\pdfoutput=1

\documentclass[11pt]{article}
\usepackage{titlesec}
\usepackage{CJKutf8}
\usepackage{titletoc}
\titleformat{\part}[hang]{\normalfont\Large\bfseries}{}{0pt}{}
\makeatletter
\@ifundefined{HCode}{%
  \usepackage{minitoc}
}{%
}
\makeatother
\usepackage[table,xcdraw]{xcolor}
\usepackage[final]{acl}

\usepackage{times}
\usepackage{latexsym}

\usepackage[T1]{fontenc}

\usepackage[utf8]{inputenc}

\usepackage{microtype}

\usepackage{inconsolata}

\usepackage{xcolor}
\usepackage{graphicx}
\usepackage{amsmath}
\usepackage{multirow}
\usepackage{tabularx}

\usepackage{longtable} 
\usepackage{array} 
\usepackage{multirow} 
\usepackage{tabularx} 
\usepackage{graphicx} 
\usepackage{graphicx}  
\usepackage{multirow}  
\usepackage{enumitem}

\title{SeedBench: A Multi-task Benchmark for Evaluating Large Language Models in Seed Science}

\author{
  \textbf{Jie Ying\textsuperscript{1}}\thanks{Equal contribution.} \quad
  \textbf{Zihong Chen\textsuperscript{1}}\footnotemark[1] \quad
  \textbf{Zhefan Wang\textsuperscript{1}}\footnotemark[1] \quad
  \textbf{Wanli Jiang\textsuperscript{1}} \quad
\\
  \textbf{Chenyang Wang\textsuperscript{1}} \quad
  \textbf{Zhonghang Yuan\textsuperscript{1}} \quad
  \textbf{Haoyang Su\textsuperscript{1}} \quad
\\
  \textbf{Huanjun Kong\textsuperscript{1}} \quad
  \textbf{Fan Yang\textsuperscript{2}} \quad
  \textbf{Nanqing Dong\textsuperscript{1,3}}\thanks{Corresponding author.}
\\
  \textsuperscript{1}Shanghai Artificial Intelligence Laboratory \quad
  \textsuperscript{2}Yazhouwan National Laboratory
  \\
  \textsuperscript{3}Shanghai Innovation Institute
}

\begin{document}
\makeatletter
\@ifundefined{HCode}{%
  \maketitle
}{%
}
\makeatother
\doparttoc
\faketableofcontents
\begin{abstract}
Seed science is essential for modern agriculture, directly influencing crop yields and global food security. However, challenges such as interdisciplinary complexity  and high costs with limited returns hinder progress, leading to a shortage of experts and insufficient technological support. While large language models (LLMs) have shown promise across various fields, their application in seed science remains limited due to the scarcity of digital resources, complex gene-trait relationships, and the lack of standardized benchmarks. To address this gap, we introduce SeedBench\footnote{\url{https://github.com/open-sciencelab/SeedBench}}—the first multi-task benchmark specifically designed for seed science. Developed in collaboration with domain experts, SeedBench focuses on seed breeding and simulates key aspects of modern breeding processes.
We conduct a comprehensive evaluation of 26 leading LLMs, encompassing proprietary, open-source, and domain-specific fine-tuned models.
Our findings not only highlight the substantial gaps between the power of LLMs and the real-world seed science problems, but also make a foundational step for research on LLMs for seed design.
\end{abstract}

\section{Introduction}
Food security is a fundamental global concern, with seeds serving as the foundation of agricultural production. However, the seed industry faces significant challenges, including its inherently interdisciplinary nature and low economic returns. These factors contribute to a persistent shortage of skilled breeding scientists, a trend expected to continue over the next decade \citep{2024Cultivating}.The critical shortage of agricultural scientists directly constrains productivity improvements and the sustainable growth of food production. 

With the rise of artificial intelligence (AI), advanced AI techniques are transforming seed science. The integration of AI-driven solutions into seed breeding aligns with projections that the next agricultural revolution will be driven by smart, digital, and precision agricultural technologies \citep{iversen2021frontier}. Large language models (LLMs), in particular, offer the ability to process vast amounts of genetic, environmental, and agronomic data, optimizing crop development. However, effectively assessing and comparing LLM capabilities requires high-quality evaluation benchmarks.


Despite the availability of LLM benchmarks for general purpose, none have been specially developed for seed breeding, which is a field critical to agricultural production and food security. Progress in this domain has been slow due to a shortage of breeding experts and limited availability of online resources. While existing agricultural benchmarks, such as AgEval (plant stress phenotyping) \citep{arshad2024ageval} and AgXQA (agricultural extension Q\&A) \citep{kpodo2024agxqa}, contribute to LLM evaluation in agriculture, they fail to address the complex decision-making and multi-step processes unique to seed breeding. The absence of a dedicated benchmark limits the ability to systematically measure LLM performance in this domain.


To bridge this gap, we introduce \textbf{SeedBench}, a multi-task benchmark designed to simulate expert decision-making across three essential seed breeding stages: (1) gene information retrieval; (2) gene function and regulation analysis; and (3) variety breeding with agronomic trait optimization. Each task category is carefully designed with information-rich, expert-validated question-answer pairs, ensuring the benchmark aligns with real-world seed breeding challenges. By providing a structured evaluation framework, SeedBench enables rigorous assessment of whether LLMs can assist human experts, accelerate breeding workflows, optimize outcomes, and advance towards autonomous intelligent seed breeding.


As the first and most comprehensive benchmark in this field, SeedBench systematically connects the capabilities of LLMs with real-world breeding challenges. Developed by experts with interdisciplinary backgrounds and Ph.D.-level expertise, SeedBench ensures both domain relevance and scientific rigor. Each question undergoes two-tier validation and refinement through machine-based assessment and human expert review, which guarantees accuracy and reliability. To accommodate diverse evaluation scenarios, SeedBench supports both one-shot and zero-shot formats, enabling a comprehensive assessment of LLMs in seed breeding.

We evaluate the performance of 26 leading LLMs on SeedBench, including 7 proprietary LLMs, 16 open-source LLMs, and 3 domain-specific models. Recognizing that LLM performance is highly sensitive to prompt structures, we further analyze multiple prompt templates to enhance robustness. Our evaluation aims to address the following research questions. 
\begin{enumerate}[label=\textbf{RQ\arabic*.}, noitemsep, topsep=0pt, leftmargin=*, align=left, listparindent=0pt]
\item What is the relationship between the reasoning ability of LLMs and the performance on seed breeding tasks? 
\item Do domain-specific fine-tuned models outperform general models in seed breeding?
\item What is the ideal model size for seed breeding tasks?
\end{enumerate}
Answering these research questions can not only facilitate the research on LLMs on seed breeding, an emerging topic, but also the understanding of usage of LLMs in the field of AI for Science.

Our main contributions are as follows:
\begin{itemize}[noitemsep, topsep=0pt, leftmargin=*]
\item SeedBench is the first benchmark designed to evaluate LLMs in seed science. 
\item SeedBench covers key seed breeding processes, ensuring reliability and accuracy through expert validation.
\item Extensive evaluations of LLMs are conducted to identify their strengths and limitations, providing insights for future AI advances in breeding.
\end{itemize}

In the following sections, Section~\ref{sec:related} reviews related work on LLMs and agricultural benchmarks. Section~\ref{sec:data} describes the construction and methodology behind SeedBench, outlining its design, task types, and validation process. Section~\ref{sec:exp} presents the experimental setup, including evaluated models and performance comparisons. Finally, we discuss findings, limitations, and future research directions for LLMs in seed breeding.

\section{Related Work}
\label{sec:related}
\subsection{Domain-Specific LLMs}
Based on the Transformer architecture \citep{vaswani2017attention}, language models have rapidly advanced, achieving key milestones in their development. Starting with foundational models such as BERT \citep{devlin2018bert} and GPT \citep{radford2018improving}, subsequent breakthroughs like GPT-4~\cite{achiam2023gpt} and DeepSeek-R1~\cite{guo2025deepseek} have demonstrated exceptional text generation.
By integrating specialized domain knowledge with continual pre-training and supervised fine-tuning, LLMs have shown potential in domains such as education \citep{gan2023large}, finance \citep{li2023large}, and healthcare \citep{mumtaz2024llms}. In the agricultural domain, LLMs are increasingly recognized for their potential to enhance food production and optimize agricultural management \citep{de2024large,kuska2024ai}. However, in contrast to the progress in these domains, leveraging LLMs for seed breeding remains an under-explored challenge in the field of AI for Science.


\begin{figure*}[ht]
  \centering
    \includegraphics[width=\linewidth]{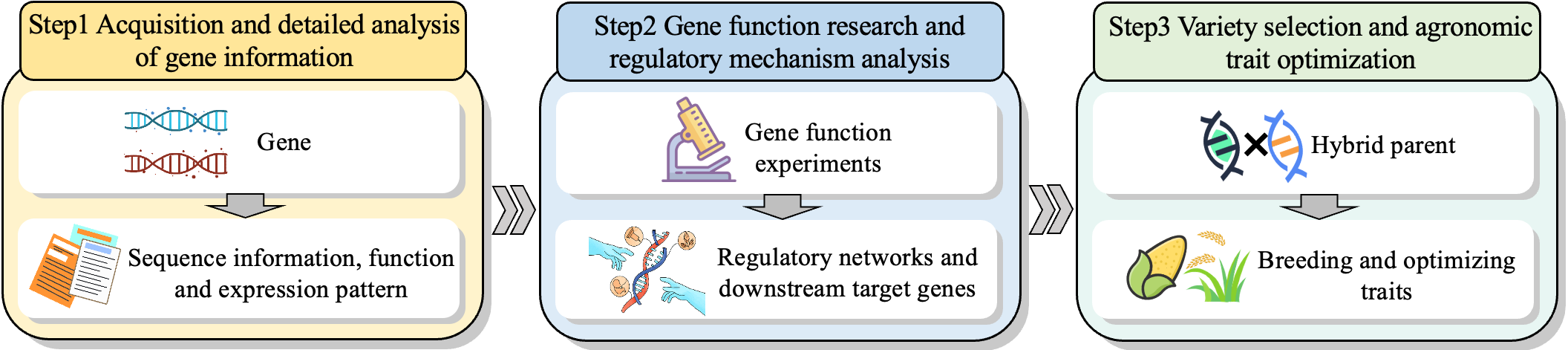}
    \caption{Breeding Expert Workflow Framework. We establish benchmark construction principles by consulting domain experts to replicate real-world seed breeding decision-making processes. (1) Gene Information Retrieval, utilizing established databases to obtain gene sequences and expression patterns; (2) Gene Function \& Regulation, employing experimental approaches (\emph{e.g.}, gene knockout, overexpression) to investigate gene roles in plant development; and (3) Variety Breeding \& Trait Optimization, implementing breeding techniques (\emph{e.g.}, hybridization, backcrossing) combined with agronomic trait selection for stable variety development.}
    \label{fig:overview}
\end{figure*}

\subsection{Domain-Specific Benchmarks}
To effectively assess and compare LLM capabilities, high-quality evaluation benchmarks are essential. While general benchmarks now cover a wide range of areas, including causal inference \citep{wang2024causalbench}, instruction following \citep{zhou2023instruction}, and safety \citep{lin-etal-2022-truthfulqa}, domain-specific benchmarks have also emerged. These specialized benchmarks are found in fields such as finance \citep{xie2024finben}, geography \citep{li2023geoglue}, healthcare \citep{chen2024gmai}, and law \citep{fei-etal-2024-lawbench}. These benchmarks help drive improvements by creating diverse datasets and tasks tailored to specific applications. However, many of these benchmarks rely heavily on website data and GPT-based re-annotation, which may limit the data diversity and depth of expertise, raising concerns about the reliability and accuracy of the results.

In addition to data limitation, another challenge is the difficulty of tasks, especially scientific tasks that require additional domain knowledge. Existing agricultural benchmarks does not cover the topic of seed breeding, a critical task in agriculture. For example, AgEval \citep{arshad2024ageval} aims for plant stress phenotyping, AgXQA \citep{kpodo2024agxqa} focuses on agricultural extension, and CROP~\citep{zhang2024empowering} evaluates the crop knowledge. 
The disparity underscores the necessity for specialized benchmarks to evaluate the performance of LLMs in seed science.

\section{SeedBench}
\label{sec:data}
This section outlines the design principles of SeedBench and details its 11 evaluation task types. SeedBench systematically evaluates LLMs in seed breeding by aligning with the typical workflow of breeding experts. It is structured into three primary categories: gene information retrieval, gene function and regulation, and variety breeding and agronomic traits. This taxonomy ensures comprehensive knowledge coverage and skill assessment.

\subsection{Breeding Process Overview}
Breeding experts typically follow three key steps in seed selection (Figure \ref{fig:overview}; further details in Appendix \ref{sec:A.1}). These steps form the basis for SeedBench’s task categorization.






\subsection{Task Taxonomy}

\begin{figure}[t]
  \centering
    \includegraphics[width=\columnwidth]{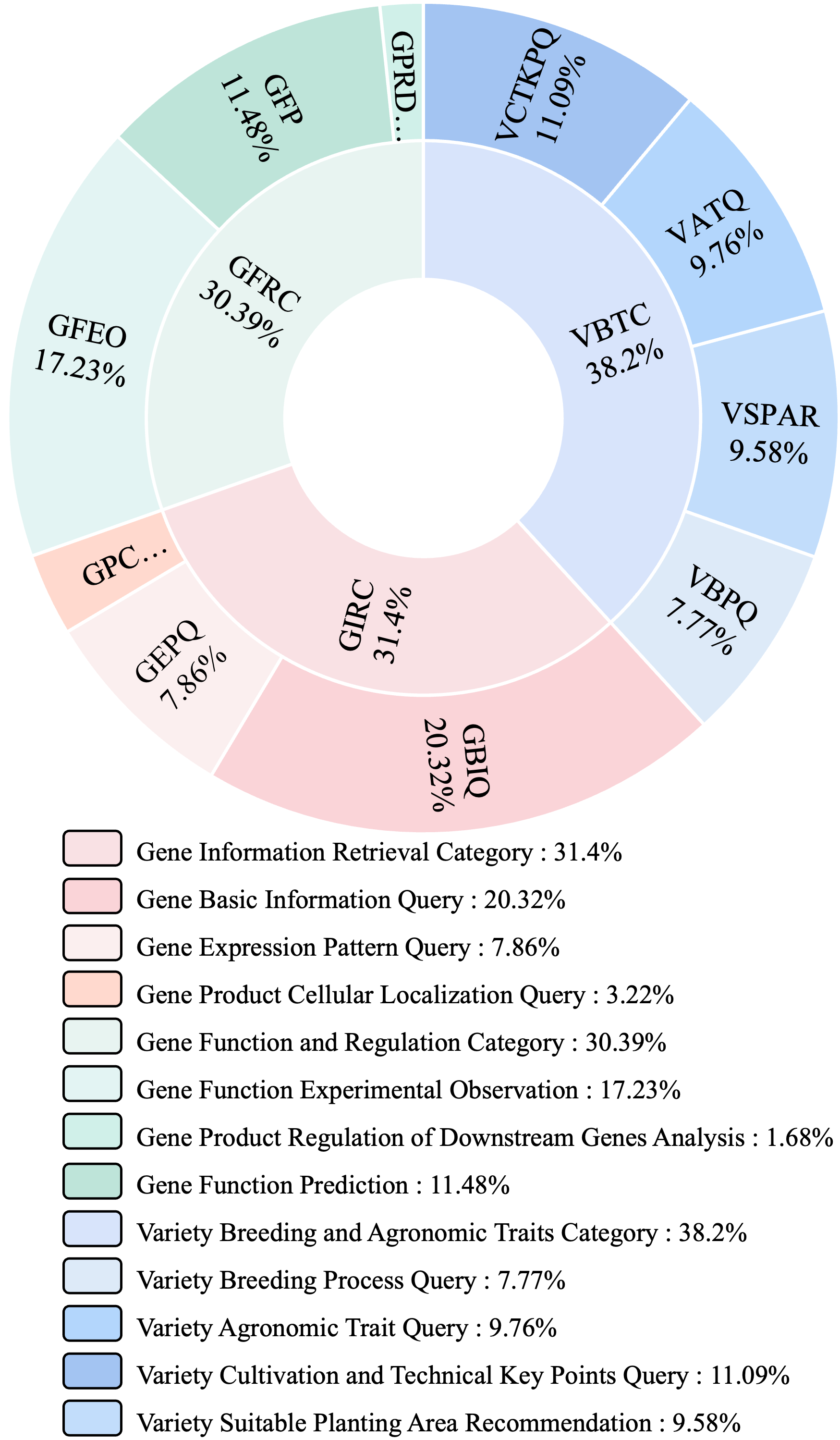}
    \caption{Benchmark Taxonomy Distribution. Three core breeding steps are further divided into ten expert-curated subcategories within SeedBench, which comprises a total of 2,264 questions. The percentages shown in the diagram represent the proportion of questions in each category relative to 2,264.}
    \label{fig:distribution}
\end{figure}

The tasks in SeedBench are categorized into three main areas, corresponding to the seed breeding workflow. These categories are further divided into ten subcategories in total, ensuring a systematic evaluation of LLM capabilities (Figure \ref{fig:distribution}). The categorization follows the practice in seed science with the help from domain experts~\citep{copeland2012principles}.

\subsubsection{Gene Information Retrieval}
LLMs retrieve essential gene information linked to specific traits, including key genes in biological processes, gene sequences, functional descriptions, and expression patterns across environments and developmental stages. They also determine the intracellular localization of gene products, mapping their distribution within the nucleus, cytoplasm, or membrane. These tasks corresponds to the initial step of the breeding process. Specific tasks include:

\begin{itemize}[noitemsep, topsep=0pt, leftmargin=*]

\item Gene Basic Information Query

\item Gene Expression Pattern Query

\item Gene Product Cellular Localization Query

\end{itemize}

\subsubsection{Gene Function and Regulation}
LLMs describe gene functions under specific experimental conditions using available data, analyze gene product regulation of downstream genes and pathways, and predict functions of uncharacterized genes. These tasks align with the second step of the breeding process. Specific tasks include: 

\begin{itemize}[noitemsep, topsep=0pt, leftmargin=*]
\item Gene Function Experimental Observation

\item Gene Product Regulation of Downstream Genes Analysis

\item Gene Function Prediction

\end{itemize}

\subsubsection{Variety Breeding and Agronomic Traits} 
LLMs gather information on breeding history, methods, objectives, and agronomic traits, including disease resistance, yield, and drought tolerance. They also propose suitable planting regions based on environmental factors. This aligns with the third step of the breeding process. Specific tasks include: 

\begin{itemize}[noitemsep, topsep=0pt, leftmargin=*]

\item Variety Breeding Process Query

\item Variety Agronomic Trait Query

\item Variety Cultivation and Technical Key Points Query

\item Variety Suitable Planting Area Recommendation

\end{itemize}

\subsection{Benchmark Construction}
\begin{figure*}[ht]
  \centering
    \includegraphics[width=1\linewidth]{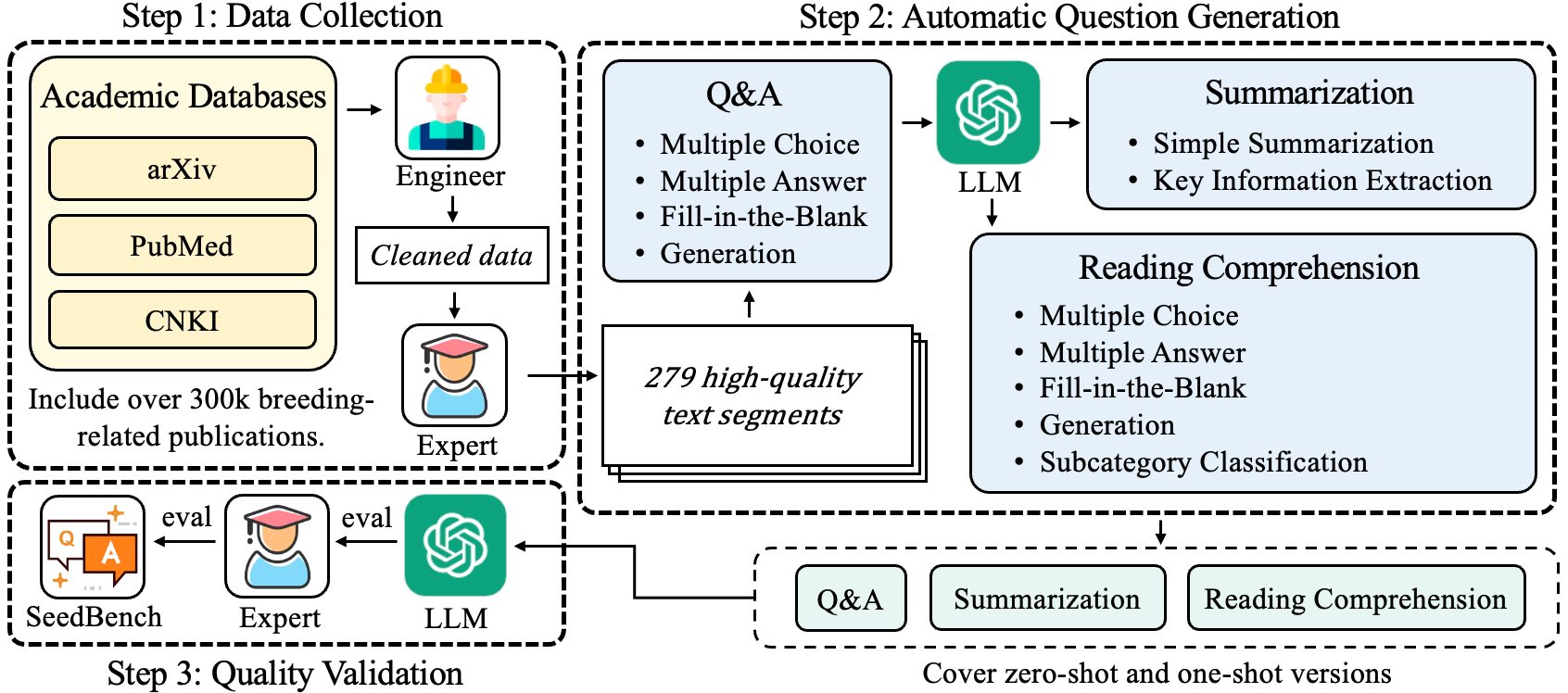}
    \caption{Benchmark Construction Pipeline. We developed SeedBench by extracting 308,727 breeding-related papers from English and Chinese sources and converting them into a unified Markdown format. The data underwent rigorous cleaning, ultimately yielding a 1.1-billion-token corpus. From this, experts curated 279 high-quality text segments, spanning 10 breeding subcategories, for generating LLM-based Q\&A tasks. Validation included both automated and expert reviews, removing low-quality entries and ensuring relevance. SeedBench offers 2,264 refined questions across 11 task types, enabling fine-grained evaluation of LLMs in seed breeding.}
    \label{fig:workflow}
\end{figure*}

SeedBench was developed through a structured three-step process: data collection, automatic question generation, and two-tier quality validation. This methodology ensures a relevant, diverse, and scientifically rigorous evaluation of LLMs.

\subsubsection{Data Collection}
SeedBench is built on a comprehensive breeding knowledge base. We extracted 308,727 breeding-related publications in English and Chinese from publicly available sources to minimize language bias\footnote{See Appendix~\ref{sec:B.2} for detailed language composition. We also discuss the potential impact of linguistic differences on the model’s performance.}. These papers were converted to Markdown using MinerU \citep{wang2024mineru} for consistency, primarily consisting of open-access academic works for credibility and reusability.

Data cleaning involved three steps: (1) heuristic filtering to remove noise, including corrupted or irrelevant data; (2) deduplication using a local-sensitive hashing method; and (3) scoring corpus segments with the CCI3-HQ-Classifier \citep{wang2024cci3} to eliminate low-quality fragments. And we used the IndustryCorpus2 Classifier\footnote{\url{https://huggingface.co/BAAI/IndustryCorpus2_Classifier}} to exclude content unrelated to seed breeding. These steps filtered out 86\% of low-quality or duplicate data, yielding a corpus of 1.1 billion tokens.

From this corpus, domain experts selected 279 high-quality text segments, sourced from 113 documents, each averaging 300 words. These segments cover 10 predefined subcategories, ensuring both depth and breadth. Each segment is highly relevant to a specific subcategory, containing multiple knowledge points essential for model evaluation. To aid in automatic question generation, domain experts manually designed 293 reference questions. 
For illustration purpose, we consistently use rice to demonstrate the benchmark construction, empirical evaluation, and case studies across the main text and Appendix, given its global importance and representativeness in seed science.\footnote{Rice is the most widely consumed crop in the world, which feeds over 3.5 billion people in the world according to Food and Agriculture Organization of the United Nations and World Bank.} The data collection and curation for maize, soy bean, wheat, and many other plants follow the same procedure.

\subsubsection{Automatic Question Generation}

\begin{table}[h]
  \centering
  \small
  \setlength{\tabcolsep}{1.5pt}
  \begin{tabular}{cccc}
    \hline
    \textbf{Type ID} & \textbf{Question Type} & \textbf{Metric} & \textbf{Count (n)} \\
    \hline
    \multicolumn{4}{l}{\textbf{Q\&A}} \\
    QA-1 & Multiple Choice & Accuracy & 199 \\
    QA-2 & Multiple Answer & Macro-F1 & 186 \\
    QA-3 & Fill-in-the-Blank & ROUGE-L & 223 \\
    QA-4 & Generation & ROUGE-L & 241 \\
    \hline
    \multicolumn{4}{l}{\textbf{Summarization}} \\
    SUM-1 & Simple Summarization & ROUGE-L & 224 \\
    SUM-2 & Key Information Extraction & ROUGE-L & 224 \\
    \hline
    \multicolumn{4}{l}{\textbf{Reading Comprehension}} \\
    RC-1 & Multiple Choice & Accuracy & 112 \\
    RC-2 & Multiple Answer & Macro-F1 & 107 \\
    RC-3 & Fill-in-the-Blank & ROUGE-L & 220 \\
    RC-4 & Generation & ROUGE-L & 239 \\
    RC-5 & Subcategory Classification & Accuracy & 278 \\
    \hline
  \end{tabular}
  \caption{Benchmark Task Types. Each high-quality text segment systematically incorporates these 11 distinct task types to ensure diversity. 'Count (n)' indicates the number of questions for this particular task type after quality validation, from the total of 2,264 questions in SeedBench. The complete distribution statistics are provided in Appendix~\ref{sec:C.2}.}
  \label{tab:tasktype}
\end{table}

The question generation phase begins with expert-designed example questions across four Q\&A types: multiple choice (QA-1), multiple answer (QA-2), fill-in-the-blank (QA-3), and generation (QA-4). This foundation extends to summarization (SUM-1, SUM-2) and reading comprehension (RC-1 to RC-5) tasks, detailed in Table~\ref{tab:tasktype}.

Utilizing GPT-4’s natural language processing capabilities, unstructured text is converted into structured knowledge dictionaries through key information extraction (Appendix~\ref{sec:F.6}), capturing genetic traits, phenotypic features, and practical applications. Based on these dictionaries, GPT-4 generates four Q\&A task types:

\begin{itemize}[noitemsep, topsep=0pt, leftmargin=*]
    \item Multiple Choice Questions

    \item Multiple Answer Questions

    \item Fill-in-the-Blank Questions

    \item Generation Questions
\end{itemize}

Then, we generated 450 summarization questions (SUM-1, SUM-2), with two per text segment, to evaluate the model’s ability to summarize breeding literature without requiring prior domain knowledge. For reading comprehension tasks, the original text segments were provided as reference documents in the augmented context. Using GPT-4, we rephrased the questions to require retrieval, analysis, reasoning, and answering based on the given document, assessing the model’s capability in long-context breeding problems. 

Additionally, we formulated 279 classification questions, where the answer corresponds to the category of each text segment, testing the model’s ability to distinguish breeding areas. All questions are available in both zero-shot and one-shot settings. SeedBench thus serves as a multidimensional benchmark, covering 10 thematic subcategories and 11 task types for fine-grained breeding assessment. All prompt templates that we used are provided in Appendix~\ref{sec:F}.

\subsubsection{Quality Validation}
Since SeedBench relies on GPT-4 for annotation, we implemented a two-stage validation process to ensure accuracy and reliability: automated machine screening and manual expert review.\\
\noindent\textbf{Automated Machine Screening.} GPT-4 first assessed coherence, logical consistency, and task adherence for each question, filtering out those with errors or contradictions. About 0.01\% of the questions were excluded at this stage.\\
\noindent\textbf{Manual Expert Review.} 
Domain experts reviewed the remaining questions for relevance and alignment with expert perspectives, removing irrelevant or weakly contextualized ones (\emph{e.g.}, ``Is rice planted in Beijing or Shanghai?'') and eliminating about 20\% of the initial set.

After validation, 2,264 high-quality questions were retained. Detailed case studies and question distribution are provided in Appendix~\ref{sec:C.2}.

\subsection{Evaluation}

The evaluation framework consists of two stages: response standardization and task-specific scoring.

\begin{itemize}[noitemsep, topsep=0pt, leftmargin=*]
\item Multiple Choice Tasks: Accuracy.

\item Multiple Answer Tasks: Macro-F1.

\item Fill-in-the-Blank: 
ROUGE-L F1 for segment-level comparison between model predictions and reference answers, averaged across segments.

\item Generation Tasks:
ROUGE-L F1 for sentence-level comparison of full generated responses against references. BERTScore included as an additional evaluation metric (Appendix \ref{sec:H.9}).
\end{itemize}
Formal mathematical definitions and post-processing methods are detailed in Appendix \ref{sec:D.2}. To account for prompt sensitivity in LLMs, Appendix \ref{sec:G} compares different prompt templates. 


\section{Experiment}
\label{sec:exp}
\subsection{Experimental Setup}
\noindent\textbf{Models.}
We compared the performance of 26 LLMs on the SeedBench, including 7 proprietary models, such as the GPT series \citep{achiam2023gpt}, Gemini series \citep{team2024gemini}, Claude-3.5-Sonnet\footnote{\url{https://www.anthropic.com/}}, and GLM-4-Plus\footnote{\url{https://open.bigmodel.cn/dev/api/normal-model/glm-4}}, as well as 16 open‑source models, including the Qwen series \citep{yang2024qwen2}, DeepSeek-V3 \citep{liu2024deepseek}, Llama series \citep{dubey2024llama}, InternLM series \citep{cai2024internlm2}, GLM-4 \citep{glm2024chatglm}, and Mistral \citep{jiang2023mistral}. Furthermore, we evaluated 3 domain-specific models, represented by the PLLaMa series \citep{yang2024pllama} and Aksara\footnote{\url{https://huggingface.co/cropinailab/aksara_v1}}. This comprehensive comparison not only highlights the relative strengths and limitations of each model group but also provides key insights for future research and application development.


\begin{table*}[!t]
  \centering
  \resizebox{1.0\textwidth}{!}{
    \small
    \setlength{\tabcolsep}{6pt}
    \begin{tabularx}{\textwidth}{l*{11}{r}}
      \hline
      \multirow{2}{*}{\textbf{Models}} & \multicolumn{10}{c}{\textbf{Breeding Subcategories}} & \multirow{2}{*}{\textbf{Average}} \\
      \cline{2-11}
       & \textbf{C1} & \textbf{C2} & \textbf{C3} & \textbf{C4} & \textbf{C5} & \textbf{C6} & \textbf{C7} & \textbf{C8} & \textbf{C9} & \textbf{C10} &  \\
      \hline
       \multicolumn{12}{l}{\textbf{Proprietary LLMs}} \\
      Claude-3.5-Sonnet & 48.77 & 57.72 & 66.02 & 57.54 & 47.82 & 49.36 & 57.47 & 60.11 & 58.06 & 58.89 & 55.45 \\
      Gemini-1.5-Pro & 47.00 & 59.55 & 62.42 & 59.56 & 43.11 & 49.55 & 53.41 & 56.18 & 52.51 & 53.71 & 53.58 \\
      Gemini-2.0-Flash & 33.67 & 27.37 & 53.04 & 32.07 & 25.87 & 44.41 & 33.57 & 36.77 & 31.78 & 31.70 & 34.24 \\
      GLM-4-Plus & 52.72 & \cellcolor{red!5}59.62 & 70.62 & \cellcolor{red!5}60.11 & 50.60 & 56.75 & \cellcolor{red!10}65.02 & \cellcolor{red!5}64.17 & \cellcolor{red!10}61.70 & \cellcolor{red!5}62.90 & \cellcolor{red!5}59.61 \\
      GPT-4 & \cellcolor{red!20}59.59 & \cellcolor{red!10}60.55 & \cellcolor{red!20}76.32 & \cellcolor{red!10}61.16 & \cellcolor{red!20}56.34 & \cellcolor{red!20}59.35 & 63.67 & \cellcolor{red!10}64.74 & \cellcolor{red!5}60.65 & \cellcolor{red!10}67.66 & \cellcolor{red!10}62.06 \\
      GPT-4o mini & \cellcolor{red!5}54.24 & 56.64 & 72.11 & 59.28 & 53.00 & \cellcolor{red!5}57.88 & 58.38 & 61.75 & 57.50 & 62.38 & 58.40 \\
      OpenAI o1-mini & 49.16 & 55.58 & 59.37 & 54.77 & 44.43 & 50.73 & 54.57 & 55.36 & 54.91 & 54.19 & 53.25 \\
      \hline
      \multicolumn{12}{l}{\textbf{Open-Source LLMs}} \\
      DeepSeek-V3-671B & \cellcolor{red!10}56.03 & \cellcolor{red!20}62.42 & \cellcolor{red!10}74.81 & \cellcolor{red!20}63.17 & \cellcolor{red!10}55.23 & \cellcolor{red!10}58.84 & \cellcolor{red!20}68.23 & \cellcolor{red!20}69.04 & \cellcolor{red!20}66.46 & \cellcolor{red!20}68.48 & \cellcolor{red!20}63.30 \\
      GLM-4-Chat-9B & 23.28 & 21.31 & 39.97 & 26.13 & 16.20 & 34.15 & 26.63 & 29.60 & 25.60 & 26.68 & 26.55 \\
      InternLM2-7B & 27.55 & 21.14 & 39.64 & 28.57 & 15.16 & 36.12 & 28.74 & 30.80 & 27.32 & 29.22 & 28.71 \\
      InternLM2.5-7B & 51.71 & 55.75 & 67.88 & 50.48 & 44.14 & 56.73 & 51.28 & 54.91 & 52.46 & 56.24 & 53.51 \\      
      Llama3.1-8B & 43.89 & 31.21 & 42.53 & 40.68 & 38.47 & 43.80 & 42.87 & 51.62 & 41.88 & 40.91 & 42.23 \\
      Llama3.1-70B & 48.72 & 55.41 & 64.77 & 53.67 & 46.73 & 54.08 & 56.94 & 57.72 & 55.31 & 57.56 & 54.30 \\
      Llama3.3-70B & 45.32 & 47.15 & 60.62 & 49.76 & 40.90 & 54.30 & 52.79 & 54.61 & 49.98 & 55.05 & 50.53 \\
      Mistral-v0.3-7B & 42.61 & 38.28 & 57.02 & 40.41 & 29.97 & 44.22 & 36.31 & 43.98 & 39.92 & 43.51 & 41.59 \\
      Qwen2-0.5B & 32.84 & 25.15 & 40.19 & 28.20 & 27.62 & 37.22 & 33.81 & 33.63 & 28.25 & 31.67 & 31.44 \\
      Qwen2-7B & 44.21 & 40.41 & 63.00 & 47.36 & 35.37 & 52.30 & 45.61 & 48.73 & 44.88 & 46.89 & 46.51 \\
      Qwen2-57B & 53.67 & 49.81 & \cellcolor{red!5}74.30 & 58.38 & 39.34 & 54.71 & \cellcolor{red!5}63.89 & 59.57 & 59.22 & 60.08 & 57.20 \\
      Qwen2-72B & 51.16 & 58.10 & 74.07 & 59.72 & \cellcolor{red!5}51.58 & 57.76 & 58.85 & 61.63 & 56.69 & 59.11 & 57.62 \\
      Qwen2.5-7B & 45.16 & 39.50 & 66.01 & 44.61 & 35.72 & 50.00 & 53.60 & 53.31 & 53.06 & 51.05 & 48.45 \\
      Qwen2.5-14B & 50.91 & 50.73 & 68.62 & 52.15 & 47.14 & 54.54 & 57.02 & 62.05 & 54.37 & 54.15 & 54.21 \\
      Qwen2.5-72B & 46.86 & 47.41 & 70.99 & 51.89 & 46.17 & 57.60 & 55.35 & 56.31 & 53.05 & 54.75 & 52.63 \\
      QwQ-32B & 32.24 & 21.06 & 47.11 & 29.14 & 28.56 & 39.68 & 38.17 & 39.56 & 34.70 & 34.52 & 33.55 \\
      \hline
      \multicolumn{12}{l}{\textbf{Domain Specific LLMs}} \\
      Aksara-v1-7B & 36.72 & 36.69 & 48.32 & 35.41 & 24.26 & 36.83 & 31.17 & 34.64 & 31.15 & 34.14 & 35.04 \\
      PLLaMa-7B & 17.85 & 13.69 & 17.99 & 16.81 & 11.66 & 21.67 & 14.34 & 17.36 & 12.39 & 16.11 & 16.46 \\
      PLLaMa-13B & 15.10 & 14.18 & 28.41 & 18.83 & 13.96 & 23.28 & 18.53 & 17.37 & 14.15 & 18.51 & 17.57 \\
      \hline
    \end{tabularx}}
  \caption{Evaluation of 26 LLMs on SeedBench. Performance (averaged across both zero-shot and one-shot configurations) is stratified by breeding subcategories, with open-source/domain-specific models evaluated through 3 repeated trials (mean scores reported). The scores represent averages across three different metrics for 11 task types. The columns delineate ten subcategories in breeding: (C1) Gene Basic Information Query, (C2) Gene Expression Pattern Query, (C3) Gene Product Cellular Localization Query, (C4) Gene Function Experimental Observation, (C5) Gene Product Regulation of Downstream Genes Analysis, (C6) Gene Function Prediction, (C7) Variety Breeding Process Query, (C8) Variety Agronomic Trait Query, (C9) Variety Cultivation and Technical Key Points Query, (C10) Variety Suitable Planting Area Recommendation. Top-$3$ performers per column are highlighted in red. Extended results (including \textbf{standard deviations}, \textbf{separate} zero-shot/one-shot scores, and task-type breakdowns) are provided in Appendix~\ref{sec:H.10}.}
  \label{tab:performance}
\end{table*}

\noindent\textbf{Implementation Details.} We evaluated the performance of all models in both zero-shot and one-shot settings. In zero-shot inference, the model input includes only the task instructions and the query. In one-shot inference, the model input consists of the task instructions, an example query with its answer, followed by the actual query.
The experiments were conducted using the OpenCompass\footnote{\url{https://opencompass.org.cn/home}} framework. For proprietary LLMs, we performed inference through their APIs. The evaluation of each model took approximately 1 hour. For open-source LLMs, the evaluation was conducted on 8 NVIDIA A100 40GB GPUs, with an average completion time of 0.5 hours. The inference hyperparameters are detailed in Appendix~\ref{sec:D.1}.

\subsection{Performance Evaluation}
Here we assess overall performance across breeding  subcategories, as detailed in Table~\ref{tab:performance}\footnote{See Tables~\ref{tab:performance_std}~\ref{tab:res_task_type_zeroshot}~\ref{tab:res_task_type_oneshot} in Appendix~\ref{sec:H.10} for complete results.}.
Among proprietary LLMs, GPT-4 achieves the highest average score on SeedBench (62.06), followed by GLM-4-plus (59.61). In contrast, open-source models show a different ranking: DeepSeek-V3 leads with an average score of 63.3, outperforming Qwen2.5-14B, which scores 54.3. Notably, DeepSeek-V3, despite being a recently released model with 671B parameters, surpasses GPT-4 on SeedBench.
On the other hand, the three domain-specific LLMs perform relatively poorly, likely due to their limited conversational and instruction following capabilities. Interestingly, OpenAI o1, despite demonstrating strong reasoning abilities in mathematics and coding, scores lower than GPT-4 on SeedBench. This suggests that its reasoning strategy does not transfer effectively to breeding-related tasks. A similar trend is observed in Gemini-2.0-Flash Thinking and QwQ-32B, both of which exhibit explicit reasoning steps in their responses yet achieve only 34.24 and 33.55, respectively.
A more detailed comparative analysis is provided in Appendix~\ref{sec:H}, where we examine performance variations within the same model series, evaluate models of identical sizes, and assess the impact of task difficulty. And Appendix~\ref{sec:E} provides a holistic error taxonomy and analysis of models' failures, summarizing eight primary causes of errors, such as Gene Name Confusion. These additional analyses collectively offer deeper insights into the factors influencing model performance on SeedBench.

\subsection{Empirical Analysis}

\subsubsection{Analysis on Reasoning Ability}
Table~\ref{tab:performance} shows that LLMs with explicit ``reasoning mode'' do not consistently outperform those without specialized chain-of-thought mechanisms (\textbf{RQ1}). For instance, models designed for multi-step reasoning, such as Gemini-2.0-Flash (34.24) and QwQ-32B (33.55), achieve lower average scores than general-purpose LLMs such as GPT-4 (62.06) and the top-performing open-source model, DeepSeek-V3-671B (63.30). This discrepancy is evident in subcategories that primarily involve fact retrieval or straightforward inference, such as Gene Basic Information Query (C1) and Variety Cultivation and Technical Key Points Query (C9).

Furthermore, the results suggest potential drawbacks of verbose chain-of-thought reasoning in tasks that require single-step inferences. Lengthy reasoning chains can introduce unnecessary content, potentially reducing performance on precision-based metrics such as ROUGE. This effect is particularly evident in tasks like Gene Product Regulation of Downstream Genes Analysis (C5), where the best-performing models (\emph{e.g.}, GPT-4 and DeepSeek-V3-671B) maintain conciseness while effectively capturing key information.
Overall, these findings indicate that while reasoning-centric prompts may be advantageous in complex multi-step tasks (\emph{e.g.}, coding or mathematical problem-solving), seed breeding queries rely on direct knowledge retrieval or limited inference.
Future research could focus on adjusting the reasoning strategy to match task complexity, dynamically adjusting reasoning chains based on task difficulty, rather than using complex reasoning for all tasks \citep{chen2024not}.

\subsubsection{Impact of Domain-Specific Fine-Tuning}
Contrary to expectations, domain-specific fine-tuned models (\emph{e.g.}, Aksara-v1-7B, PLLaMa-7B, PLLaMa-13B) perform worse than general-purpose models on 
SeedBench. As shown in Table~\ref{tab:performance}, these specialized models exhibit significantly lower overall scores (\emph{e.g.}, 35.04 for Aksara-v1-7B and 17.57 for PLLaMa-13B). In comparison, mid-tier open-source LLMs such as Llama3.1-70B (54.30) and Qwen2.5-14B (54.21) outperform them, while top-performing general-purpose models like DeepSeek-V3-671B (63.30) and GPT-4 (62.06) achieve even higher scores. Notably, Aksara-v1-7B’s best subcategory score (48.32 in C3: Gene Product Cellular Localization Query) remains below many open-source models’ performance in similar tasks.
We hypothesize that a key factor behind this underperformance is the deterioration of conversational and instruction-following capabilities, after fine-tuning on domain-specific data. In one-shot evaluations, PLLaMas frequently misinterpret prompts, treating in-context examples as direct queries. This shows that domain specialization may weaken general instruction adherence, which is essential for tasks requiring customized outputs and complex user interactions. Additionally, post-training on narrowly defined tasks or using data not validated by experts may further diminish models' performance. Thus, we suggest that maintaining general-purpose abilities during fine-tuning, expanding the breadth of training data, and incorporating expert validation may effectively improve the performance of domain-specific fine-tuned models. Further investigations into the fine-tuning strategy and the scope and quality of the training corpus are required to fully answer \textbf{RQ2}. 



\begin{figure}[t]
    \includegraphics[width=\columnwidth]{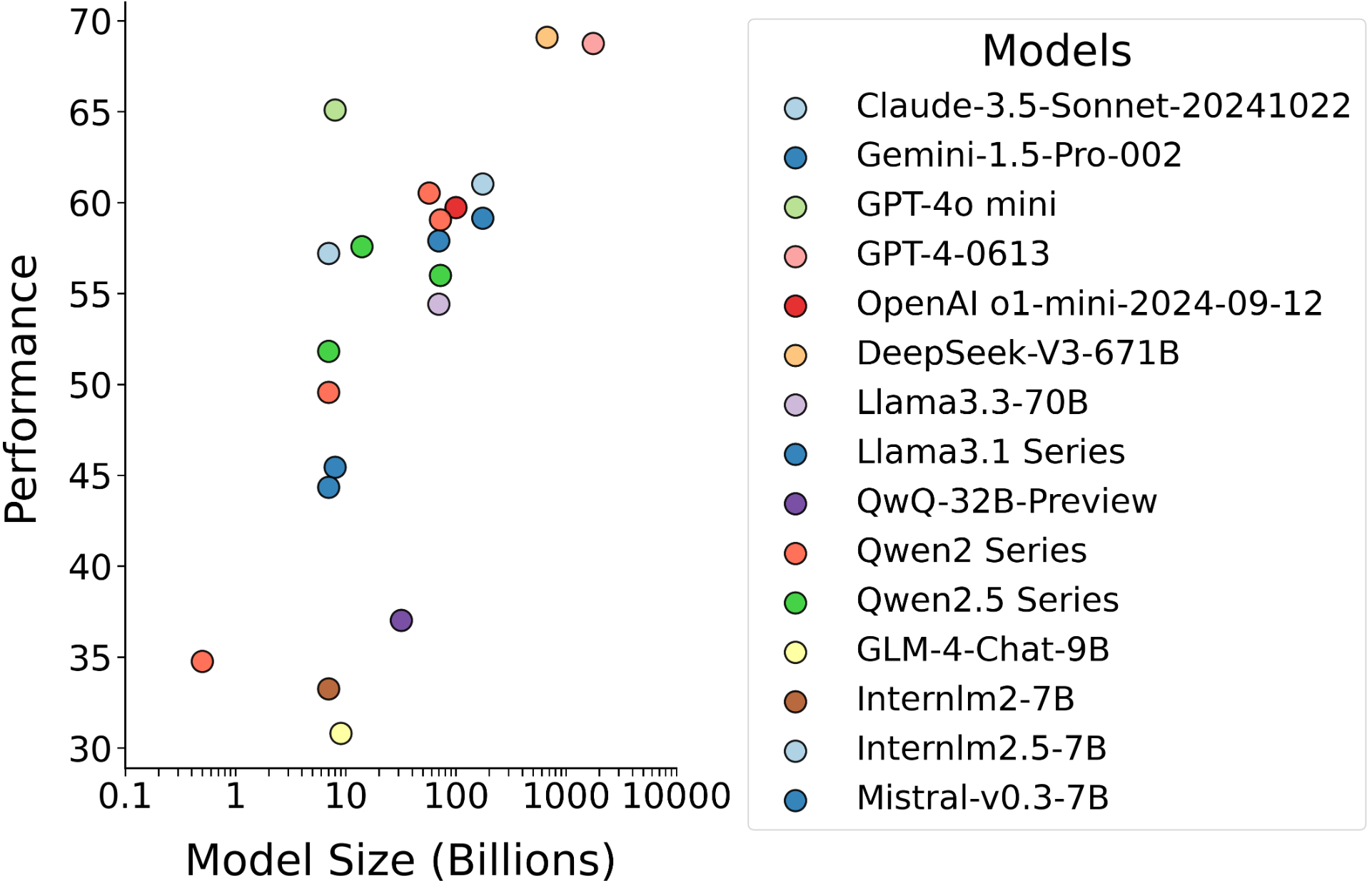}
    \caption{Performance vs. Model Size. 
    We empirically validate scaling laws in seed breeding tasks, showing a logarithmic correlation between model size and average scores. The optimal model size for breeding tasks lies between 7B and 14B, balancing performance and computational efficiency.}
    \label{fig:modelsize}
\end{figure}

\subsubsection{Impact of Model Size}
As shown in Figure~\ref{fig:modelsize}, models with 7B–14B parameters achieve the best trade-off between performance and computational efficiency (\textbf{RQ3}). Models in this range, such as Qwen2.5-14B and InternLM-2.5-7B, perform robustly with manageable resource requirements.
In contrast, smaller models (\emph{e.g.}, Qwen2-0.5B) underperform, and much larger models (\emph{e.g.}, Qwen2.5-72B) show diminishing returns. Additionally, the lack of performance improvement across scaled Qwen models (14B to 72B) suggests a distribution mismatch between our benchmark data and Qwen’s enlarged training corpus. We conjecture that the quality of training corpus is more important than model size for domain-specific tasks. In addition, we analyze performance differences between models of varying sizes, series, and subcategory difficulty in Appendices~\ref{sec:H.1} to~\ref{sec:H.8}.

\section{Discussion}
\label{sec:discuss}
The discrepancy between the seed scientists' expectations and the reality of LLMs is evident. From the perspective of breeding experts, the immediate applicability of domain-specific fine-tuned models to seed science remains constrained. This may arise from factors such as training on narrowly defined tasks, reliance on data lacking expert validation, or catastrophic forgetting of general capabilities during fine-tuning. Conversely, while current general LLMs have shown good potential in text understanding and basic reasoning, they still fall short of meeting the deep and specialized requirements of actual breeding work. Several key gaps between seed scientists' expectations and the capabilities of general LLMs are highlighted below.\\
\noindent \textbf{Domain Depth.} While LLMs cover a broad range of agricultural topics, their knowledge depth is insufficient for specialized breeding tasks. Complex issues like molecular breeding or trait introgression often require high-quality domain knowledge repositories, an area where current models are lacking.
One potential solution is to integrate structured knowledge graphs mapping relationships across phenotypic, genomic, and environmental data.\\
\noindent \textbf{Multimodal Integration.} Breeding in practice heavily relies on sensory evaluations (such as detecting grain morphology, texture, and odor) and environmental data (such as climate and soil conditions). Current LLMs, primarily based on text input, cannot effectively integrate images, sensor readings, and field observations, limiting their performance in multimodal decision-making.\\
\noindent \textbf{Explainability and Risk Management.} 
The breeding decision cycle is costly and prolonged. Without transparent reasoning and risk evaluation mechanisms, misleading outputs could result in substantial losses. Producers must be able to verify and trace model conclusions to ensure safe and controllable implementation, \emph{e.g.}, via safe RL.

\section{Conclusion}
\label{sec:conclu}
In this study, we propose SeedBench, the first multi-task LLM benchmark tailored for seed science. The contributions of SeedBench to AI for Science are twofold. First, it demonstrates the complete benchmark construction process for seed science, a knowledge-intensive field. The knowledge behind can be transferred to other science domains, such as life science or physical science, to build a comprehensive and reliable scientific benchmark. Second, it evaluates the capabilities of LLMs on addressing seed breeding tasks. The results not only provide insightful empirical findings on the tasks on interest, but also pose future research directions on designing scientific LLM. By bridging the gap between the power of LLMs and the real-world scientific problem in seed science, we aim to make a foundational effort for successful implementation of LLMs for seed design in the future.
\section*{Limitations}
Given the highly specialized nature of breeding research, we primarily use the peer-reviewed scientific literature to ensure the accuracy and credibility of the data. Online sources often lack systematic review and professional validation. In the future, we will explore more reliable online databases and expert knowledge bases to further diversify our data sources.

 
There remains a gap between the expectations of the researchers and the actual capabilities of LLMs. Closing this gap calls for more domain-focused and in-depth professional datasets, as well as additional breeding-specific knowledge during model training. Another key direction is to incorporate sequential decision making and iterative learning into the models, so that they can adapt to planting cycles and experimental feedback. Additionally, developing LLMs that support multimodal inputs—such as phenotypic, genomic, and environmental data—would be crucial for complex breeding scenarios. Finally, improving model interpretability and safety mechanisms will be essential for building trust when the model provides breeding recommendations.

\section*{Ethical Considerations}
This research adheres to the ethical principles outlined in the \textit{ACL Code of Ethics}. We have taken steps to ensure that our work does not cause harm, particularly in the context of seed science, where the implications of our research could impact agricultural practices and food security.

We have carefully considered the potential risks of our approach, especially with respect to the use of LLMs in agricultural research. One of the main ethical concerns we addressed is the potential for bias in LLMs, which could affect seed breeding decisions or lead to misinterpretations in agricultural data. To mitigate this, we evaluated multiple LLMs, to better understand their limitations and potential biases, and have explicitly highlighted areas where further improvement is needed.

In addition, we ensured that all data used in our experiments were sourced responsibly, with due regard for privacy and intellectual property. The artifacts we used are governed under the CC-BY 4.0, Apache 2.0 and MIT licenses, which support open and ethical use of such resources. The creation of the SeedBench benchmark involved collaboration with domain experts to ensure that it accurately represents the complexities of seed breeding, while minimizing any unintended consequences Moreover, we have taken care to ensure that the use of these artifacts aligns with the intended purpose of advancing research in seed breeding and AI applications.

Finally, we have considered the broader social impact of our work, recognizing that our research could influence farming practices, genetic resource management, and food security. We have outlined the necessary steps to avoid harm and ensure responsible application of our findings.

\section*{Acknowledgments}
This work was supported by Shanghai Artificial Intelligence Laboratory, the Yazhouwan National Laboratory Project (grant no. 2310CF01), and the Hainan Yazhou Bay Seed Laboratory Project (grant no. B21HJ0001). The authors thank anonymous reviewers for their constructive comments. The authors also thank Yazhouwan National Laboratory for data collection and validation of human experts. The authors also thank Songyang Zhang from Shanghai Artificial Intelligence Laboratory for academic discussion.

\bibliography{custom}

\newpage

\appendix
\onecolumn

\label{sec:appendix}
\startcontents
\part{Appendix}
\mtcaddpart{}
\parttoc
\newpage
\section{Background Definitions}
\label{sec:A}
\subsection{Definition of Breeding Competencies}
\label{sec:A.1}
We define in Table~\ref{tab:core_breeding_competencies} the core breeding competencies required to address complex, knowledge-intensive breeding tasks. The breeding process is divided into three steps: the acquisition and analysis of genetic information, the exploration of gene function and regulatory mechanisms, and variety selection alongside agronomic trait optimization. Each stage corresponds to specific knowledge and skill requirements.

\begin{table}[htbp!]
\centering
\resizebox{\textwidth}{!}{%
\begin{tabular}{p{4.3cm} p{3.7cm} p{3.7cm} p{6cm}}
\hline
\textbf{Stage} & \textbf{Corresponding Task} & \textbf{Competency} & \textbf{Description} \\
\hline
\multirow{3}{4.3cm}{\raggedright \textbf{Acquisition and\\ Analysis of\\ Genetic Information}}
& Basic Gene Information Retrieval 
& Ability to Retrieve Basic Gene Information 
& Is able to retrieve and integrate a gene’s basic information (e.g., sequence, functional annotation, chromosomal location) based on user inputs such as gene names or identifiers. \\
\cline{2-4}
& Gene Expression Pattern Retrieval
& Ability to Analyze Gene Expression Patterns 
& Possesses the capacity to explore and interpret gene expression data under different tissues, developmental stages, or environmental conditions. \\
\cline{2-4}
& Subcellular Localization of Gene Products
& Ability to Determine Gene Product Localization and Characteristics
& Uses existing protein or molecular data to infer the specific location of a gene product within the cell (e.g., nucleus, cytoplasm, cell membrane) and, by applying biological knowledge, deduces its potential functions (e.g., transcriptional regulation, signal transduction, material transport). \\
\hline

\multirow{3}{4.3cm}{\raggedright \textbf{Exploration of\\ Gene Function and\\ Regulatory Mechanisms}}
& Observations from Gene Function Experiments
& Ability to Interpret Experimental Results on Gene Function
& Accurately describes how a gene influences plant traits under specific conditions. \\
\cline{2-4}
& Analysis of Gene Product Regulation on Downstream Genes
& Ability to Analyze Gene Regulatory Networks and Downstream Genes
& Relies on existing research or deductive reasoning to evaluate the regulatory effects of a target gene product on downstream genes or related pathways. \\
\cline{2-4}
& Gene Function Prediction
& Ability to Predict and Validate Unknown Gene Functions
& Draws on known gene sequences, expression characteristics, and analogies with characterized genes to predict functions of genes that have not yet been extensively studied. \\
\hline

\multirow{4}{4.3cm}{\raggedright \textbf{Variety Selection\\ and Agronomic\\ Trait Optimization}}
& Retrieval of Variety Breeding Processes
& Ability to Oversee Variety Breeding Processes
& Collects and summarizes the breeding history, methods, and improvement goals for a specific variety. \\
\cline{2-4}
& Querying Agronomic Traits of Varieties
& Ability to Screen and Evaluate Agronomic Traits
& Compares the agronomic traits (e.g., disease resistance, yield, drought tolerance) of a target variety or multiple varieties to assess their value in practical production, selecting the most suitable breeding materials or gene combinations. \\
\cline{2-4}
& Querying Key Points of Variety Cultivation and Techniques
& Ability to Plan Variety Cultivation Management and Key Technical Points
& Gathers and analyzes crucial information on cultivation management (e.g., fertilization strategies, irrigation schedules, pest control) to optimize field trials or practical planting outcomes. \\
\cline{2-4}
& Recommending Suitable Planting Regions for Varieties
& Ability to Evaluate and Recommend Planting Regions
& Takes into account environmental factors and agronomic traits to propose suitable planting regions for a given variety, while evaluating its potential for wider application. \\
\hline

\end{tabular}%
}
\caption{Core Breeding Competencies}
\label{tab:core_breeding_competencies}
\end{table}

\subsection{Example Definitions of Breeding Terminology}
\label{sec:A.2}
We define in Table~\ref{tab:breeding_terminology} the specialized breeding terms mentioned in this paper, along with explanations to provide a unified reference for subsequent experimental design, task classification, and result interpretation.

\begin{table}[htbp!]
\centering
\resizebox{\textwidth}{!}{%
\begin{tabular}{p{4cm} p{12cm}}
\hline
\textbf{Terminology} & \textbf{Definition} \\
\hline
\textbf{Gene Information} 
& Refers to the fundamental data of a given gene, including its sequence, structure, functional descriptions, and expression patterns. This information is typically obtained from genomic databases or sequencing results, serving as the starting point for subsequent gene function research and variety improvement. \\
\hline
\textbf{Gene Expression Pattern} 
& Describes a gene’s expression levels and dynamic changes across various tissues, developmental stages, or environmental conditions. By analyzing this pattern, one can determine the gene’s potential impact on target traits (e.g., drought tolerance). \\
\hline
\textbf{Gene Regulatory Network} 
& A molecular regulatory system formed by interactions among gene products (e.g., transcription factors, enzymes). It determines how crops respond to environmental stimuli or developmental requirements and serves as an essential foundation for precision breeding analyses. \\
\hline
\textbf{Hybridization \& Backcrossing} 
& Involves breeding new offspring by crossing or pollinating different parental lines. Backcrossing is carried out with a desirable parent line over multiple generations to stabilize or strengthen specific traits. This approach is traditional yet effective, often combined with modern molecular marker techniques. \\
\hline
\textbf{Agronomic Trait} 
& Refers to key field performance characteristics of crops, such as plant height, number of tillers, yield, disease resistance, and maturity. These traits are crucial indicators for evaluating the economic value and potential for broader adoption of new varieties. \\
\hline
\textbf{Marker-Assisted Breeding} 
& Uses molecular markers (e.g., SNP, SSR) for genotypic screening to accelerate the breeding process and improve selection accuracy. This method is frequently employed during hybridization and backcrossing to rapidly pinpoint genes associated with targeted traits. \\
\hline
\textbf{LLM-Aided Breeding} 
& Leverages large language models (e.g., GPT) to analyze and reason over literature, databases, and experimental data, supporting gene information analysis, gene function prediction, and variety selection decisions. It is expected to shorten breeding cycles and enhance breeding efficiency. \\
\hline
\end{tabular}%
}
\caption{Breeding Terminology Definitions}
\label{tab:breeding_terminology}
\end{table}

\newpage
\section{Source Data Collection}
\label{sec:B}
\subsection{Construction and Annotation of High-Quality Text Segments}
\label{sec:B.1}
In Figure~\ref{fig:source_data_collection} below, \texttt{Content} is a snippet extracted by agricultural experts from selected scientific publications. \texttt{Example Question} is an illustrative question provided by the experts, demonstrating from which perspective or angle one may inquire about the snippet. \texttt{Classification} indicates the subfield of agriculture to which the snippet belongs, and \texttt{Reference} identifies the scientific publication from which the snippet originates.

\begin{figure}[htbp!]
\centering
\includegraphics[width=0.95\textwidth]{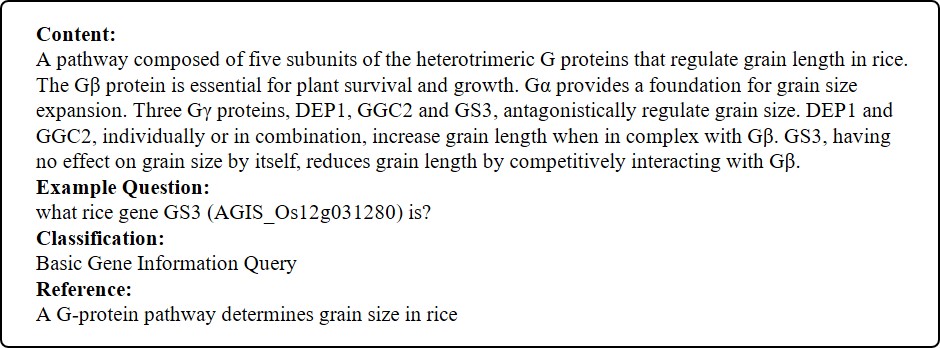}
\includegraphics[width=0.95\textwidth]{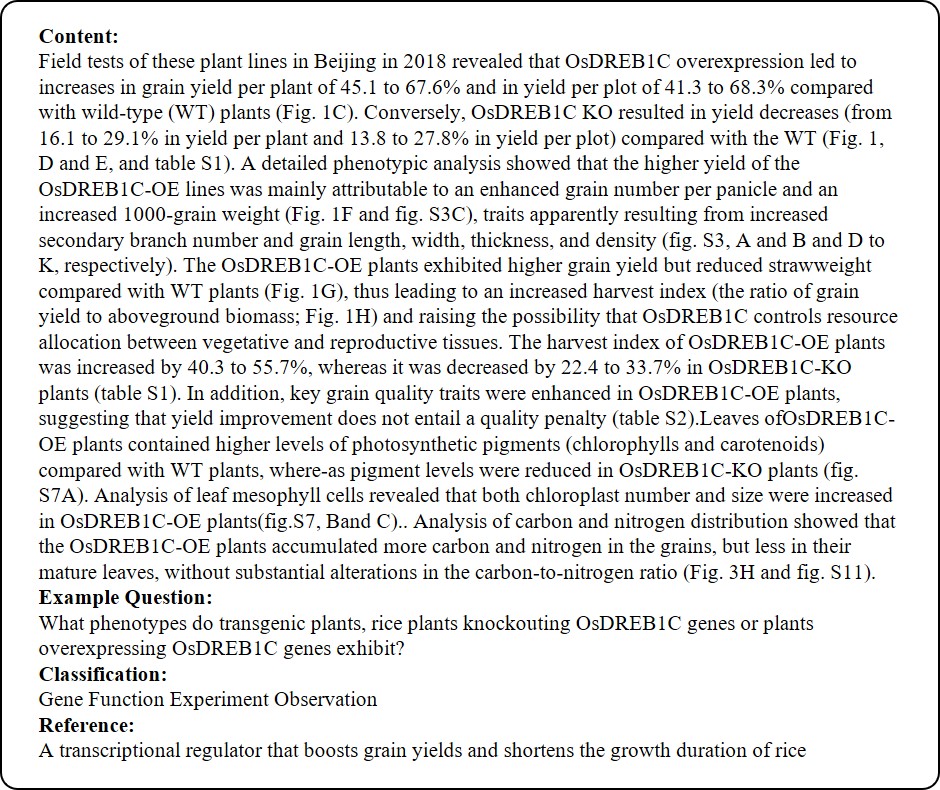}
\caption{An illustration of the Content, Example Question, Classification, and Reference fields.}
\label{fig:source_data_collection}
\end{figure}

\subsection{Language Composition and Impact}
\label{sec:B.2}
(1) The initial corpus of 308,727 articles comprises 63\% English and 37\% Chinese, with this imbalance reflecting the greater availability of English publications; (2) After cleaning, the 1.1 billion-token corpus consists of 75\% English and 25\% Chinese, with this shift due to the higher accuracy of the MinerU in processing English texts; (3) The final 279 segments used in SeedBench include 49\% English and 51\% Chinese, achieving balance through manual selection by breeding experts; (4) The 2,264 questions in SeedBench include 45\% English and 55\% Chinese.

Although LLMs exhibit strong cross-lingual capabilities and the linguistic differences in breeding questions do not alter the underlying scientific logic, we observed response drift when posing the same question in English and Chinese (with cleared histories). This phenomenon suggests potential issues in cross-lingual consistency, which deserves further research---especially when applying LLMs in specific domains where alignment across languages is critical:
\begin{itemize}[leftmargin=1.5em]
    \item \textbf{Question1-EN}: \textit{What effect did the overexpression of OsDREB1C have on the levels of photosynthetic pigments in the leaves of the plants?} \\
    A. Increased pigment levels \\
    B. Decreased pigment levels \\
    C. Pigment levels fluctuated unpredictably \\
    D. No significant change in pigment levels

    \item \textbf{Question1-CN}: \textit{\begin{CJK}{UTF8}{gbsn}OsDREB1C 的过表达对植物叶片中光合色素水平有何影响？\end{CJK}} \\
    A. \begin{CJK}{UTF8}{gbsn}增加色素水平\end{CJK} \\
    B. \begin{CJK}{UTF8}{gbsn}减少色素水平\end{CJK} \\
    C. \begin{CJK}{UTF8}{gbsn}色素水平不可预测地波动\end{CJK} \\
    D. \begin{CJK}{UTF8}{gbsn}色素水平没有显著变化\end{CJK}

    \item \textbf{Question2-EN}: \textit{The expression profile of OsDT11 in different rice tissues was analyzed by \underline{\hspace{3cm}}.}
    
    \item \textbf{Question2-CN}: \textit{\begin{CJK}{UTF8}{gbsn}OsDT11 在不同水稻组织中的表达谱通过\underline{\hspace{3cm}}分析。\end{CJK}}
\end{itemize}

\vspace{1em}
\begin{table}[h]
\centering
\small
\setlength{\tabcolsep}{5pt}
\caption{Model responses to parallel English and Chinese domain-specific questions.}
\begin{tabular}{lcccc}
\hline
\textbf{Model} & \textbf{Ans1-EN} & \textbf{Ans1-CN} & \textbf{Ans2-EN} & \textbf{Ans2-CN} \\
\hline
DeepSeek-V3-671B & A & A & qRT-PCR & \begin{CJK}{UTF8}{gbsn}实时荧光定量PCR\end{CJK} \\
GPT-4 & A & A & qRT-PCR & \begin{CJK}{UTF8}{gbsn}转录组学\end{CJK} \\
OpenAI o1-mini & A & A & qRT-PCR & \begin{CJK}{UTF8}{gbsn}Northern blot分析\end{CJK} \\
Gemini-1.5-Pro & A & A & qRT-PCR & qRT-PCR \\
\hline
\end{tabular}
\end{table}

\section{Quality Verification}
\label{sec:C}
The manual QA verification process involves a comprehensive assessment by experts to ensure the clarity, relevance, and rationality of questions. This includes evaluating whether the questions are designed to elicit accurate and valuable answers. Experts also check the correctness of the answers by consulting authoritative sources and applying professional knowledge to ensure they are error-free. Additionally, for questions that include multiple-answer options, it is essential to assess the rationality of these options. This assessment guarantees a clear distinction between them and ensures their close relevance to the correct answer. To enhance transparency regarding the expert correction process and the resolution of disagreements, we outline the following three points: (1) The composition of the expert panel, consisting of six Ph.D.-level experts in seed breeding; (2) An iterative review process, wherein each question is independently evaluated by at least two experts; (3) A disagreement resolution mechanism that addresses subjectivity in expert assessments, resolved by adopting the intersection of differing expert corrections. Additionally, we have open-sourced samples discarded during the expert correction process as “bad cases” on GitHub, offering readers further insight and reference.

\subsection{Manual Quality Verification}
\label{sec:C.1}
\begin{figure}[h!]
\centering
\includegraphics[width=0.95\textwidth]{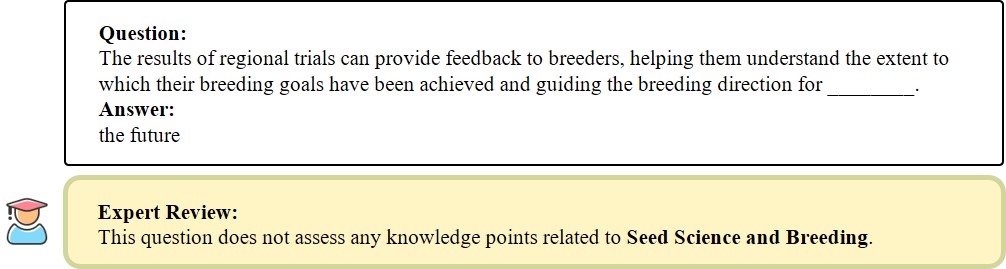}
\end{figure}

\begin{figure}[h!]
\centering
\includegraphics[width=0.95\textwidth]{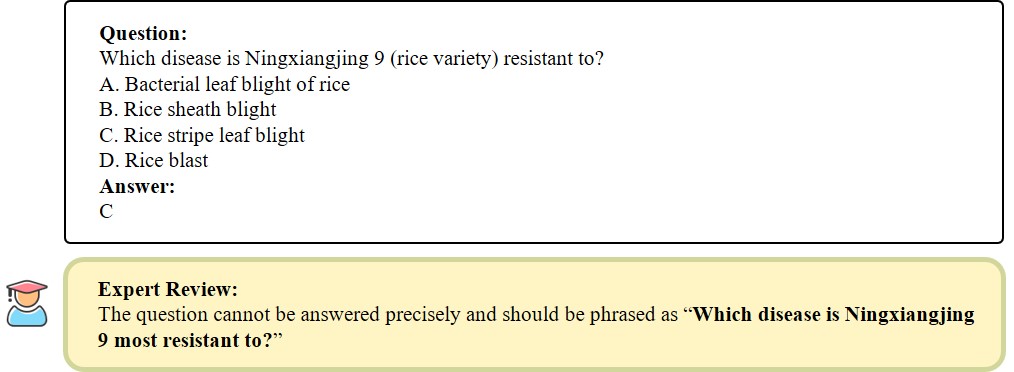}

\end{figure}

\begin{figure}[h!]
\centering
\includegraphics[width=0.95\textwidth]{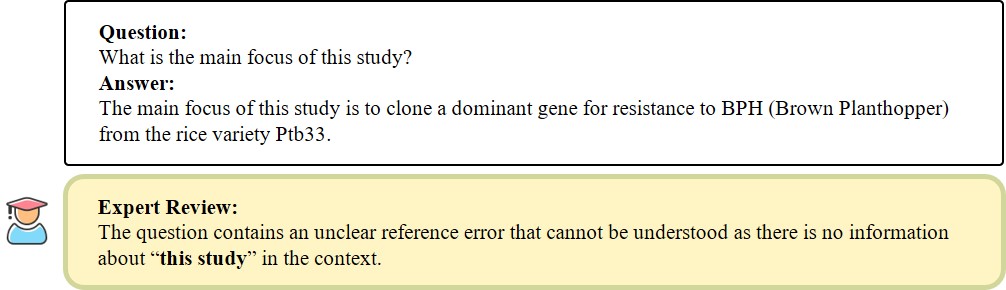}
\end{figure}

\begin{figure}[h!]
\centering
\includegraphics[width=0.95\textwidth]{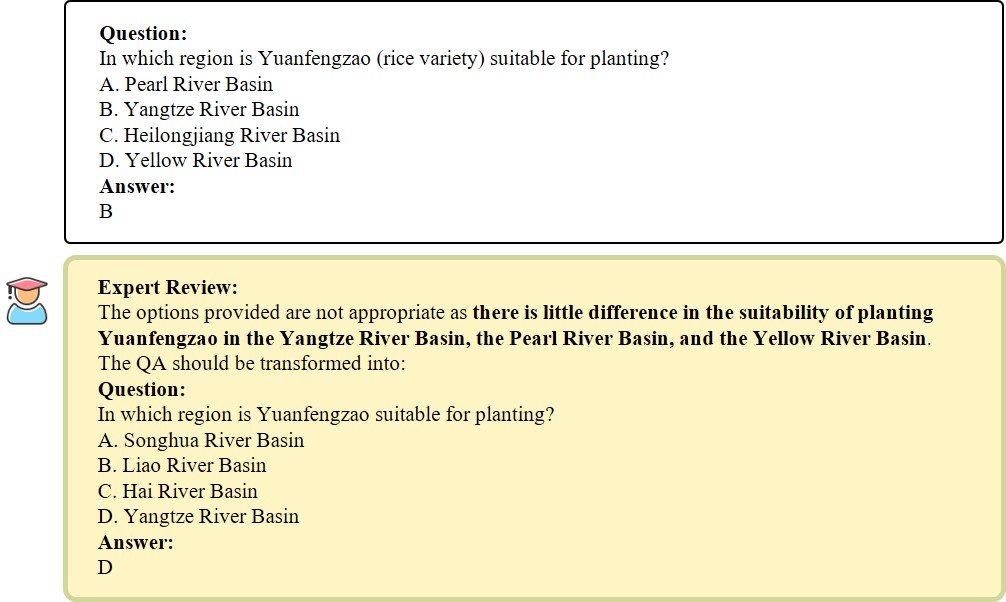}

\end{figure}

\newpage
\subsection{Descriptive Statistics}
\label{sec:C.2}
\begin{table}[h!]
\centering
\caption{Descriptive Statistics of Generated Questions and Answers}
\label{tab:table1}
\resizebox{\textwidth}{!}{%
\begin{tabular}{l r r r r r r r r r r r}
\hline
\textbf{} & \textbf{C1} & \textbf{C2} & \textbf{C3} & \textbf{C4} & \textbf{C5} & \textbf{C6} & \textbf{C7} & \textbf{C8} & \textbf{C9} & \textbf{C10} & \textbf{total} \\
\hline
QA-1 & 37 & 9 & 0 & 23 & 2 & 33 & 15 & 32 & 38 & 11 & 200 \\
     & (1.63\%) & (0.4\%) & (0.0\%) & (1.02\%) & (0.09\%) & (1.46\%) & (0.66\%) & (1.41\%) & (1.68\%) & (0.49\%) & (8.83\%) \\
\hline
QA-2 & 37 & 10 & 5 & 44 & 2 & 26 & 17 & 11 & 9 & 26 & 187 \\
     & (1.63\%) & (0.44\%) & (0.22\%) & (1.94\%) & (0.09\%) & (1.15\%) & (0.75\%) & (0.49\%) & (0.4\%) & (1.15\%) & (8.26\%) \\
\hline
QA-3 & 38 & 32 & 0 & 34 & 5 & 13 & 20 & 23 & 34 & 25 & 224 \\
     & (1.68\%) & (1.41\%) & (0.0\%) & (1.5\%) & (0.22\%) & (0.57\%) & (0.88\%) & (1.02\%) & (1.5\%) & (1.1\%) & (9.89\%) \\
\hline
QA-4 & 66 & 12 & 14 & 47 & 5 & 34 & 14 & 16 & 14 & 20 & 242 \\
     & (2.92\%) & (0.53\%) & (0.62\%) & (2.08\%) & (0.22\%) & (1.5\%) & (0.62\%) & (0.71\%) & (0.62\%) & (0.88\%) & (10.69\%) \\
\hline
SUM-1 & 39 & 17 & 6 & 36 & 4 & 25 & 20 & 26 & 28 & 24 & 225 \\
      & (1.72\%) & (0.75\%) & (0.27\%) & (1.59\%) & (0.18\%) & (1.1\%) & (0.88\%) & (1.15\%) & (1.24\%) & (1.06\%) & (9.94\%) \\
\hline
SUM-2 & 39 & 17 & 6 & 36 & 4 & 25 & 20 & 26 & 28 & 24 & 225 \\
      & (1.72\%) & (0.75\%) & (0.27\%) & (1.59\%) & (0.18\%) & (1.1\%) & (0.88\%) & (1.15\%) & (1.24\%) & (1.06\%) & (9.94\%) \\
\hline
RC-1 & 25 & 7 & 0 & 15 & 1 & 17 & 9 & 16 & 17 & 6 & 113 \\
     & (1.1\%) & (0.31\%) & (0.0\%) & (0.66\%) & (0.04\%) & (0.75\%) & (0.4\%) & (0.71\%) & (0.75\%) & (0.27\%) & (4.99\%) \\
\hline
RC-2 & 18 & 4 & 4 & 32 & 1 & 15 & 7 & 6 & 7 & 14 & 108 \\
     & (0.8\%) & (0.18\%) & (0.18\%) & (1.41\%) & (0.04\%) & (0.66\%) & (0.31\%) & (0.27\%) & (0.31\%) & (0.62\%) & (4.77\%) \\
\hline
RC-3 & 38 & 32 & 0 & 33 & 5 & 13 & 20 & 23 & 34 & 23 & 221 \\
     & (1.68\%) & (1.41\%) & (0.0\%) & (1.46\%) & (0.22\%) & (0.57\%) & (0.88\%) & (1.02\%) & (1.5\%) & (1.02\%) & (9.76\%) \\
\hline
RC-4 & 65 & 12 & 14 & 47 & 5 & 33 & 14 & 16 & 14 & 20 & 240 \\
     & (2.87\%) & (0.53\%) & (0.62\%) & (2.08\%) & (0.22\%) & (1.46\%) & (0.62\%) & (0.71\%) & (0.62\%) & (0.88\%) & (10.6\%) \\
\hline
RC-5 & 58 & 26 & 24 & 43 & 4 & 26 & 20 & 26 & 28 & 24 & 279 \\
     & (2.56\%) & (1.15\%) & (1.06\%) & (1.9\%) & (0.18\%) & (1.15\%) & (0.88\%) & (1.15\%) & (1.24\%) & (1.06\%) & (12.32\%) \\
\hline
\textbf{total} & 460 & 178 & 73 & 390 & 38 & 260 & 176 & 221 & 251 & 217 & 2264 \\
          & (20.32\%) & (7.86\%) & (3.22\%) & (17.23\%) & (1.68\%) & (11.48\%) & (7.77\%) & (9.76\%) & (11.09\%) & (9.58\%) & (100.0\%) \\
\hline
\end{tabular}%
}
\end{table}

\newpage
\section{Additional Experimental Setup}
\label{sec:D}
\subsection{Model Hyperparameters}
\label{sec:D.1}
We conduct evaluations on different large language models using \texttt{opencompass} as the primary tool. In order to ensure reproducibility and provide a reference for future research, the main hyperparameters used in our experiments, along with their meanings, are listed in Table~\ref{tab:hyperparameters}:

\begin{table}[htbp!]
\centering
\caption{Main Hyperparameters for LLM Evaluation}
\label{tab:hyperparameters}
\begin{tabular}{p{3.5cm} p{9cm} p{2cm}}
\hline
\textbf{Parameter} & \textbf{Meaning} & \textbf{Value} \\
\hline
\texttt{max\_seq\_len} & The maximum context length (upper limit of input tokens) & 7168 \\
\texttt{max\_out\_len} & The maximum output length (upper limit of generated tokens) & 2048 \\
\texttt{batch\_size} & The batch size (number of requests processed per generation) & 80 \\
\hline
\end{tabular}
\end{table}

For proprietary LLMs, the maximum generation tokens for each model were set to 2048, with a batch size of 80. The temperature was set to 0.7, and top-$p$ and top-$k$ were set to 0.8 and 10, respectively.  For open-source LLMs, the maximum generation tokens were also set to 2048, with a batch size of 80. The temperature was set to 0.7, and top-$p$ and top-$k$ were set to 0.8 and 10, respectively. 

\subsection{Evaluation Metrics}
\label{sec:D.2}
The evaluation process consists of two steps: the first step is answer extraction. After collecting the model's responses, we first perform post-processing to extract the model's replies. The second step is metric calculation. During evaluation, we have designed different evaluation methods for different question types:

\textbf{Single-choice Questions.}
We use Accuracy as the evaluation metric to calculate the correctness of the model's answers. The formula for Accuracy is:

\[
\text{Accuracy} = \frac{1}{N_{\text{total}}} \sum_{i=1}^{N_{\text{total}}} \delta(y_i^{\text{pred}}, y_i^{\text{true}})
\]

where \(N_{\text{total}}\) is the total number of questions, \(y_i^{\text{pred}}\) is the predicted answer for the \(i\)-th question by the model, and \(y_i^{\text{true}}\) is the true answer for the \(i\)-th question. \(\delta(a, b)\) is the Kronecker delta function, where \(\delta(a, b) = 1\) if \(a = b\), and \(\delta(a, b) = 0\) otherwise.

\textbf{Multiple-choice Questions.}
We use the \textbf{F1 score}, which balances the correct and incorrect answers from the model to provide a more comprehensive reflection of model performance. The formula for the F1 score is:

\[
F1 = 2 \cdot \frac{P \cdot R}{P + R}
\]

where precision (\(P\)) is:

\[
P = \frac{1}{N} \sum_{i=1}^{N} \frac{|Y_i^{\text{pred}} \cap Y_i^{\text{true}}|}{|Y_i^{\text{pred}}|}
\]

and recall (\(R\)) is:

\[
R = \frac{1}{N} \sum_{i=1}^{N} \frac{|Y_i^{\text{pred}} \cap Y_i^{\text{true}}|}{|Y_i^{\text{true}}|}
\]

where \(N\) is the total number of samples, \(Y_i^{\text{pred}}\) is the set of predicted answers for the \(i\)-th sample, and \(Y_i^{\text{true}}\) is the set of true answers for the \(i\)-th sample.

\textbf{Fill-in-the-Blank and Generation Questions.}
We use ROUGE for evaluation and calculate the average F1 score of ROUGE-L. For fill-in-the-blank questions, since answers typically contain multiple blanks or paragraphs, we adopt a segmented evaluation method, comparing the similarity of each segment between the model's predicted answer and the reference answer, and averaging the scores of multiple segments to more accurately reflect the model's performance for each answer item. For generation questions, we use a sentence-level evaluation method, directly calculating the similarity between the model's generated full answer and the reference answer. This method allows for a comprehensive evaluation of the overall fluency and semantic accuracy of the generated content.

The formula for ROUGE-L F1 is:

\[
\text{ROUGE-L} = \frac{(1 + \beta^2) \cdot \text{P} \cdot \text{R}}{\beta^2 \cdot \text{P} + \text{R}}
\]

where precision (\(P\)) is:

\[
P = \frac{1}{N} \sum_{i=1}^{N} \frac{\text{LCS}(X_i, Y_i)}{|X_i|}
\]

and recall (\(R\)) is:

\[
R = \frac{1}{N} \sum_{i=1}^{N} \frac{\text{LCS}(X_i, Y_i)}{|Y_i|}
\]

where \(N\) is the total number of samples, \(X_i\) and \(Y_i\) represent the predicted and reference answers for the \(i\)-th sample, and \(\text{LCS}(X_i, Y_i)\) is the length of the longest common subsequence between \(X_i\) and \(Y_i\). \(|X_i|\) and \(|Y_i|\) are the lengths of \(X_i\) and \(Y_i\), respectively. \(\beta\) is the balance factor between precision and recall, typically set to 1.

\newpage
\section{Error Analysis of Large Language Model Outputs}
\label{sec:E}

In this chapter, we thoroughly analyze the errors produced by large language models in different scenarios, according to the three task types defined above. By presenting typical cases and discussing both the causes and potential improvements, we provide a reference for optimizing the quality of LLM outputs in subsequent work.

\subsection{Errors in the First Task Type}
\label{sec:E.1}
The first task type primarily involves gene information retrieval capabilities. The model needs to demonstrate fundamental biological knowledge as well as accurate information retrieval. We observe several representative error cases, outlined below:

\paragraph{Gene Name Confusion}
When dealing with similar or homonymous gene names or gene identifiers, the model incorrectly provides the wrong gene sequences and functional descriptions. This mistake can lead to misalignment in subsequent gene function validation.
\begin{figure}[htbp]
\centering
\includegraphics[width=0.95\textwidth]{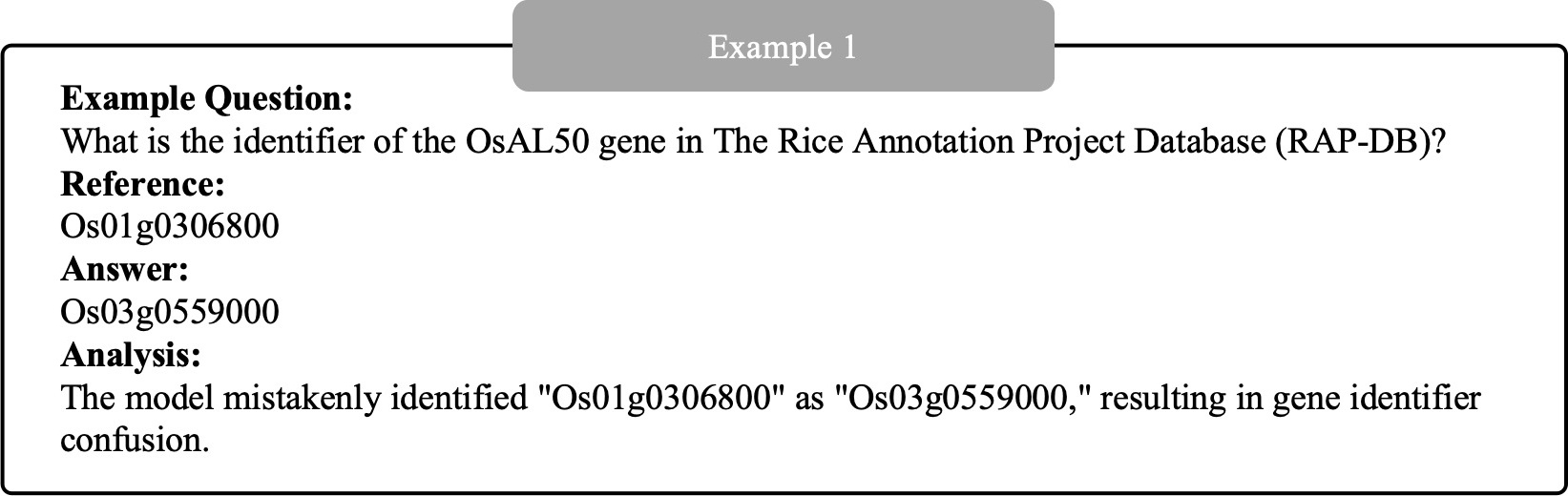}
\includegraphics[width=0.95\textwidth]{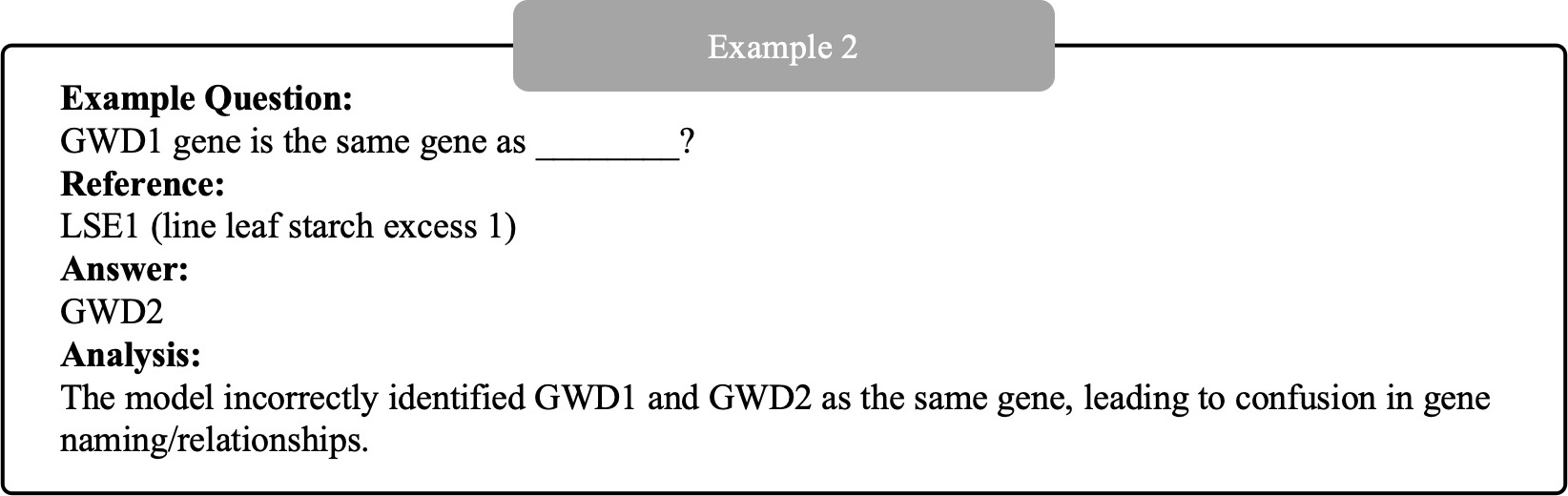}
\includegraphics[width=0.95\textwidth]{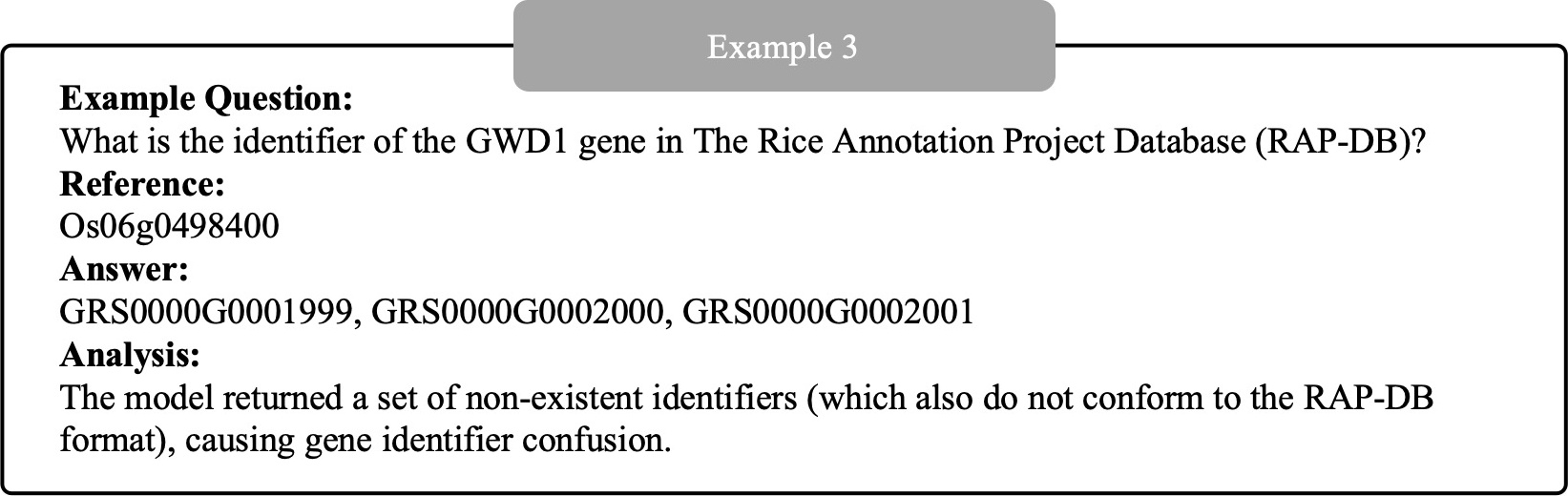}
\end{figure}

\begin{figure}[htbp]
\centering
\includegraphics[width=0.95\textwidth]{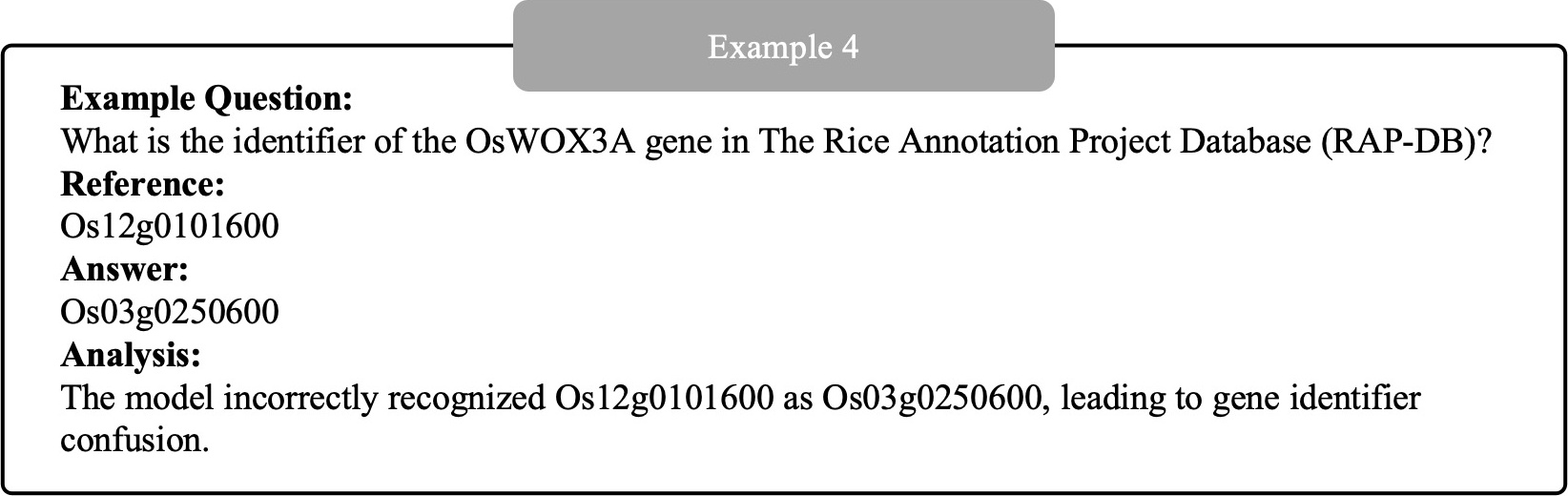}
\includegraphics[width=0.95\textwidth]{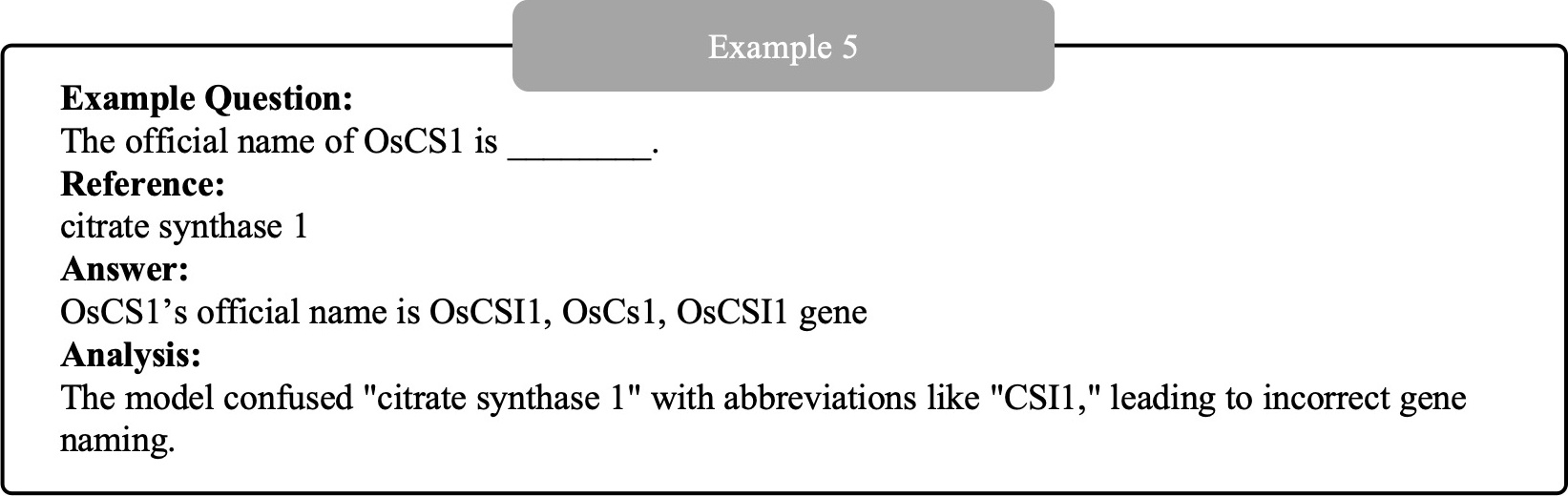}
\includegraphics[width=0.95\textwidth]{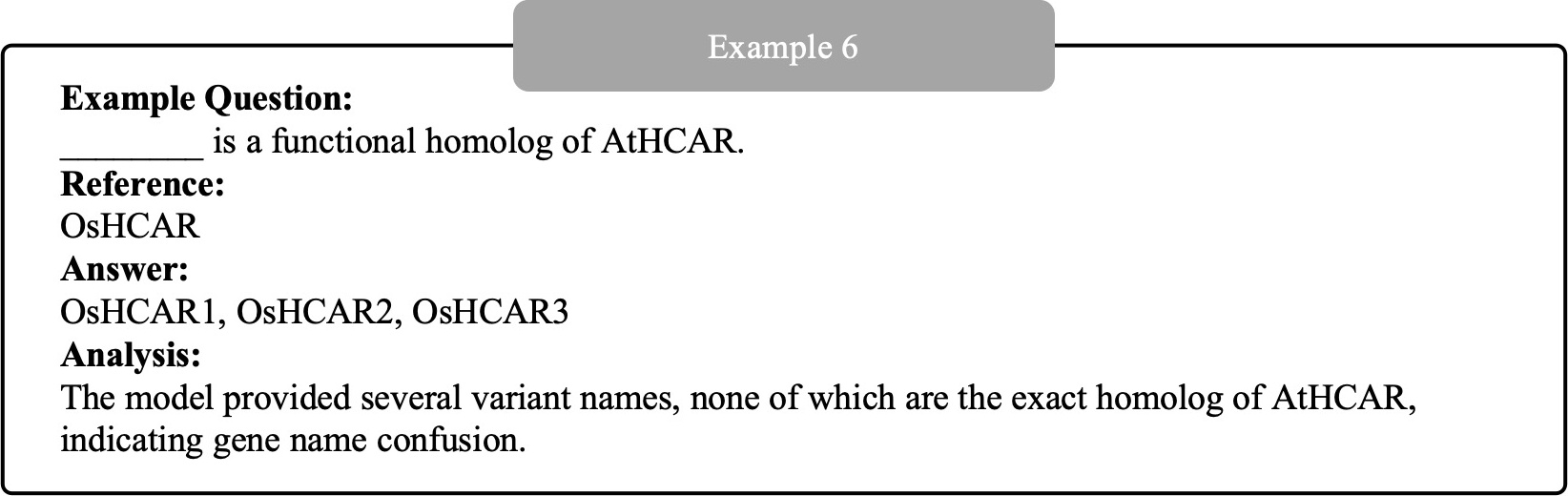}
\includegraphics[width=0.95\textwidth]{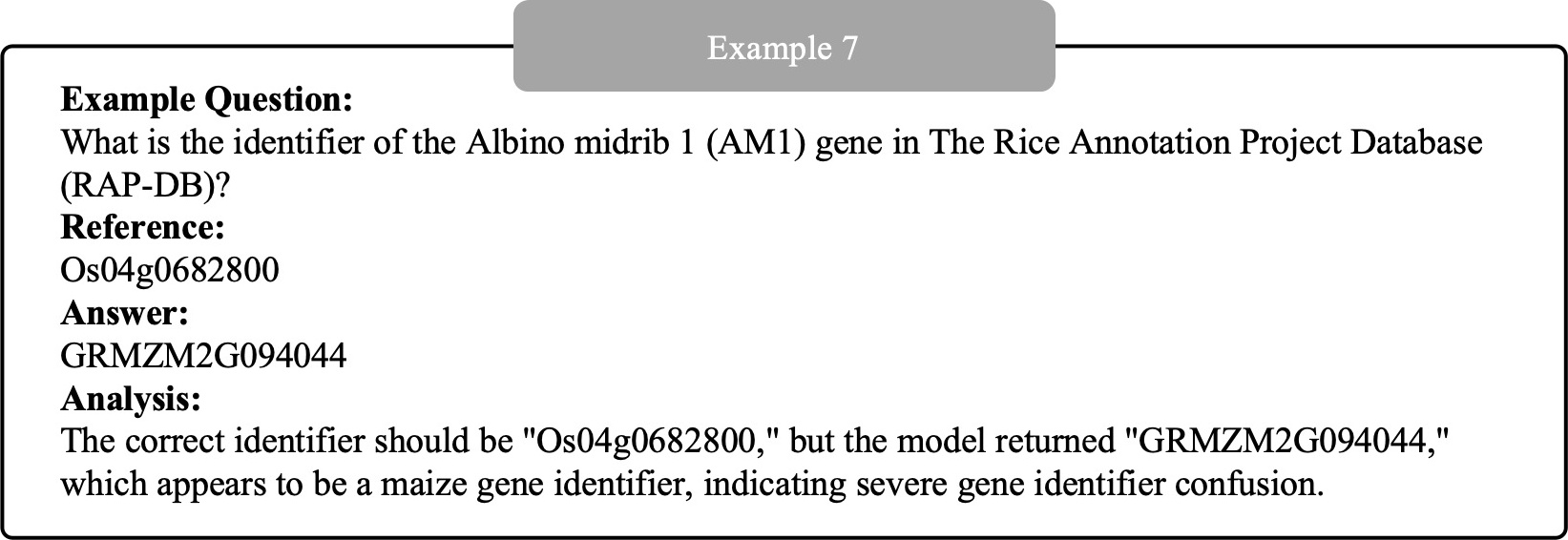}
\caption{Example illustrating gene name confusion.}
\label{fig:gene-name-confusion}
\end{figure}

\newpage
\paragraph{Errors in Gene Sequence and Positional Information}
The model provides incorrect answers regarding the physical location information of genes (chromosome number, start and end coordinates, sequence length, etc.), such as incorrect gene coordinates or sequence length discrepancies with actual databases. This may impact genome annotation and structural variation analysis, leading to experimental design biases.

\begin{figure}[htbp!]
\centering
\includegraphics[width=0.95\textwidth]{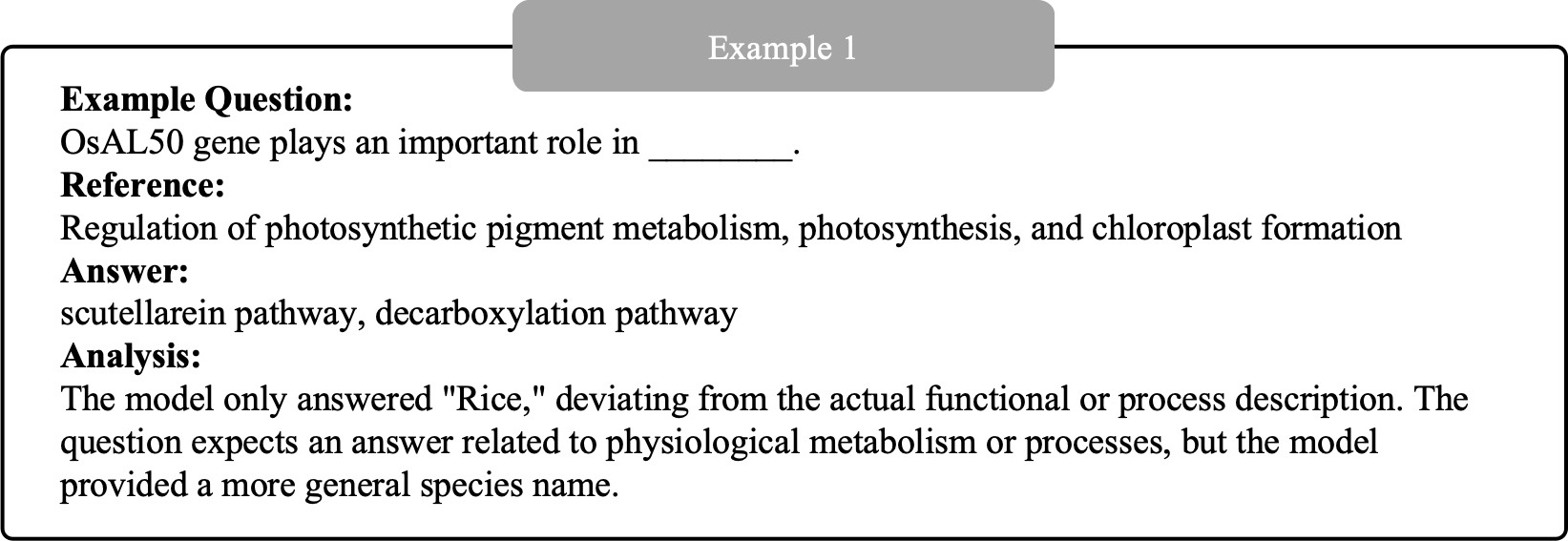}
\includegraphics[width=0.95\textwidth]{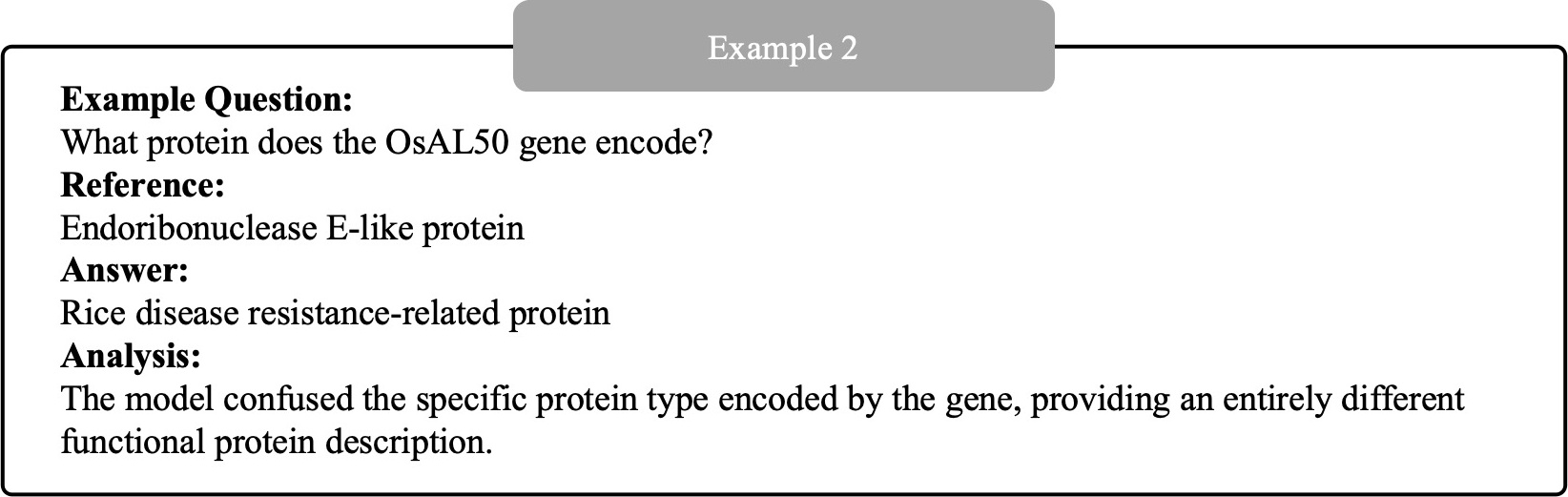}
\includegraphics[width=0.95\textwidth]{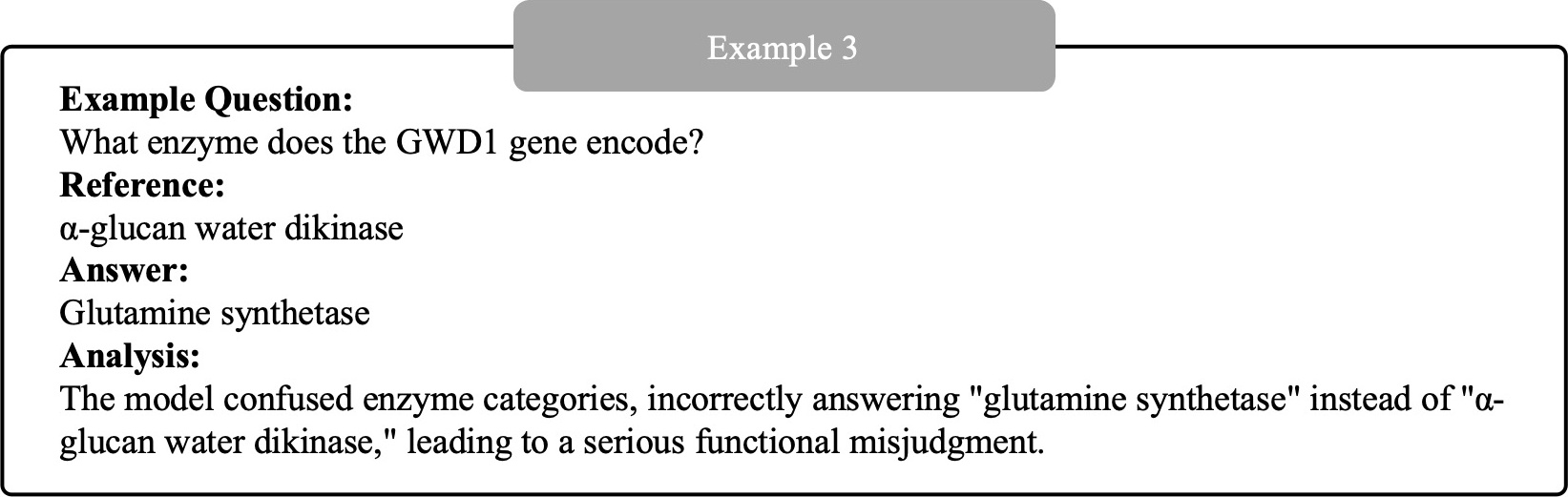}
\includegraphics[width=0.95\textwidth]{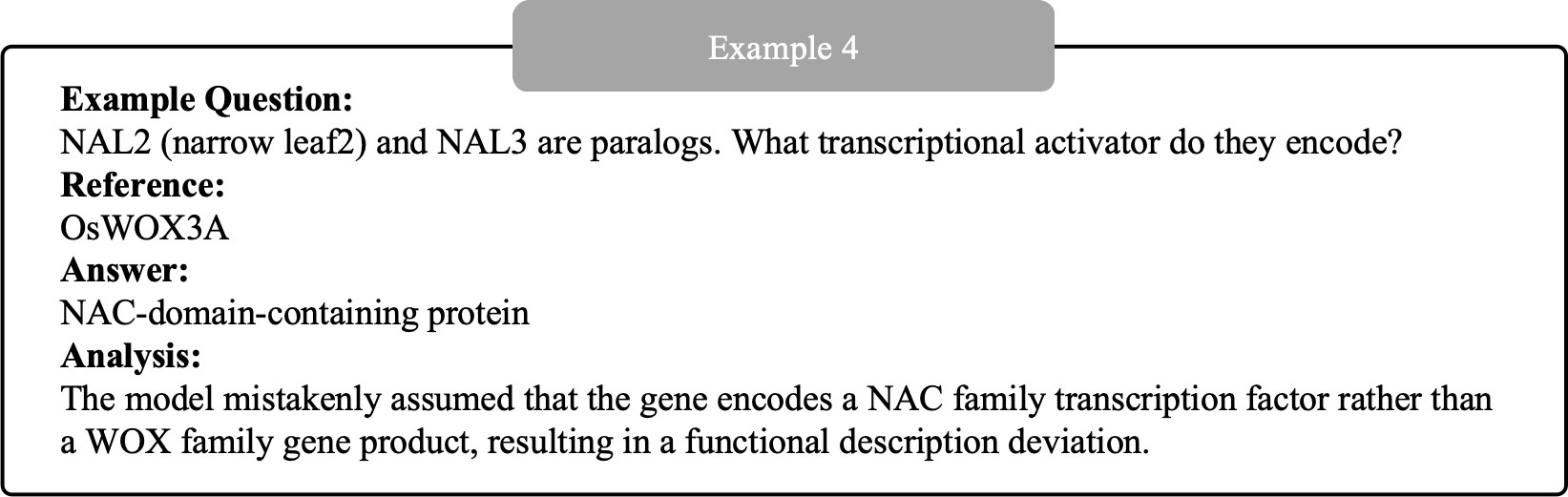}
\end{figure}

\newpage
\begin{figure}[htbp]
\centering
\includegraphics[width=0.95\textwidth]{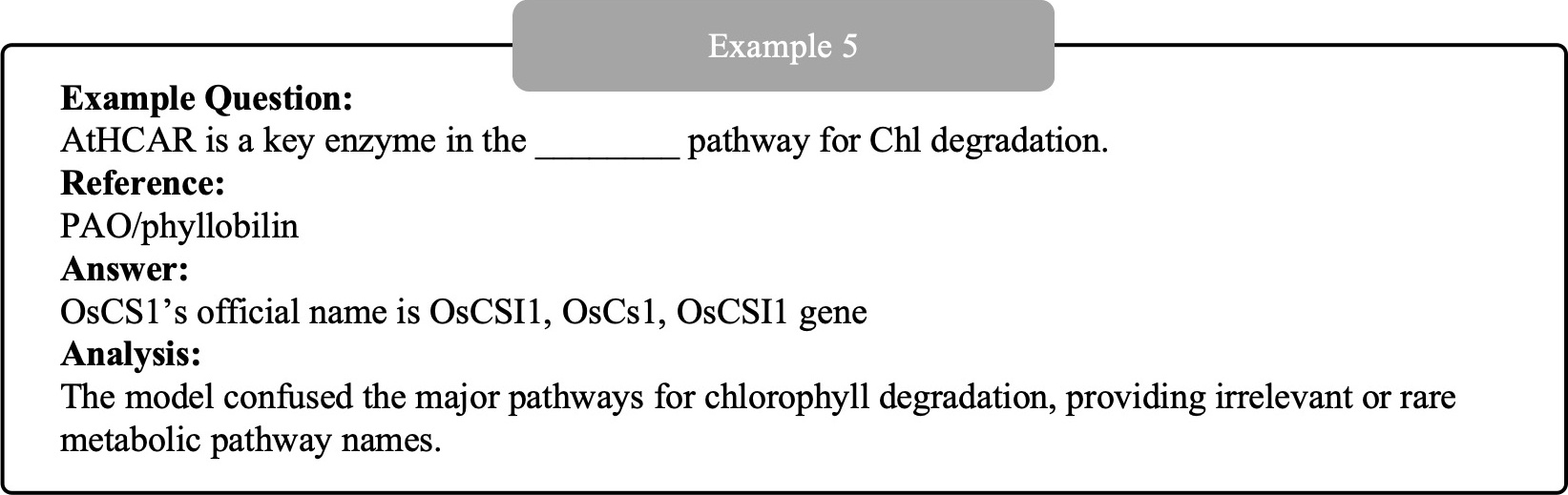}
\includegraphics[width=0.95\textwidth]{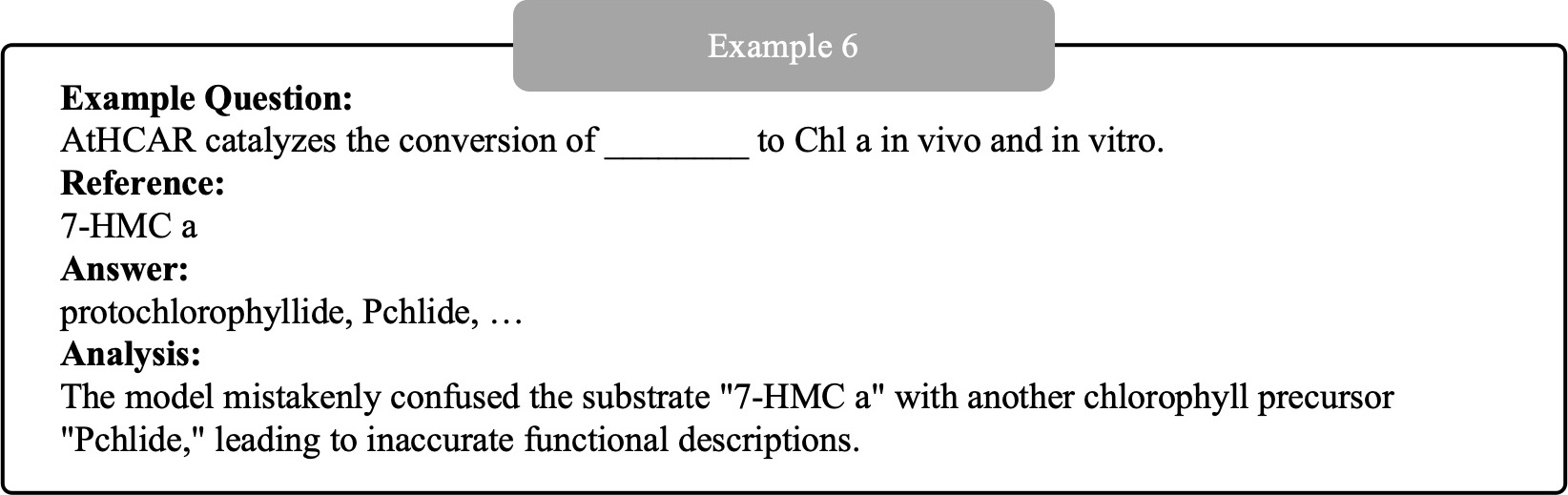}
\includegraphics[width=0.95\textwidth]{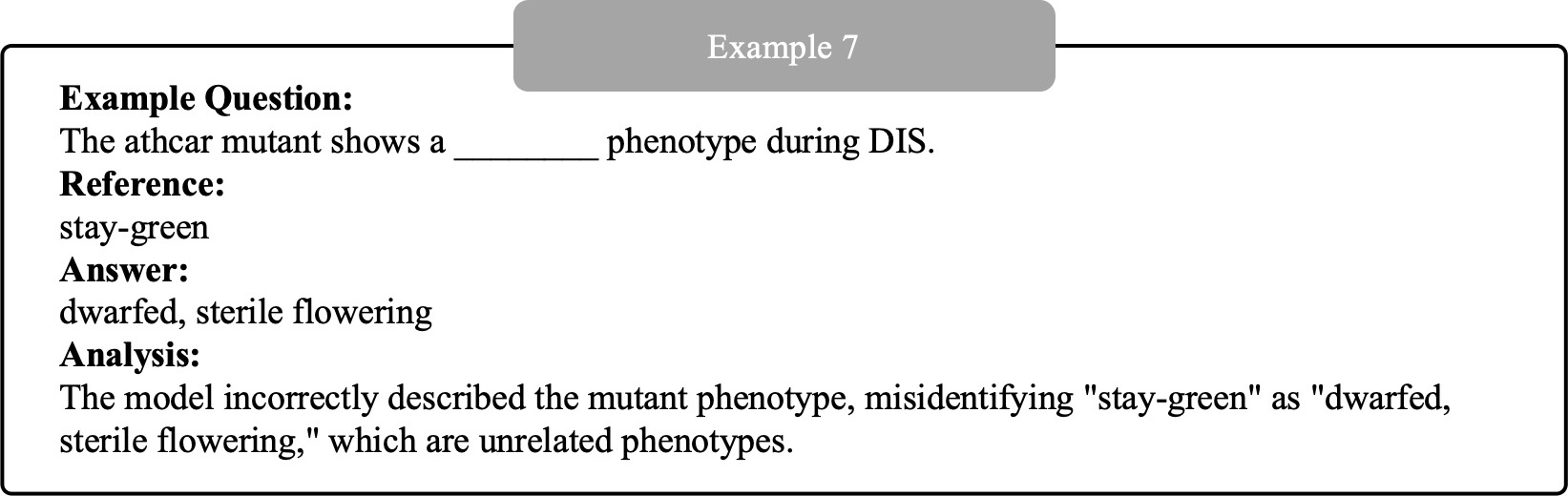}
\caption{Example illustrating cross-species gene information errors.}
\label{fig:cross-species-info-1}
\end{figure}

\newpage
\paragraph{Errors in Gene Function and Regulation}
The model exhibits errors in describing gene functions, protein products, or regulatory mechanisms, such as misclassifying gene functions, misunderstanding their roles in signaling pathways, or incorrectly predicting protein products. These issues may mislead researchers' understanding of gene biological functions, potentially leading to incorrect downstream experiments and data analysis.

\begin{figure}[htbp!]
\centering
\includegraphics[width=0.95\textwidth]{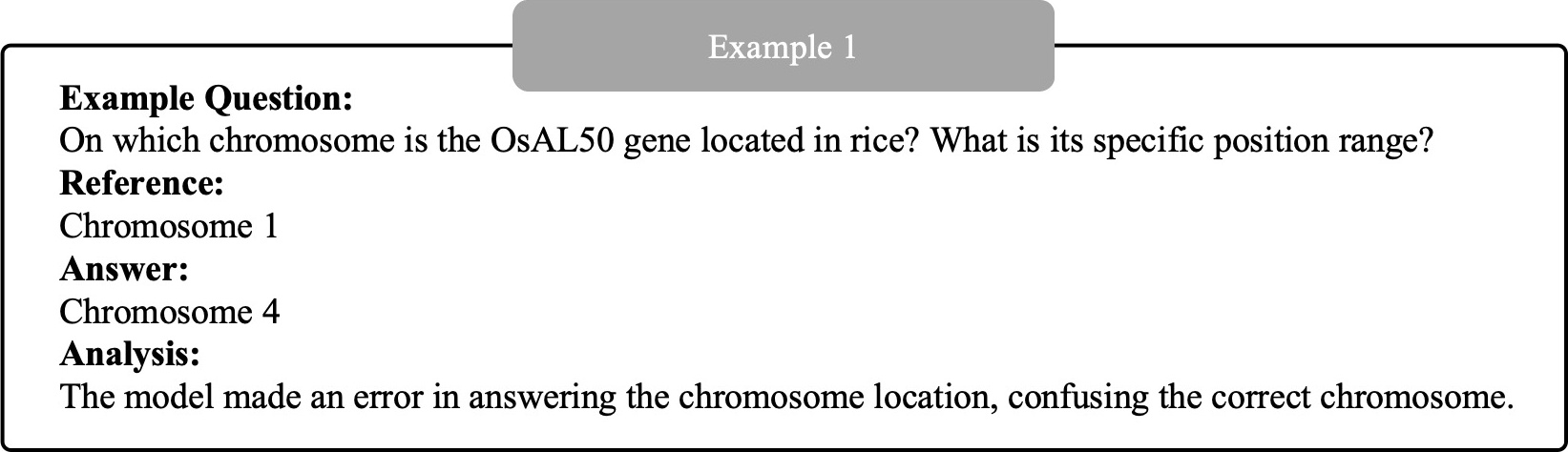}
\includegraphics[width=0.95\textwidth]{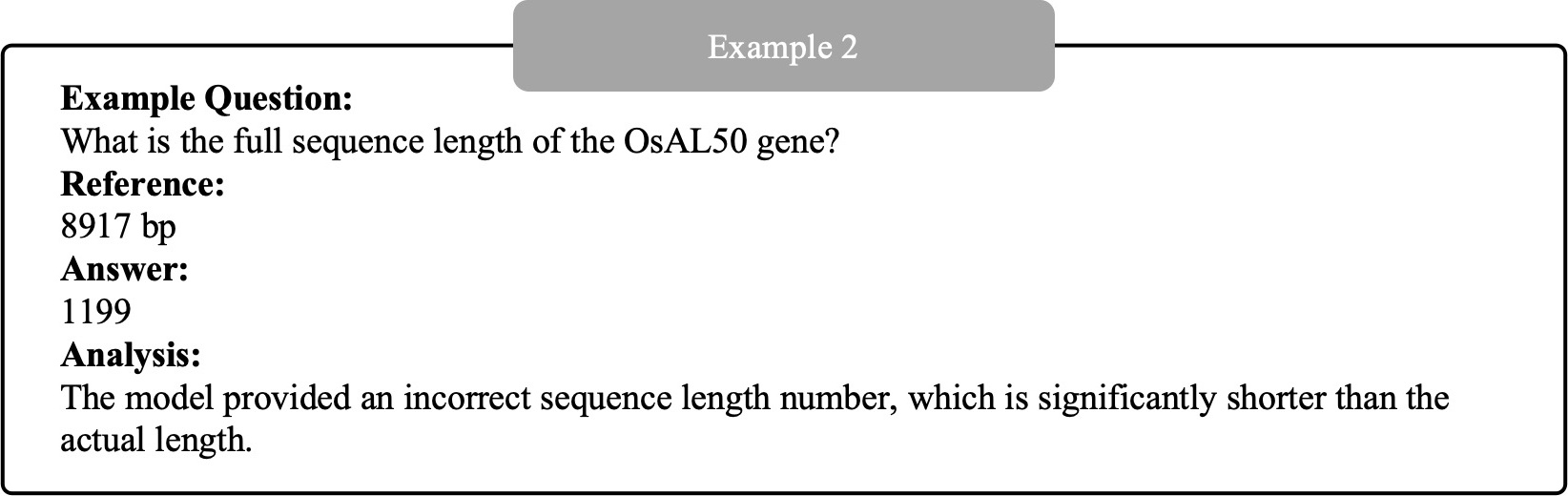}
\includegraphics[width=0.95\textwidth]{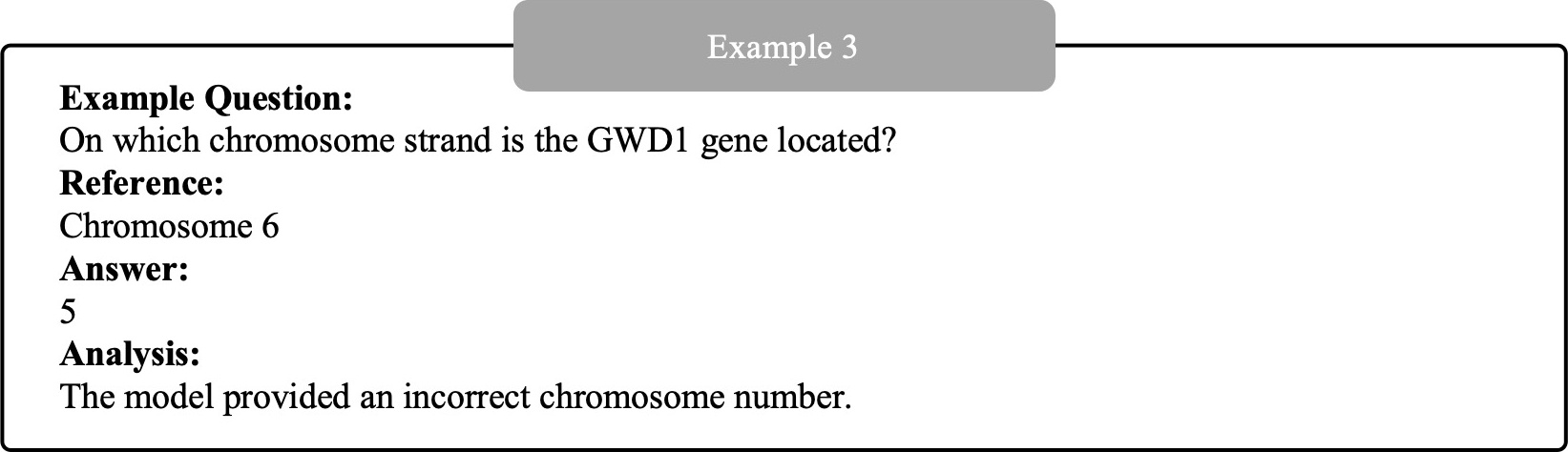}
\includegraphics[width=0.95\textwidth]{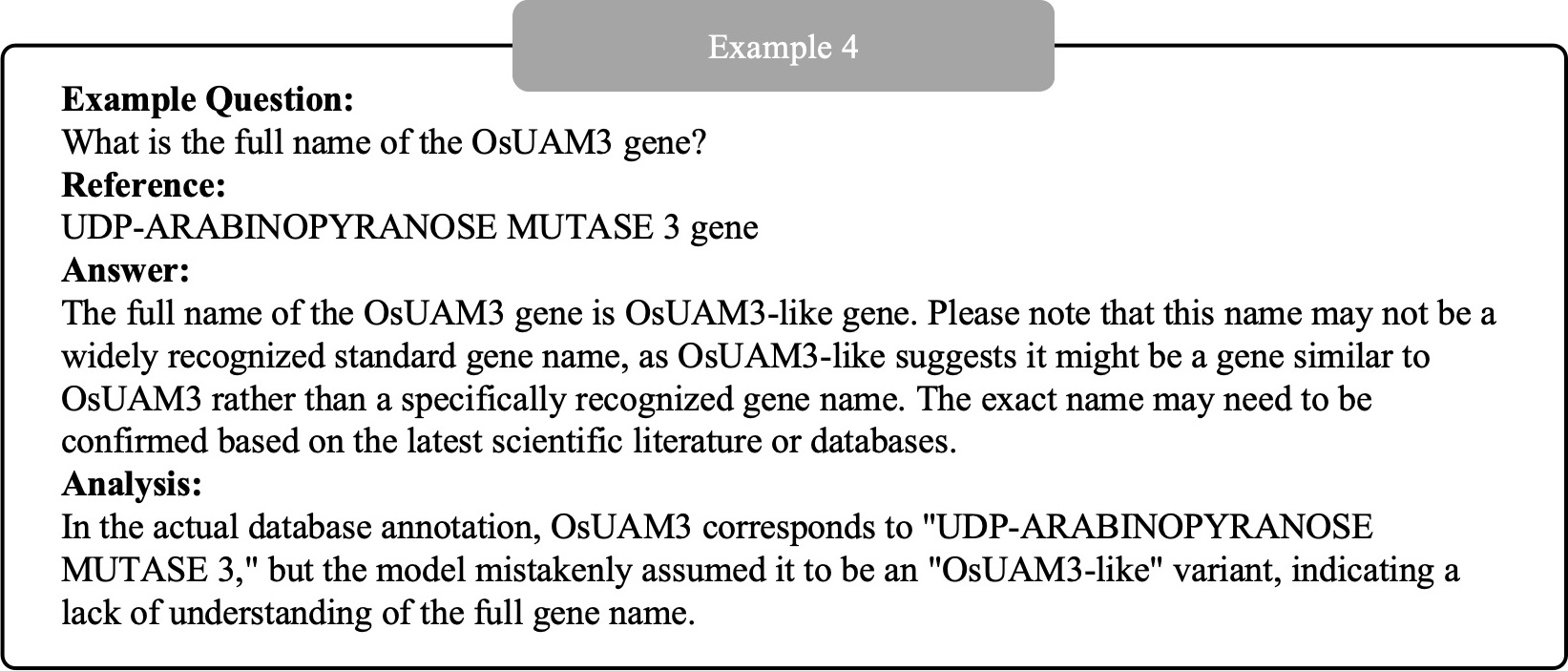}
\end{figure}

\newpage
\begin{figure}[htbp]
\centering
\includegraphics[width=0.95\textwidth]{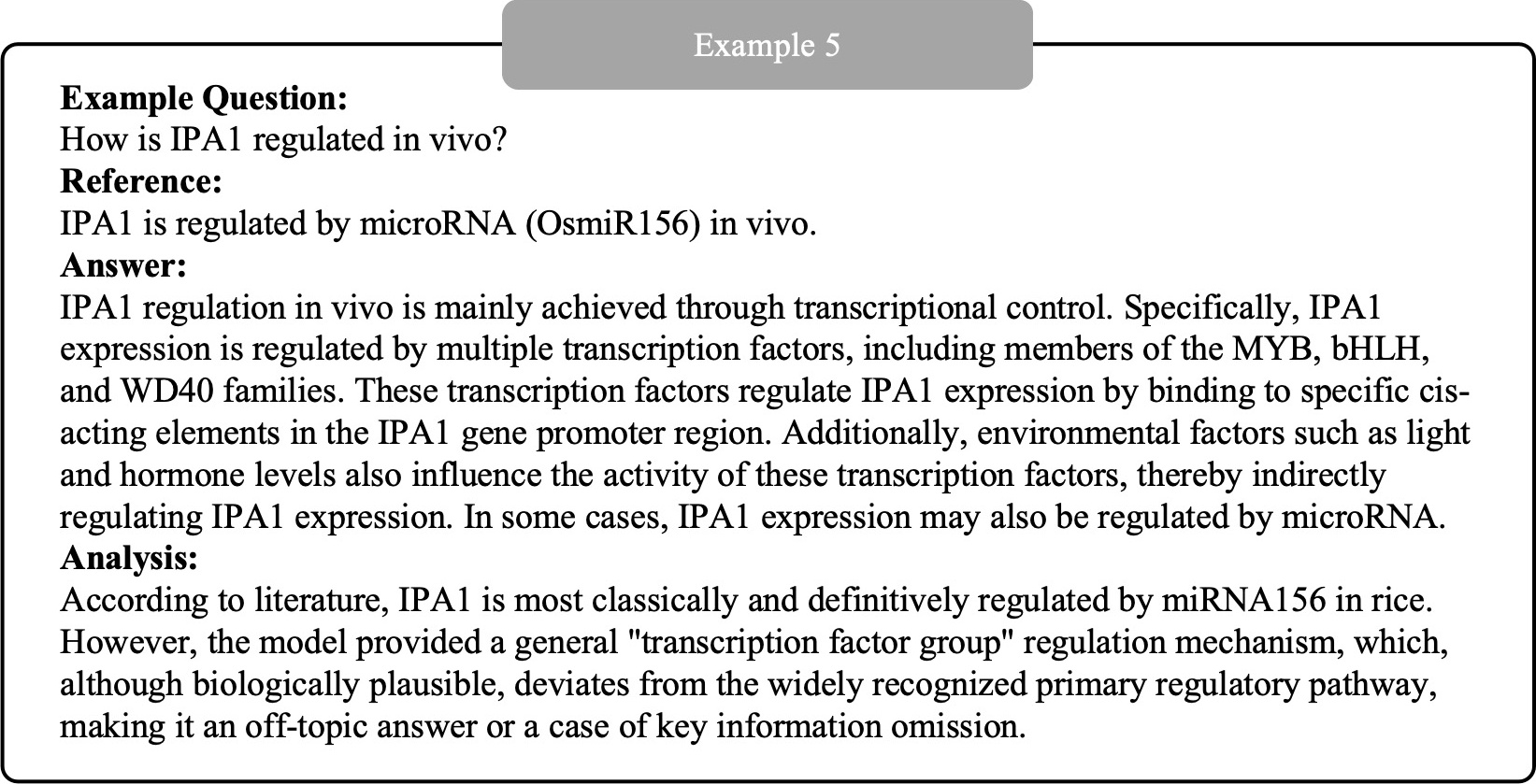}
\caption{Examples of gene function and regulatory errors.}
\label{fig:cross-species-info-2}
\end{figure}

\newpage
\subsection{Errors in the Second Task Type}
\label{sec:E.2}
The second task type concerns gene function and regulatory mechanisms, requiring more advanced logical reasoning and deeper domain knowledge. The model is expected to demonstrate “causal understanding” or “upstream/downstream interactions” thinking in gene function validation and regulatory network analysis. Similarly, we observe several representative error types, listed below:

\paragraph{Knowledge Errors}
Because the model lacks thorough understanding of the background or domain-specific knowledge, its answers can be incorrect. In Table~X (not shown), we list erroneous examples where the model misinterprets domain-specific terminology or experimental data, resulting in inaccurate responses.
\begin{figure}[htbp!]
\centering
\includegraphics[width=0.95\textwidth]{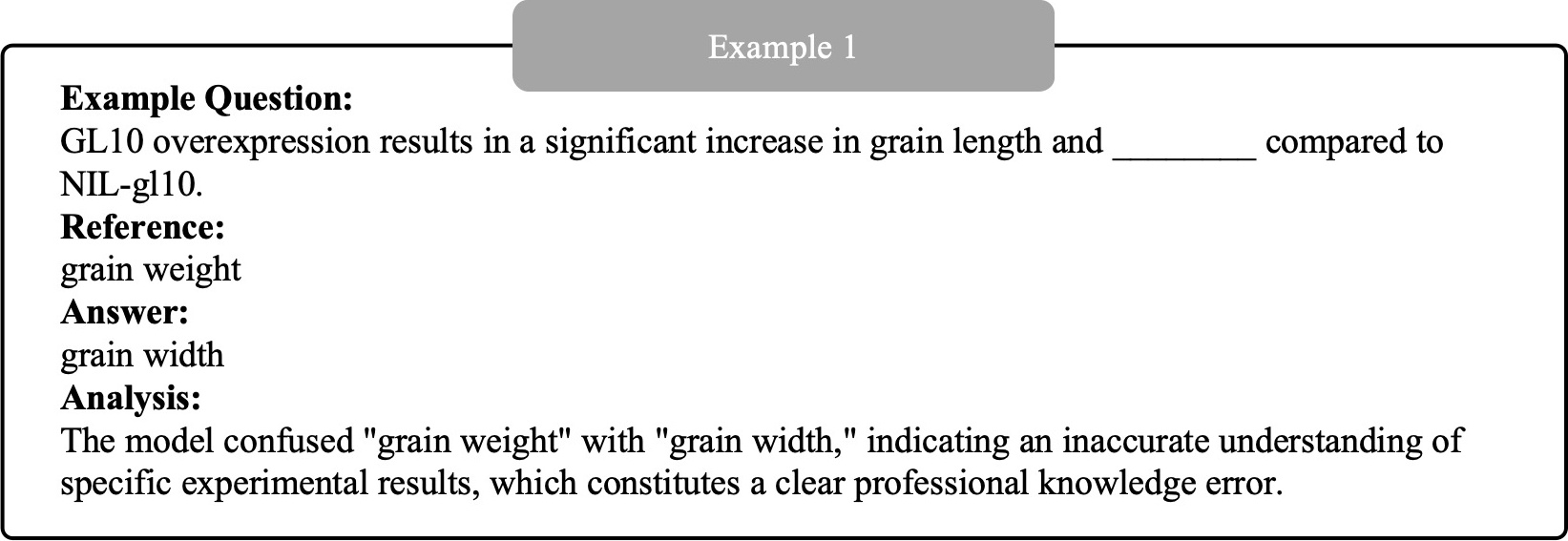}
\includegraphics[width=0.95\textwidth]{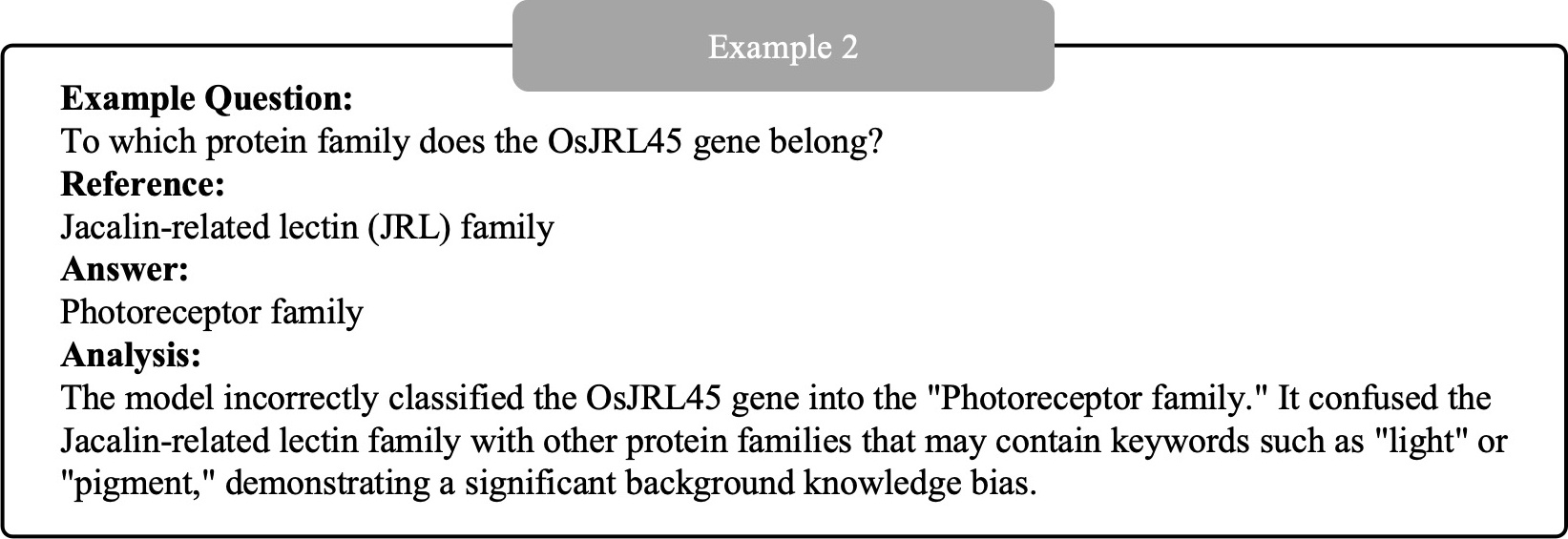}
\includegraphics[width=0.95\textwidth]{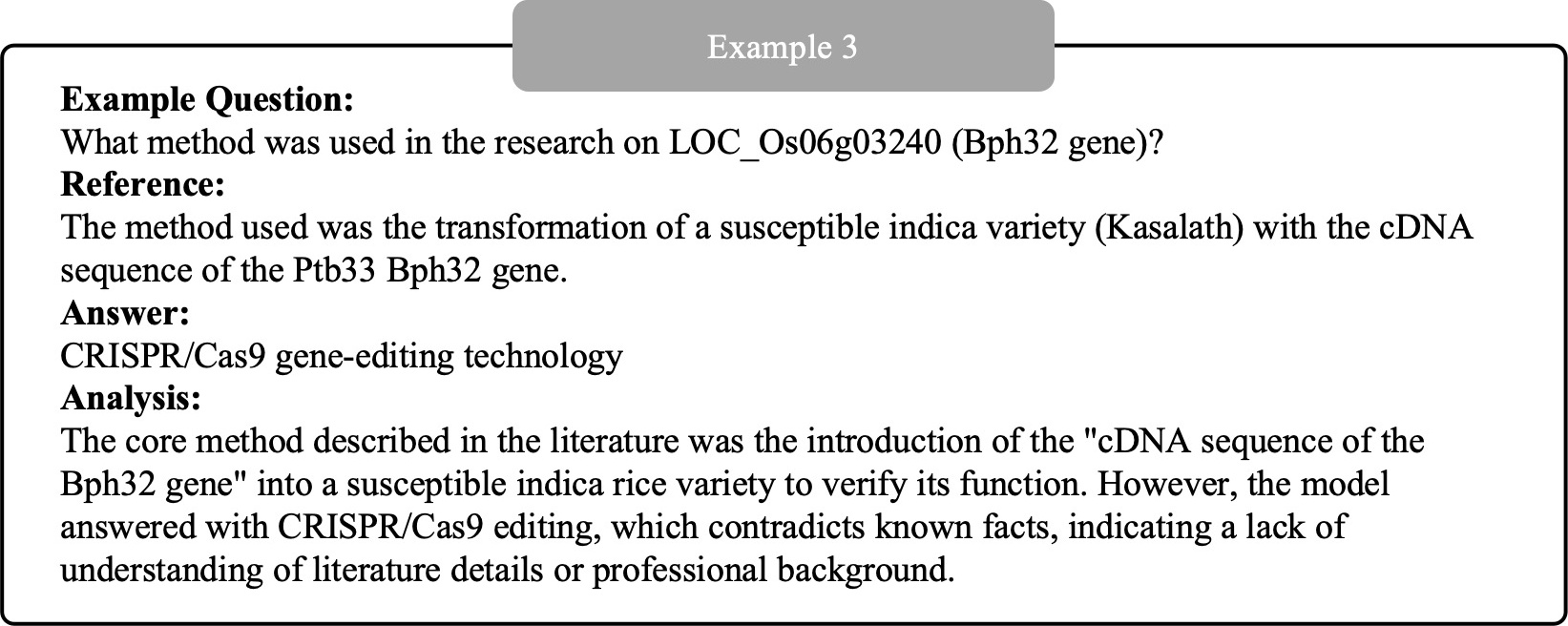}
\end{figure}

\newpage
\begin{figure}[htbp!]
\centering
\includegraphics[width=0.95\textwidth]{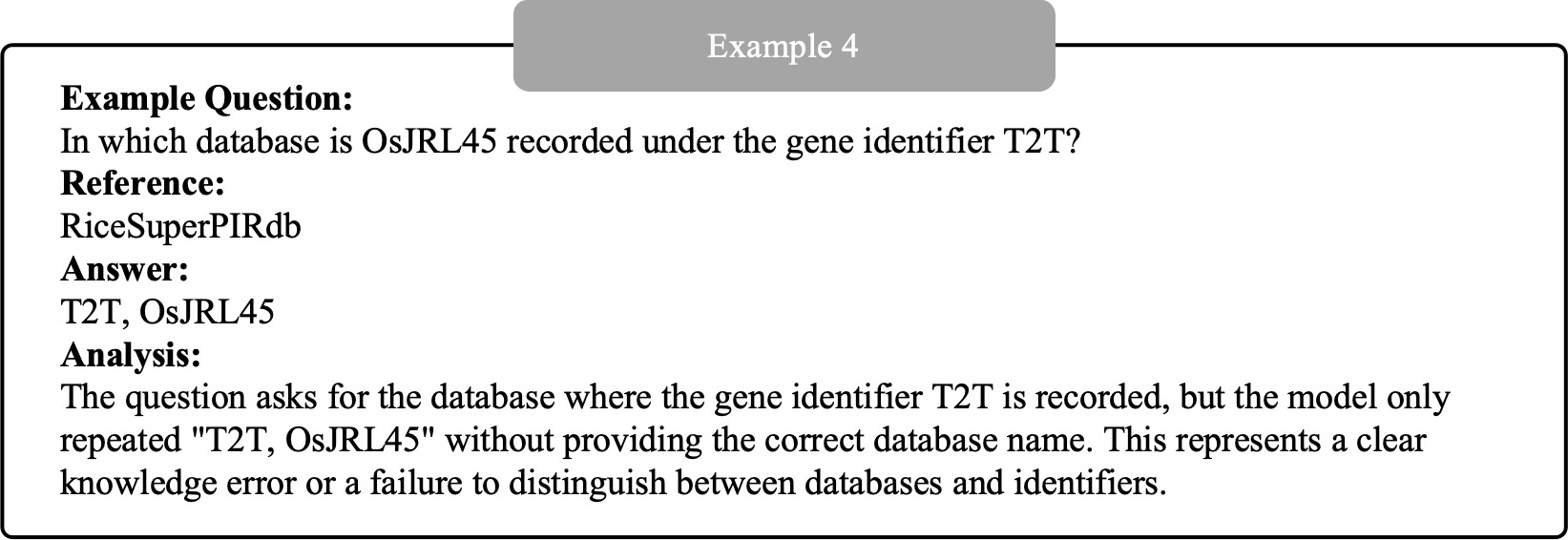}
\includegraphics[width=0.95\textwidth]{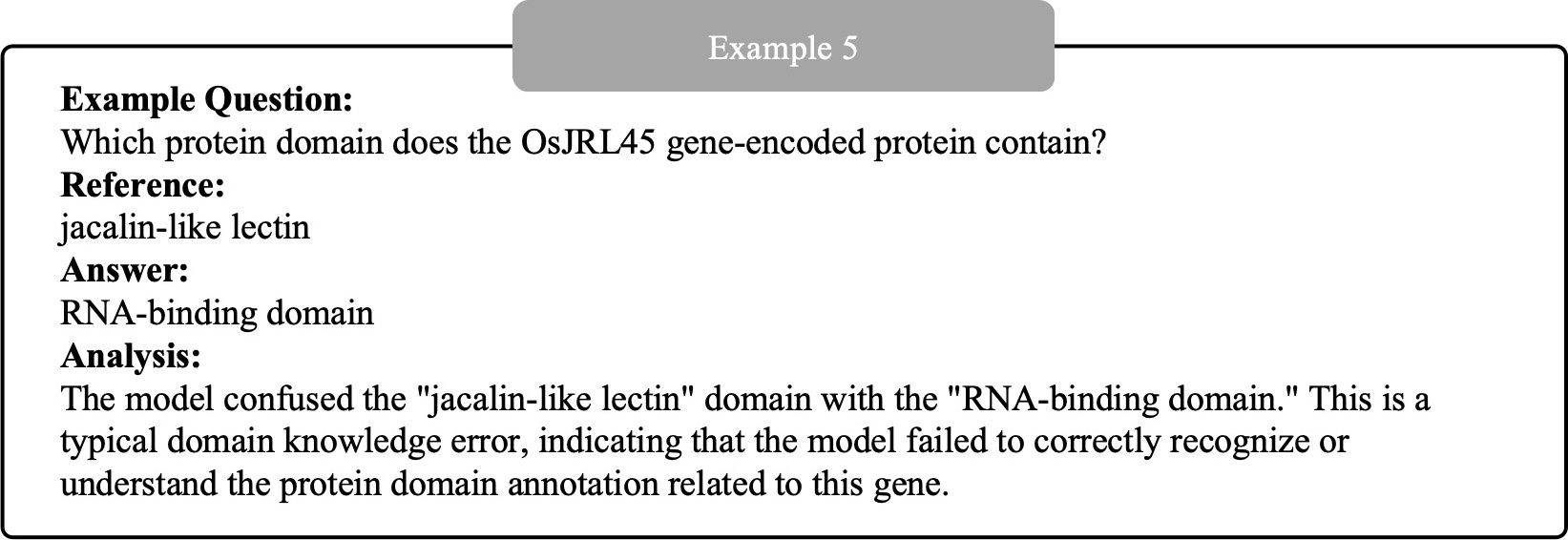}
\caption{Example illustrating knowledge errors in gene function and regulation.}
\label{fig:knowledge-errors}
\end{figure}

\newpage
\paragraph{Insufficient Answer Precision}
When the question requires detailed specifics, such as particular experimental methods, tools, or procedures, the model fails to provide precise information and instead offers overly generic descriptions. As a result, the answer does not meet the exact requirements of the question, indicating over-generalization.


\begin{figure}[htbp!]
\centering
\includegraphics[width=0.95\textwidth]{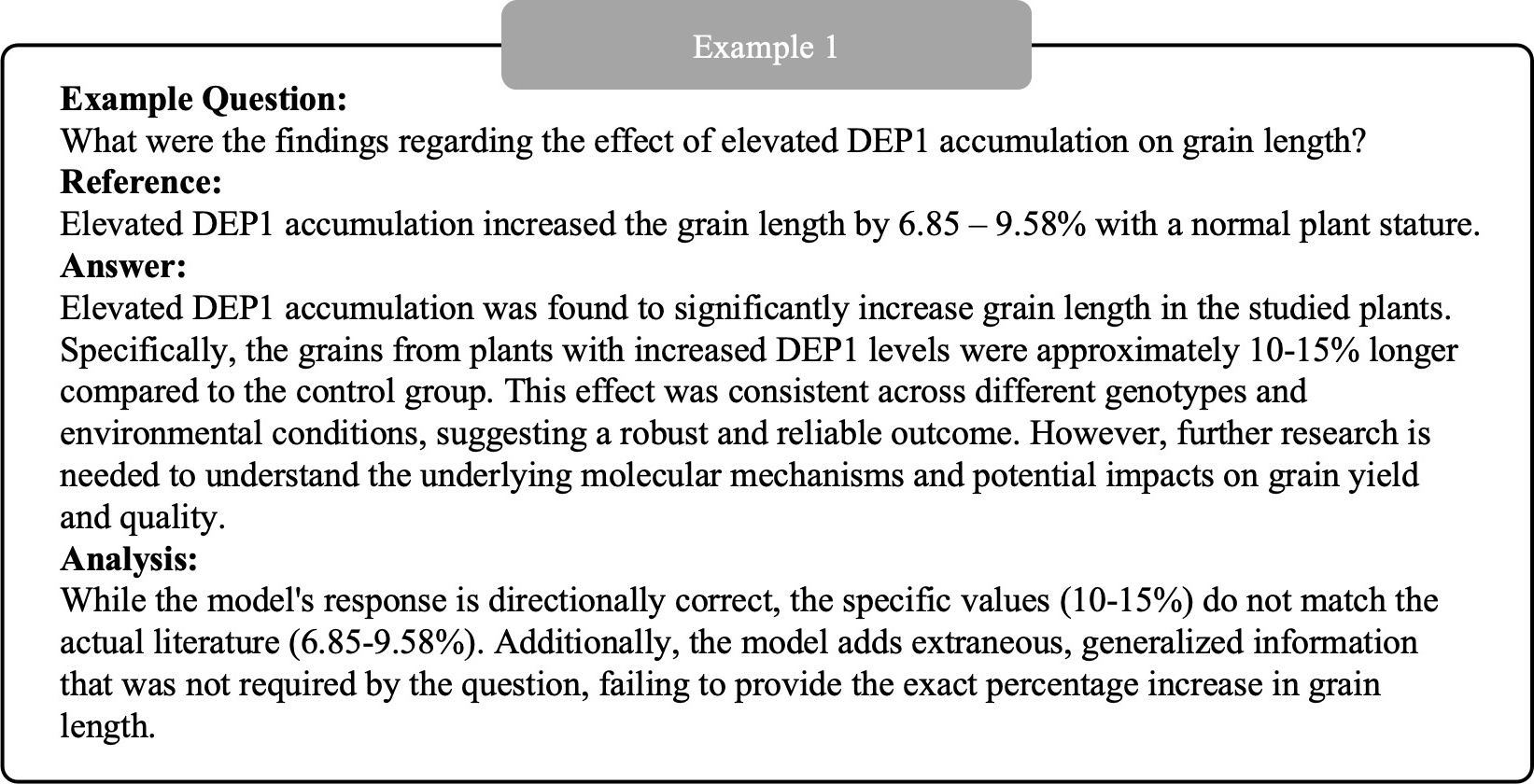}
\includegraphics[width=0.95\textwidth]{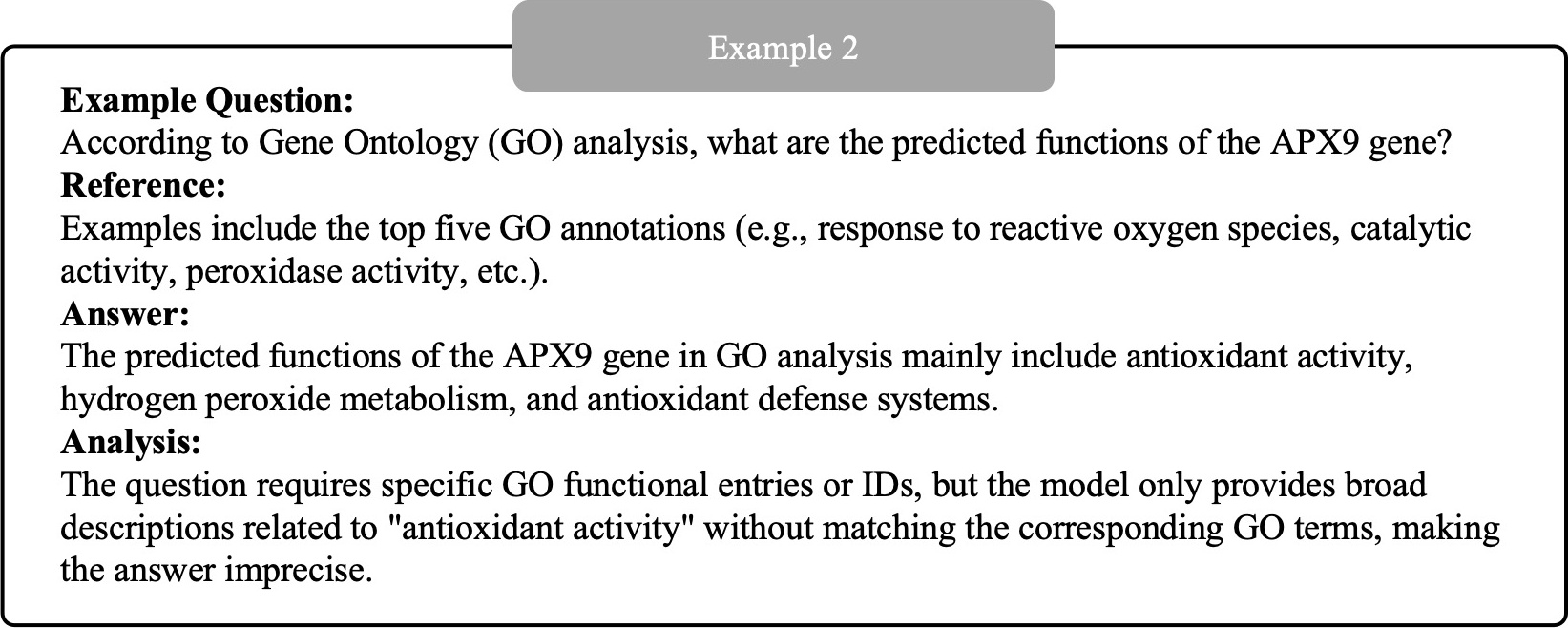}
\includegraphics[width=0.95\textwidth]{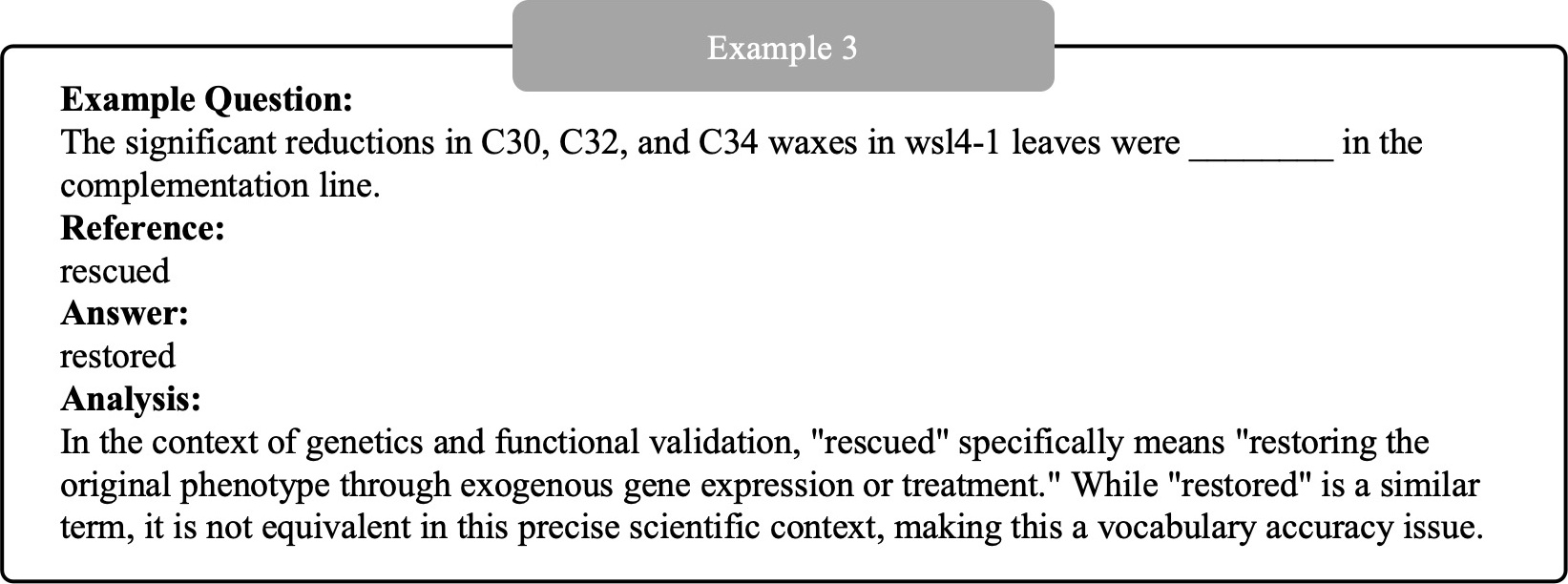}
\caption{Example illustrating lack of precision in model answers.}
\label{fig:precision-insufficiency}
\end{figure}

\newpage
\paragraph{Limited Reasoning Ability}
The model does not incorporate the given information to perform logical analysis and contextual interpretation, thereby producing responses that do not meet practical needs. It either fails to link the contextual clues correctly or ignores implicit details in the question.
\begin{figure}[htbp]
\centering
\includegraphics[width=0.95\textwidth]{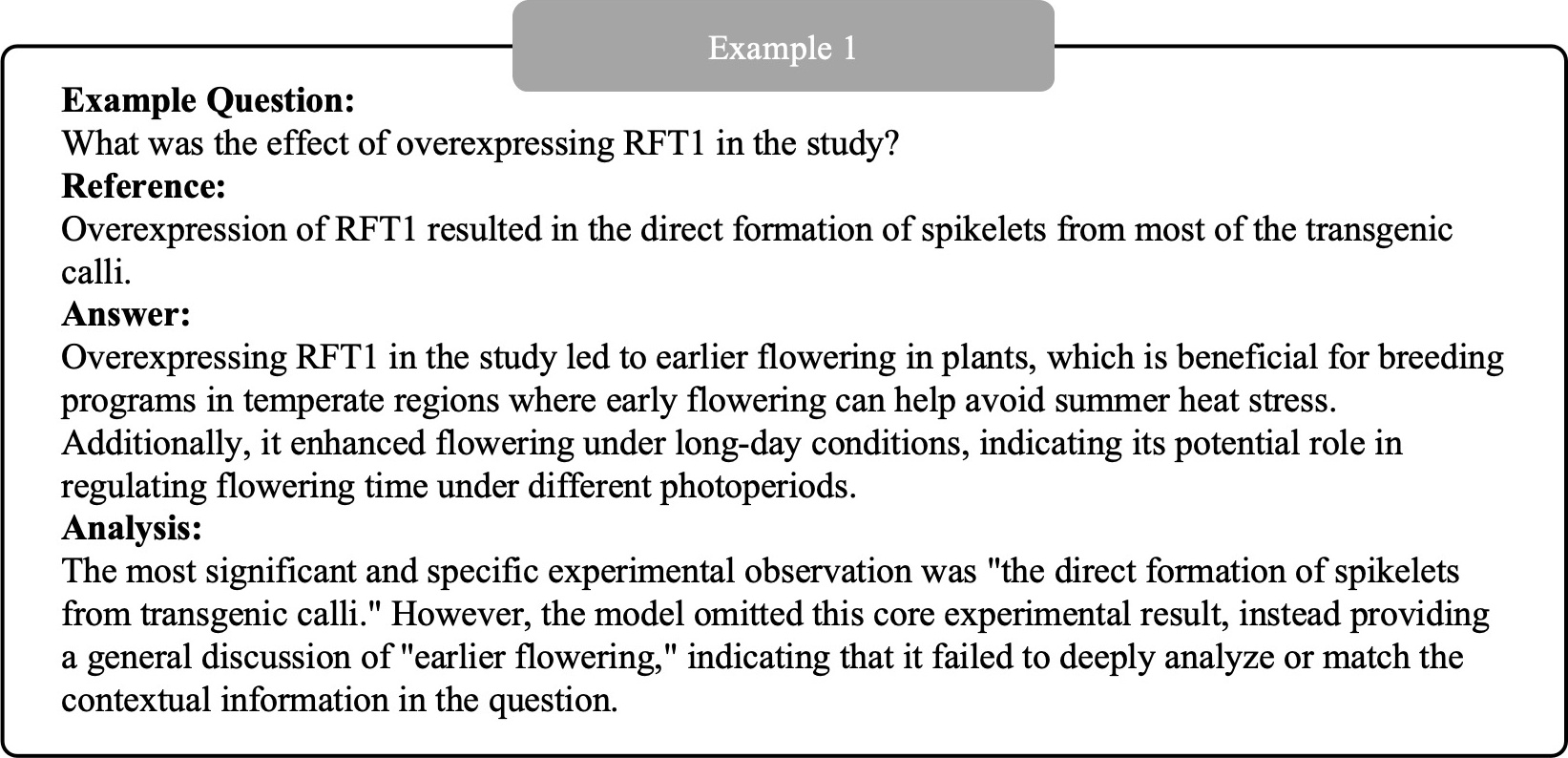}
\includegraphics[width=0.95\textwidth]{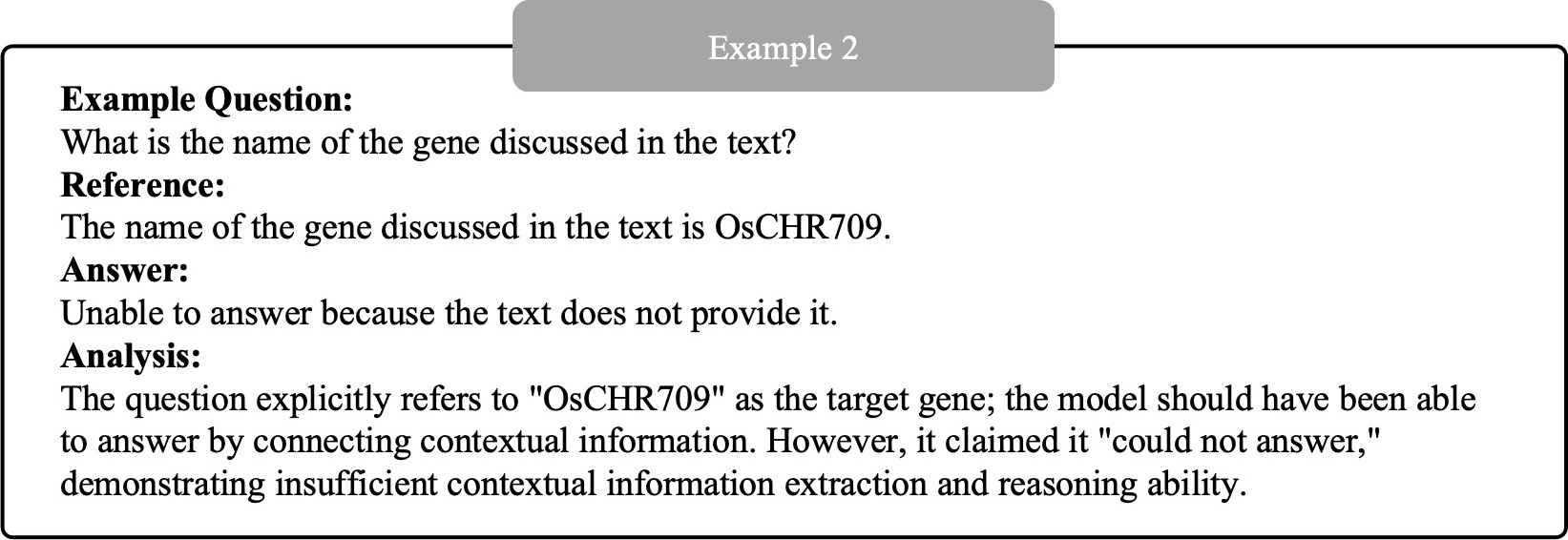}
\caption{Example illustrating limited reasoning ability.}
\label{fig:reasoning-limits}
\end{figure}

\newpage
\subsection{Errors in the Third Task Type}
\label{sec:E.3}
The third task type encompasses a wide range of needs related to variety selection and agronomic traits, requiring the model to integrate gene-level knowledge with real-world agricultural management experience. We observe several representative error cases, listed below:

\paragraph{Insufficient Semantic Understanding}
The model does not fully grasp the core descriptions or context in the question, leading to incorrect interpretations of critical information. Consequently, the final output deviates from the reference solution.
\begin{figure}[htbp]
\centering
\includegraphics[width=0.95\textwidth]{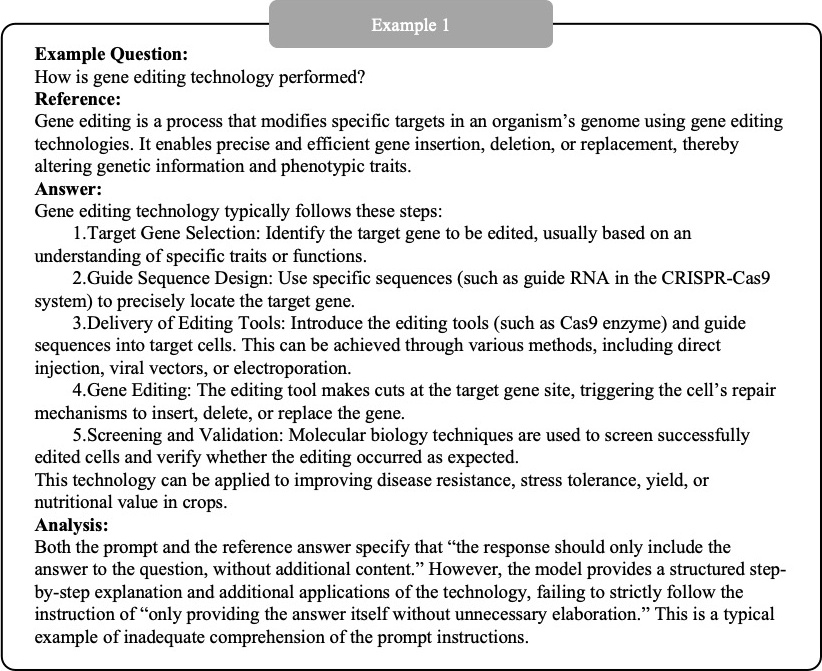}
\includegraphics[width=0.95\textwidth]{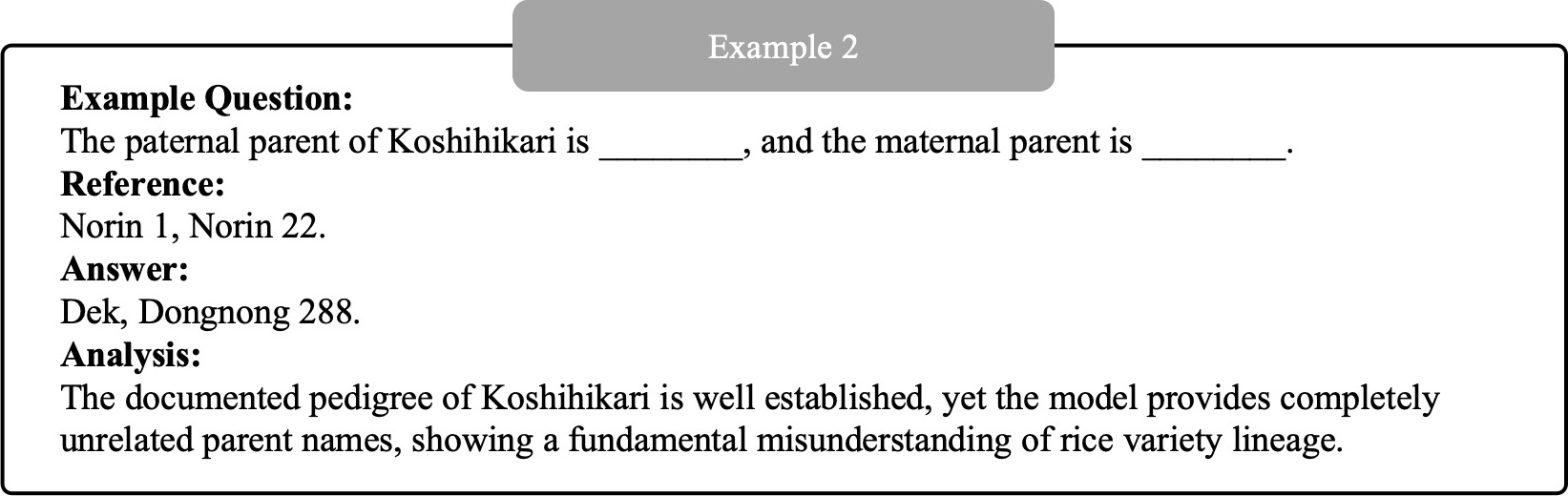}
\caption{Example illustrating insufficient semantic understanding in variety selection tasks.}
\label{fig:semantic-insufficiency}
\end{figure}

\newpage
\paragraph{Knowledge Errors}
Similar to the other task types, this category also shows frequent knowledge-related mistakes. Lacking the necessary domain background or relevant information leads the model to give incorrect answers.
\begin{figure}[htbp]
\centering
\includegraphics[width=0.95\textwidth]{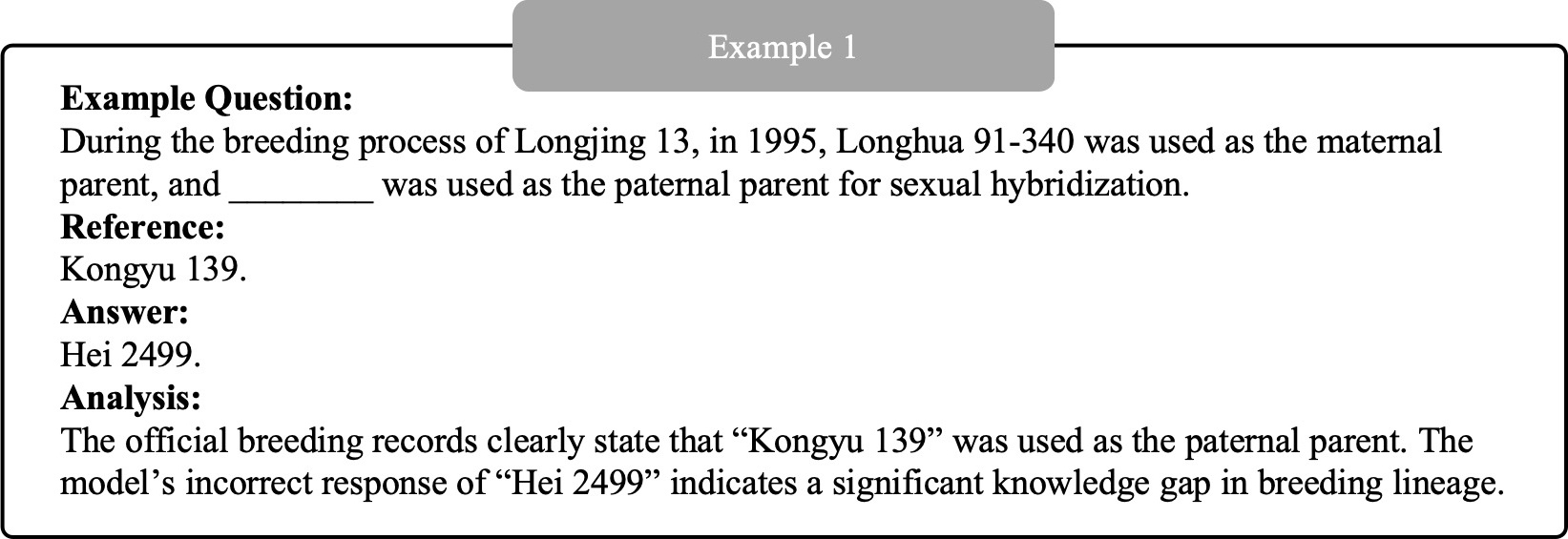}
\includegraphics[width=0.95\textwidth]{appendix_imgs/part_e/c2-2.jpg}
\includegraphics[width=0.95\textwidth]{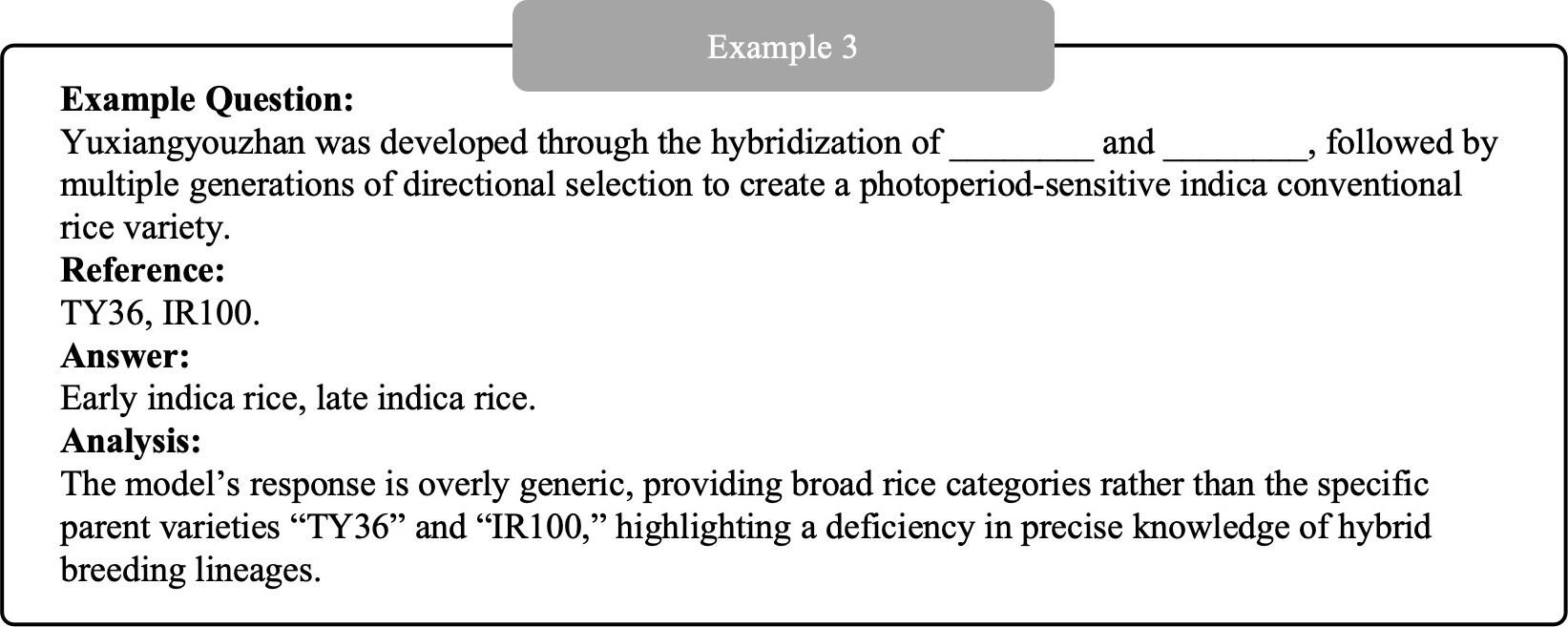}
\end{figure}

\newpage
\begin{figure}[htbp]
\centering
\includegraphics[width=0.95\textwidth]{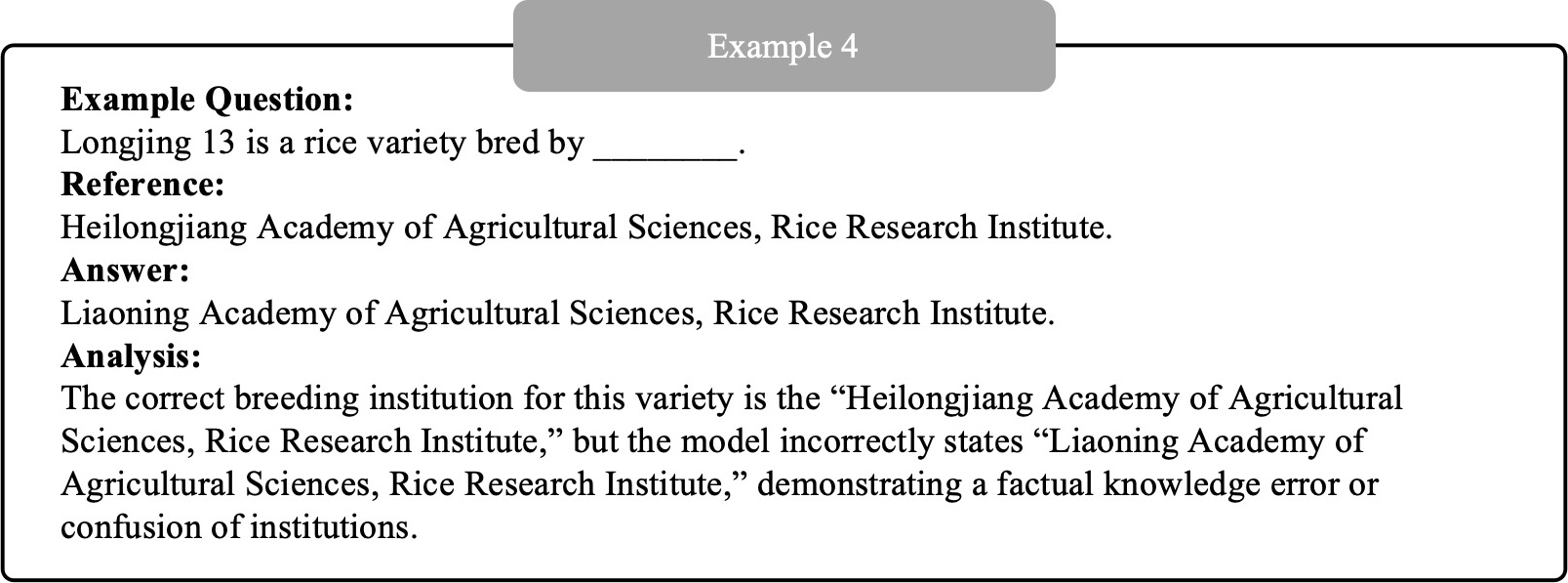}
\includegraphics[width=0.95\textwidth]{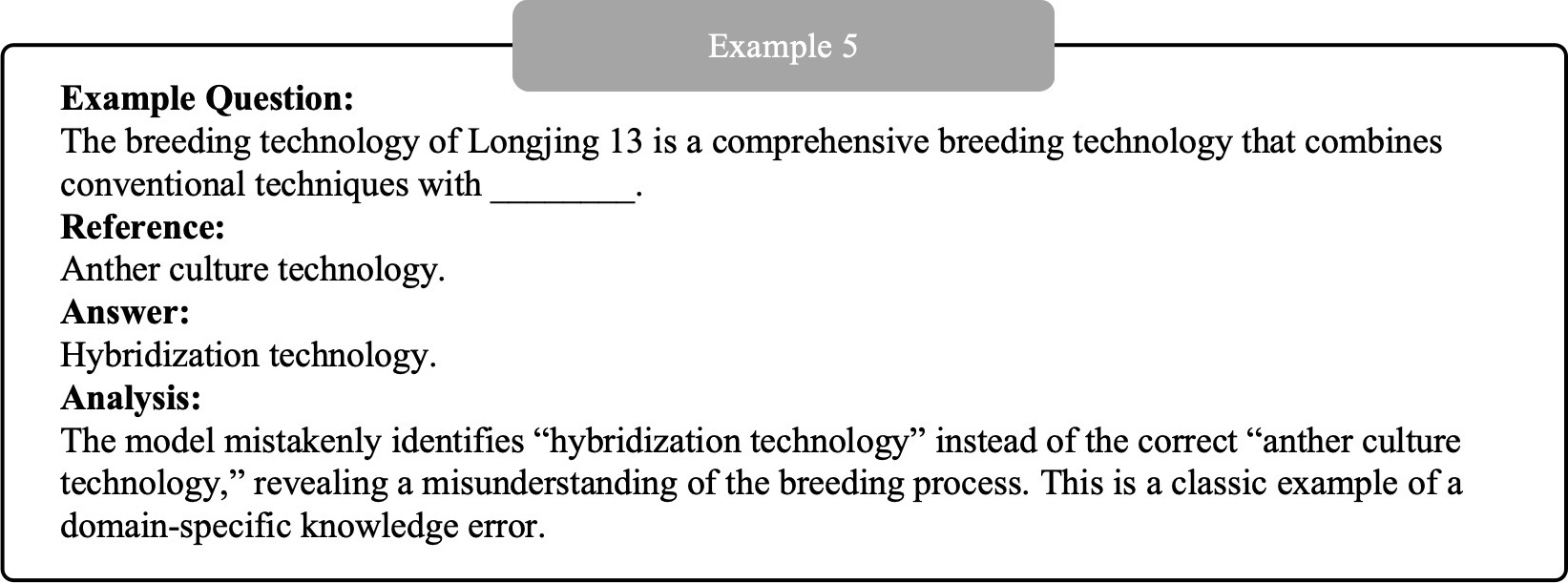}
\caption{Example illustrating knowledge errors in variety selection and agronomic traits.}
\label{fig:variety-knowledge-errors}
\end{figure}

\newpage
\section{Prompt Templates}
\label{sec:F}

\subsection{Multiple-Choice Question Generation (Task QA-1)}
\label{sec:F.1}
\begin{figure}[htbp!]
\centering
\includegraphics[width=0.95\textwidth]{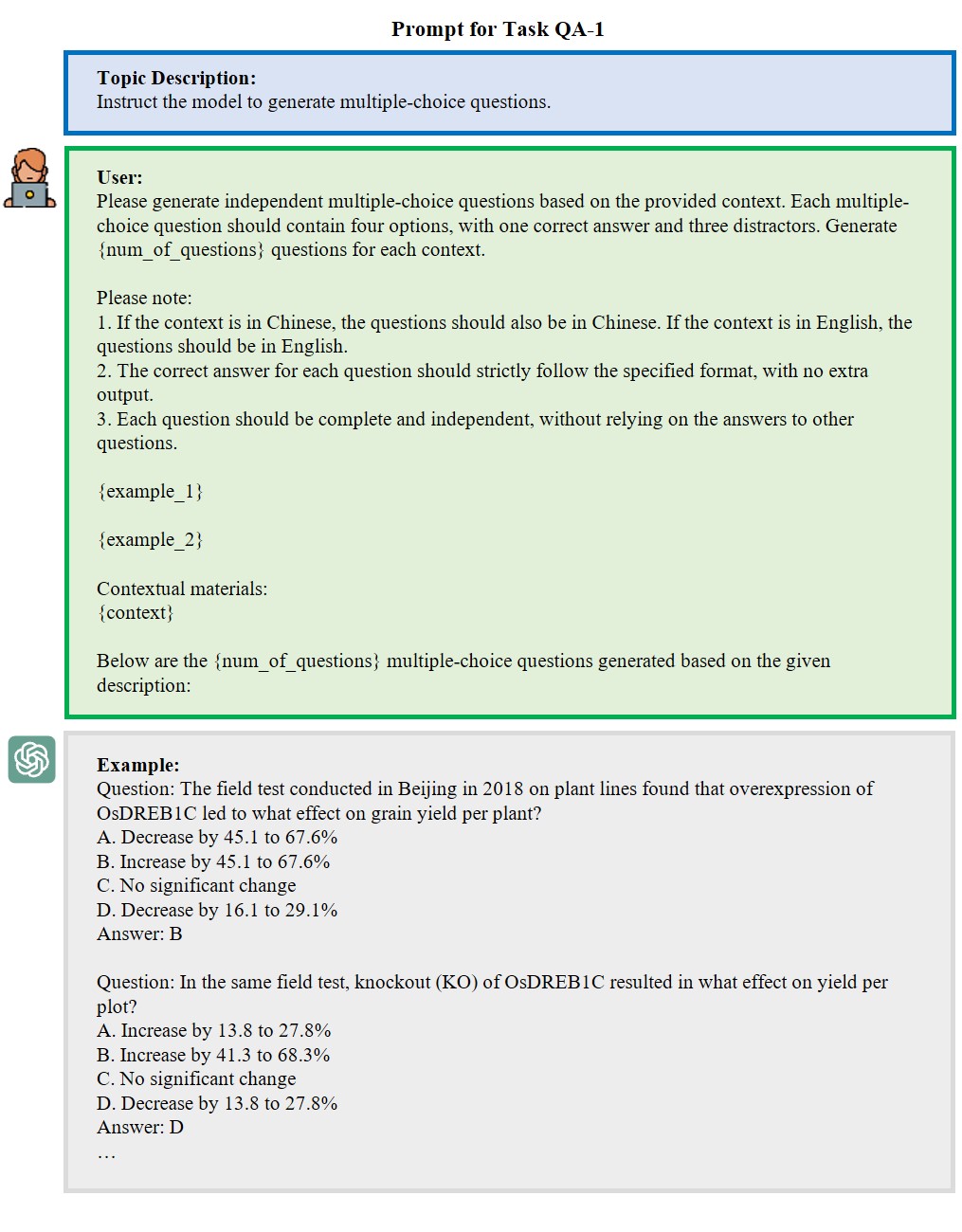}
\caption{Illustration for multiple-choice question generation.}
\label{fig:single_choice}
\end{figure}

\newpage
\subsection{Multiple-Answer Question Generation (Task QA-2)}
\label{sec:F.2}
\begin{figure}[htbp!]
\centering
\includegraphics[width=0.95\textwidth]{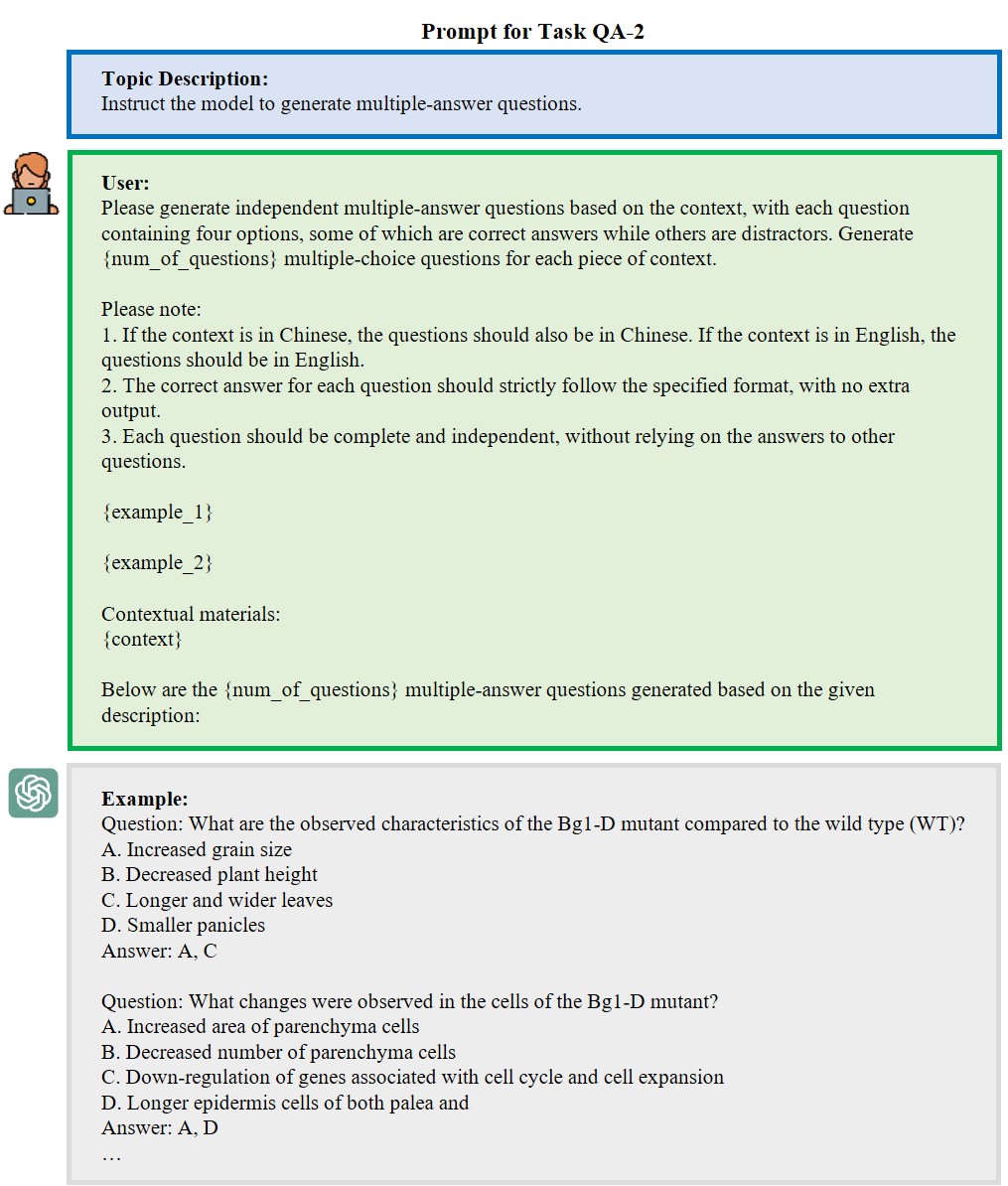}
\caption{Illustration for multiple-answer question generation.}
\label{fig:multiple_choice}
\end{figure}

\newpage
\subsection{Fill-in-the-Blank Question Generation (Task QA-3)}
\label{sec:F.3}
\begin{figure}[htbp!]
\centering
\includegraphics[width=0.95\textwidth]{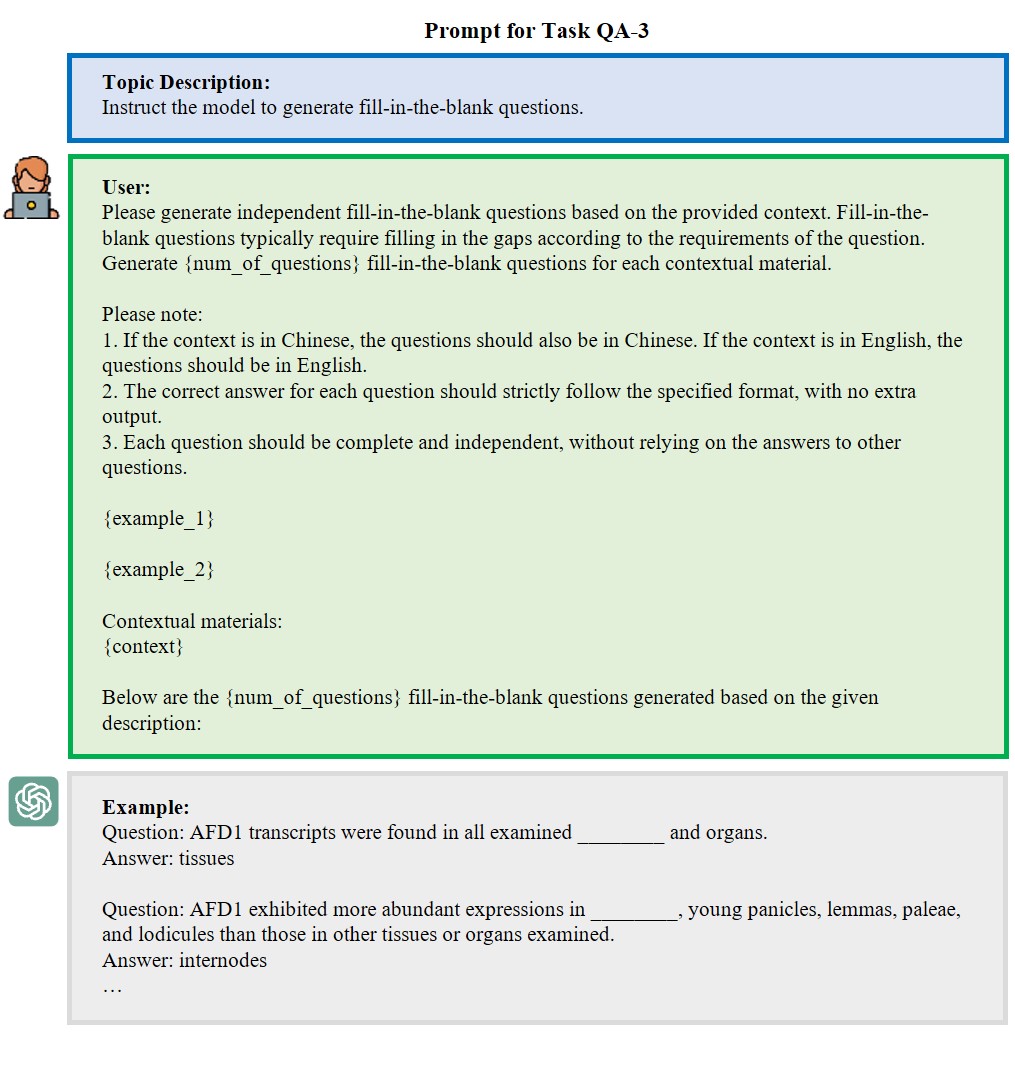}
\caption{Illustration for fill-in-the-blank question generation.}
\label{fig:fill_blank}
\end{figure}

\newpage
\subsection{Text-based Q\&A Generation (Task QA-4)}
\label{sec:F.4}
\begin{figure}[htbp!]
\centering
\includegraphics[width=0.95\textwidth]{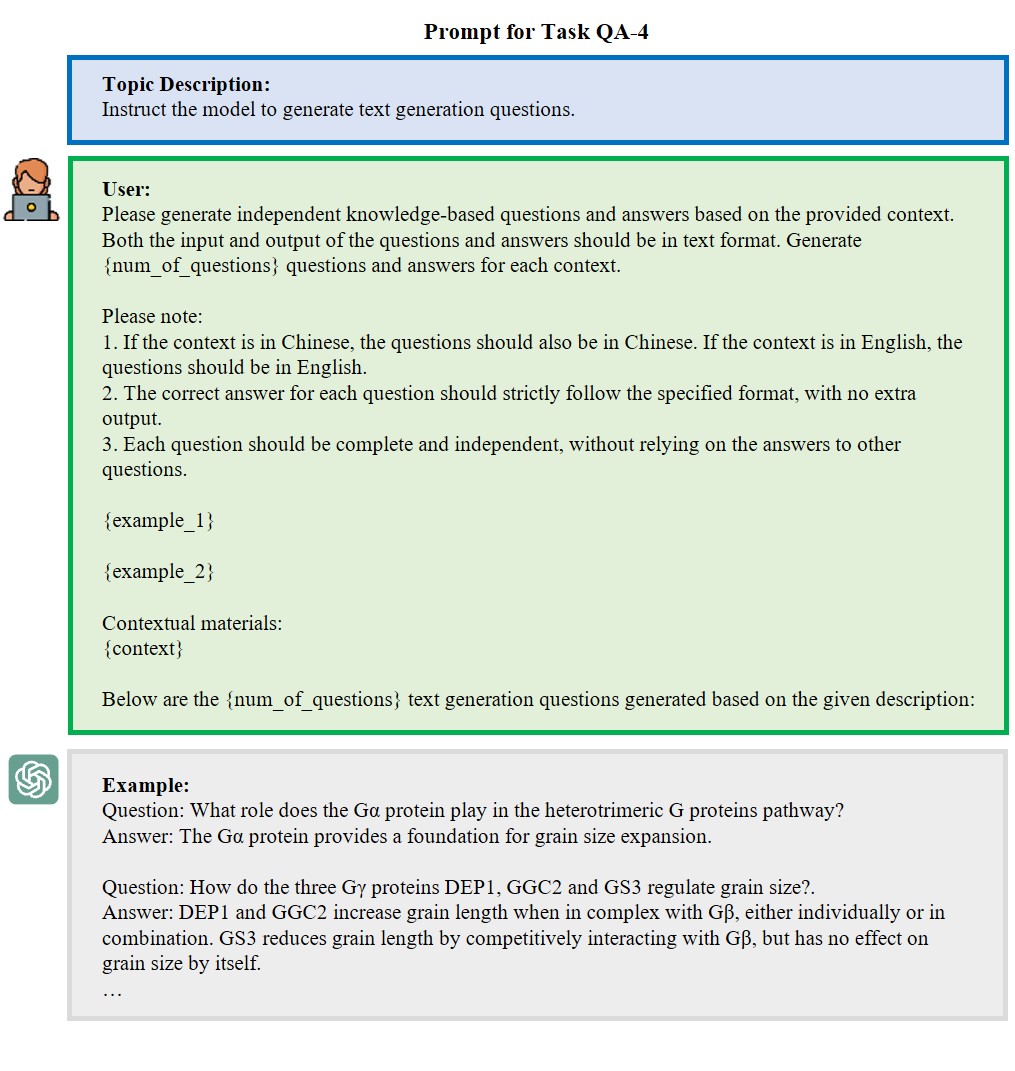}
\caption{Illustration for text-based Q\&A generation.}
\label{fig:text_qa}
\end{figure}

\newpage
\subsection{Naive Summarization (Task SUM-1)}
\label{sec:F.5}
\begin{figure}[htbp!]
\centering
\includegraphics[width=0.95\textwidth]{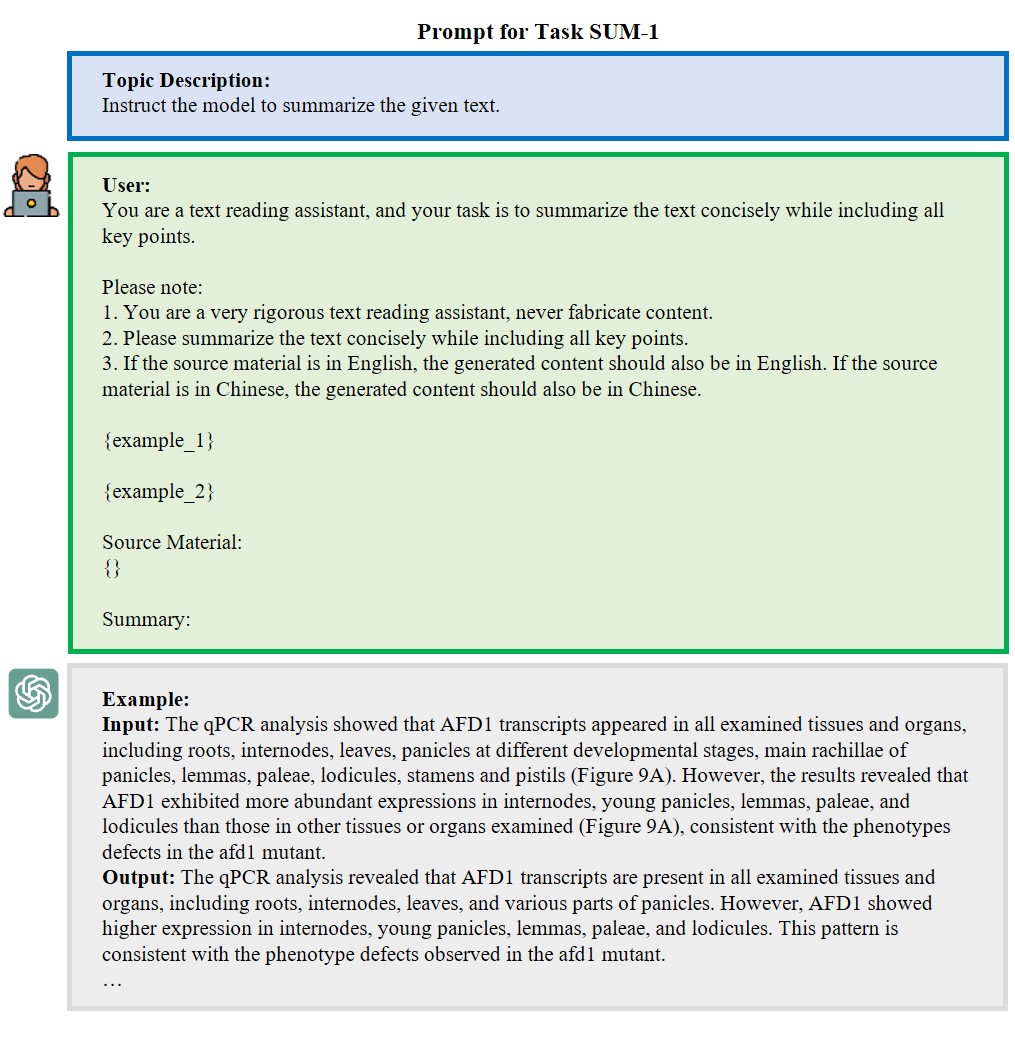}
\caption{Illustration for naive summarization.}
\label{fig:naive_summarization}
\end{figure}

\newpage
\subsection{Key Information Extraction (Task SUM-2)}
\label{sec:F.6}
\begin{figure}[htbp!]
\centering
\includegraphics[width=0.95\textwidth]{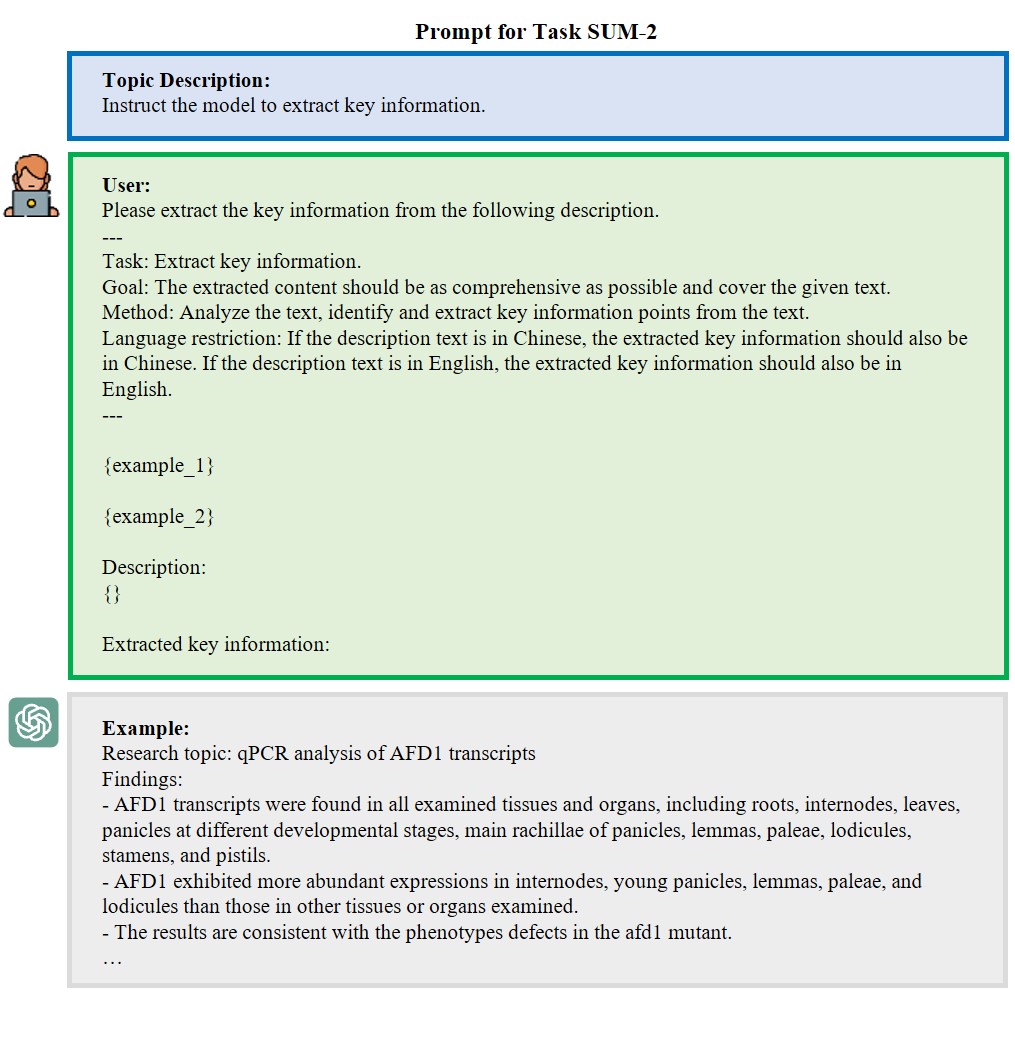}
\caption{Illustration for key information extraction.}
\label{fig:key_info_extraction}
\end{figure}

\newpage
\subsection{Question Rewriting (Tasks RC-1, RC-2, RC-3, RC-4)}
\label{sec:F.7}
\begin{figure}[htbp!]
\centering
\includegraphics[width=0.95\textwidth]{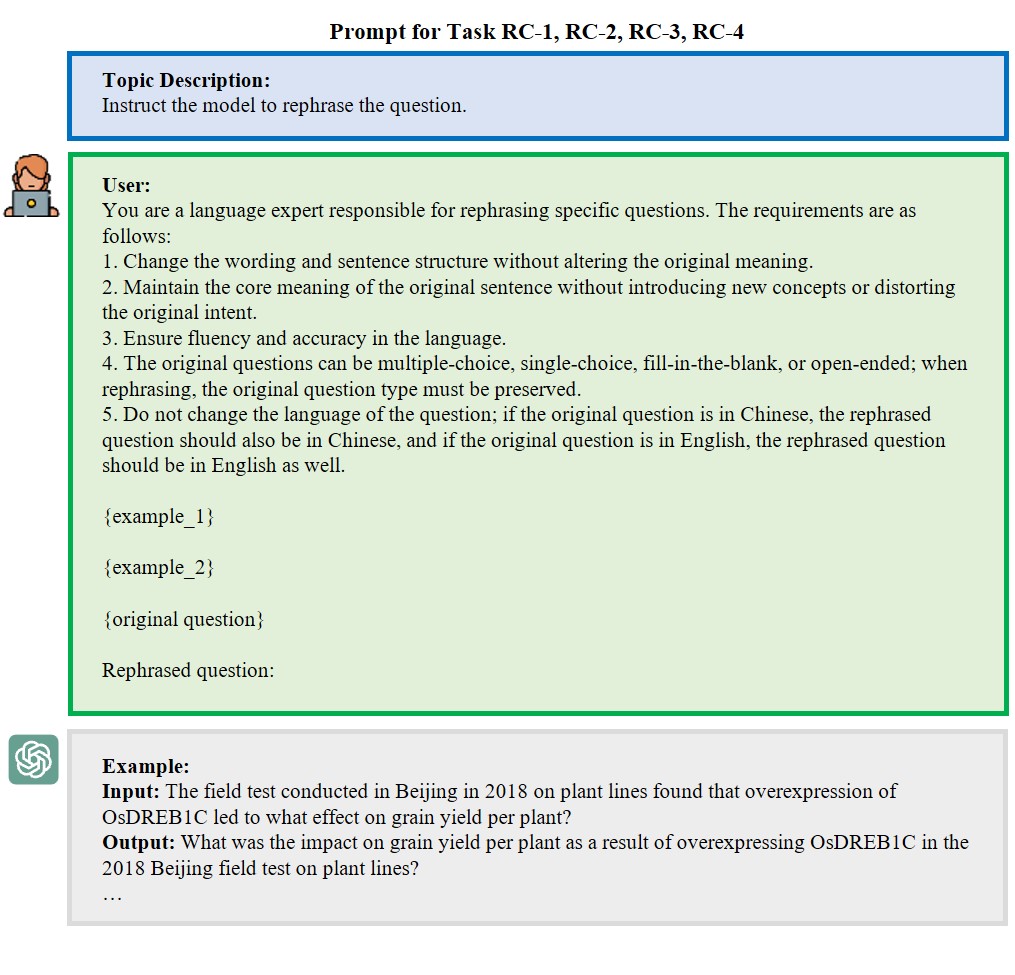}
\caption{Illustration for question rewriting across multiple subtasks.}
\label{fig:question_rewriting}
\end{figure}

\newpage
\section{Robustness Evaluation of Prompts}
\label{sec:G}

This section analyzes the sensitivity of model outputs to different prompt templates and evaluates how various prompt styles impact model performance. Specifically, we modify the wording and style of prompts and compare how different styles affect the consistency and stability of model-generated results.

\begin{figure}[htbp!]
\centering
\includegraphics[width=0.95\textwidth]{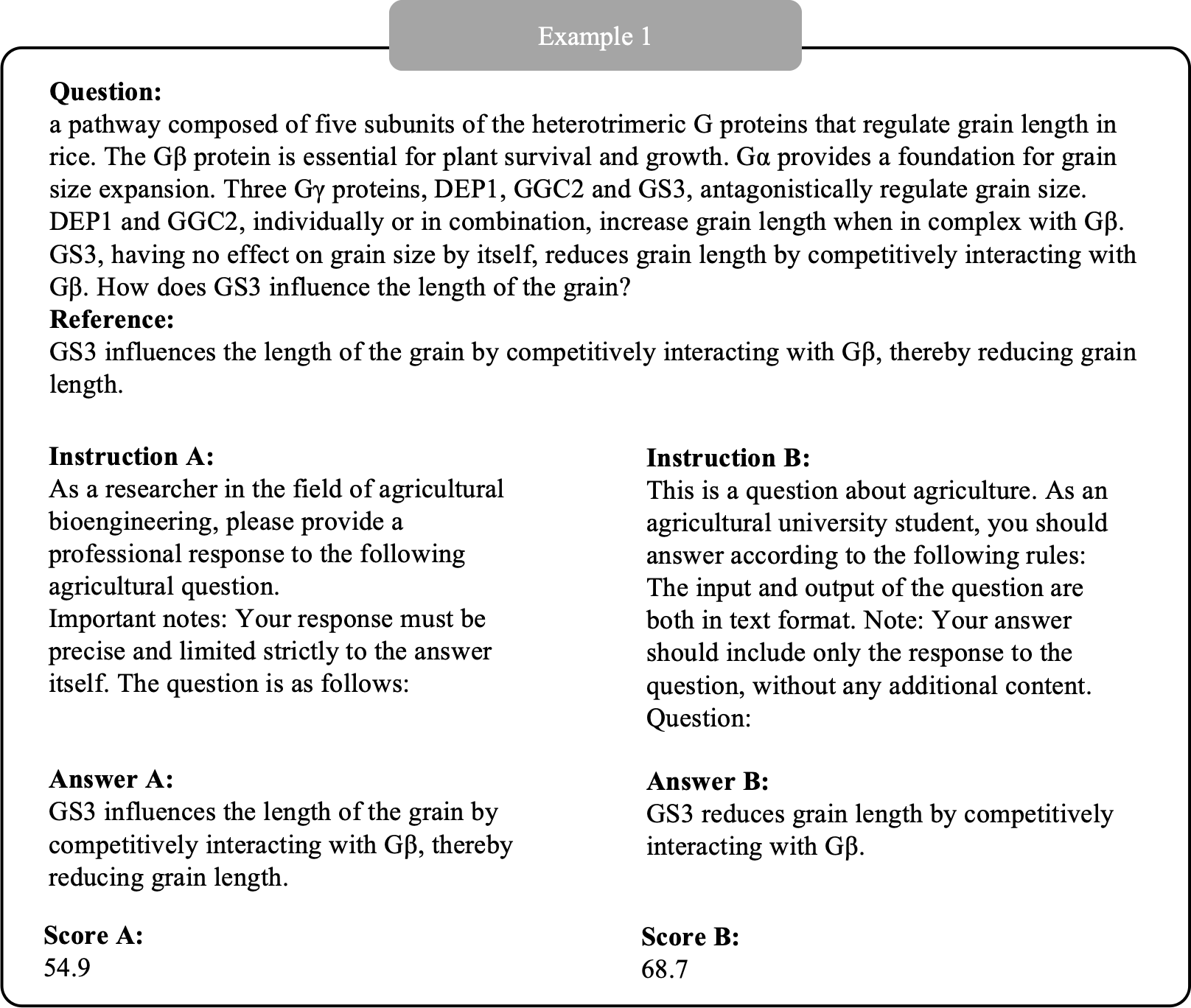}
\end{figure}

\newpage
\begin{figure}[htbp!]
\centering
\includegraphics[width=0.95\textwidth]{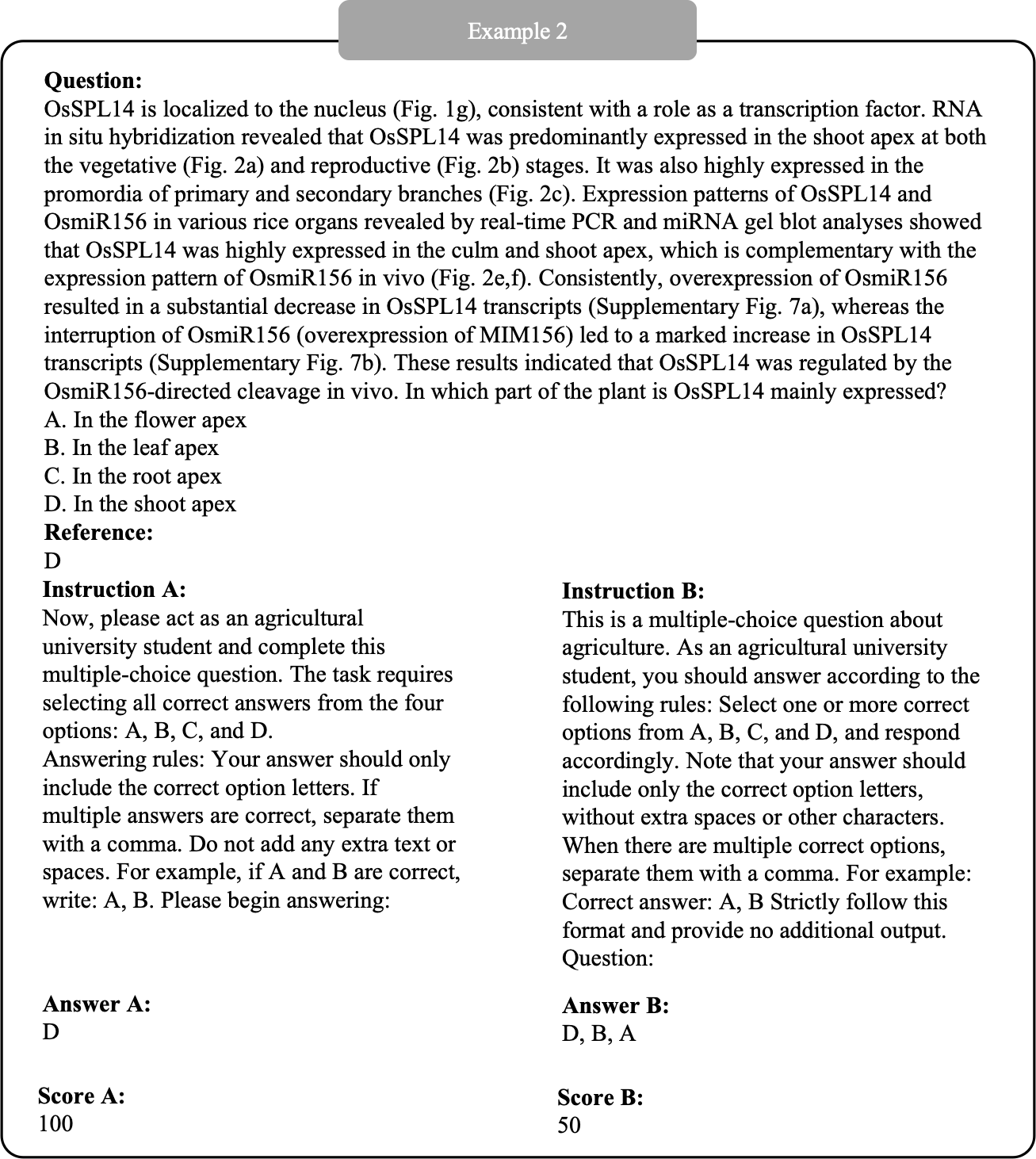}
\end{figure}

\newpage
\begin{figure}[htbp!]
\centering
\includegraphics[width=0.95\textwidth]{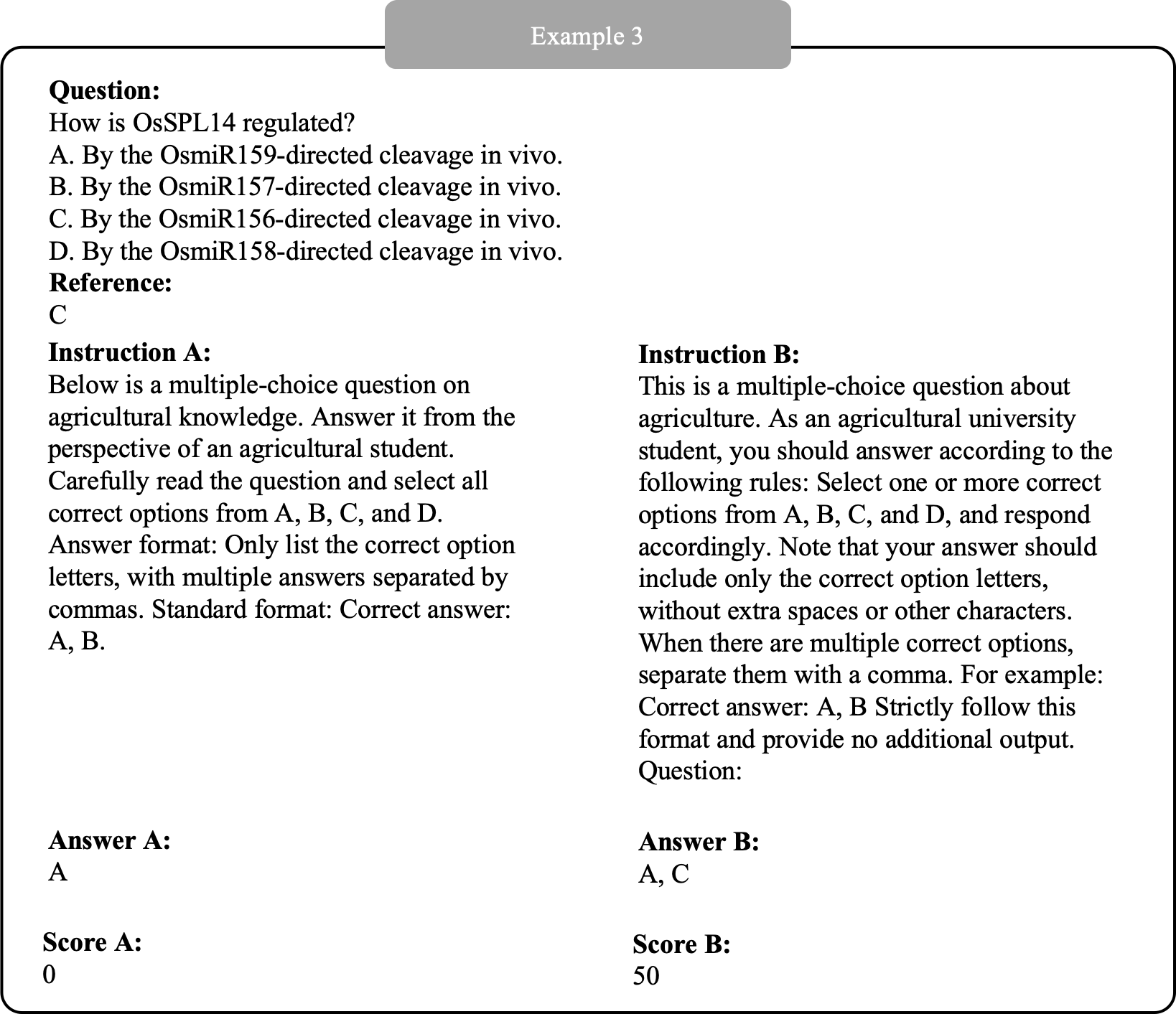}
\end{figure}

\newpage
\begin{figure}[htbp!]
\centering
\includegraphics[width=0.95\textwidth]{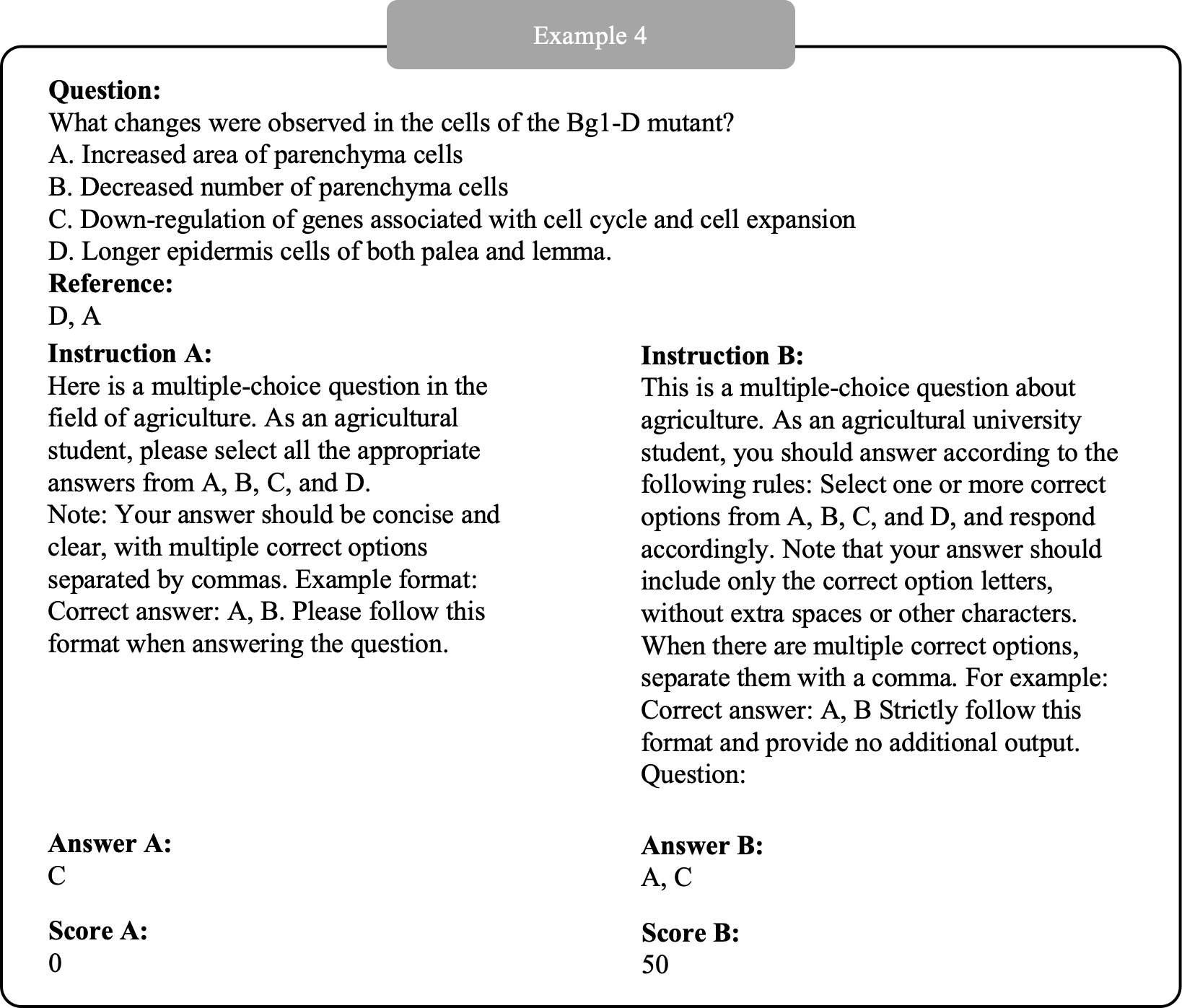}
\end{figure}

\newpage
\begin{figure}[htbp!]
\centering
\includegraphics[width=0.95\textwidth]{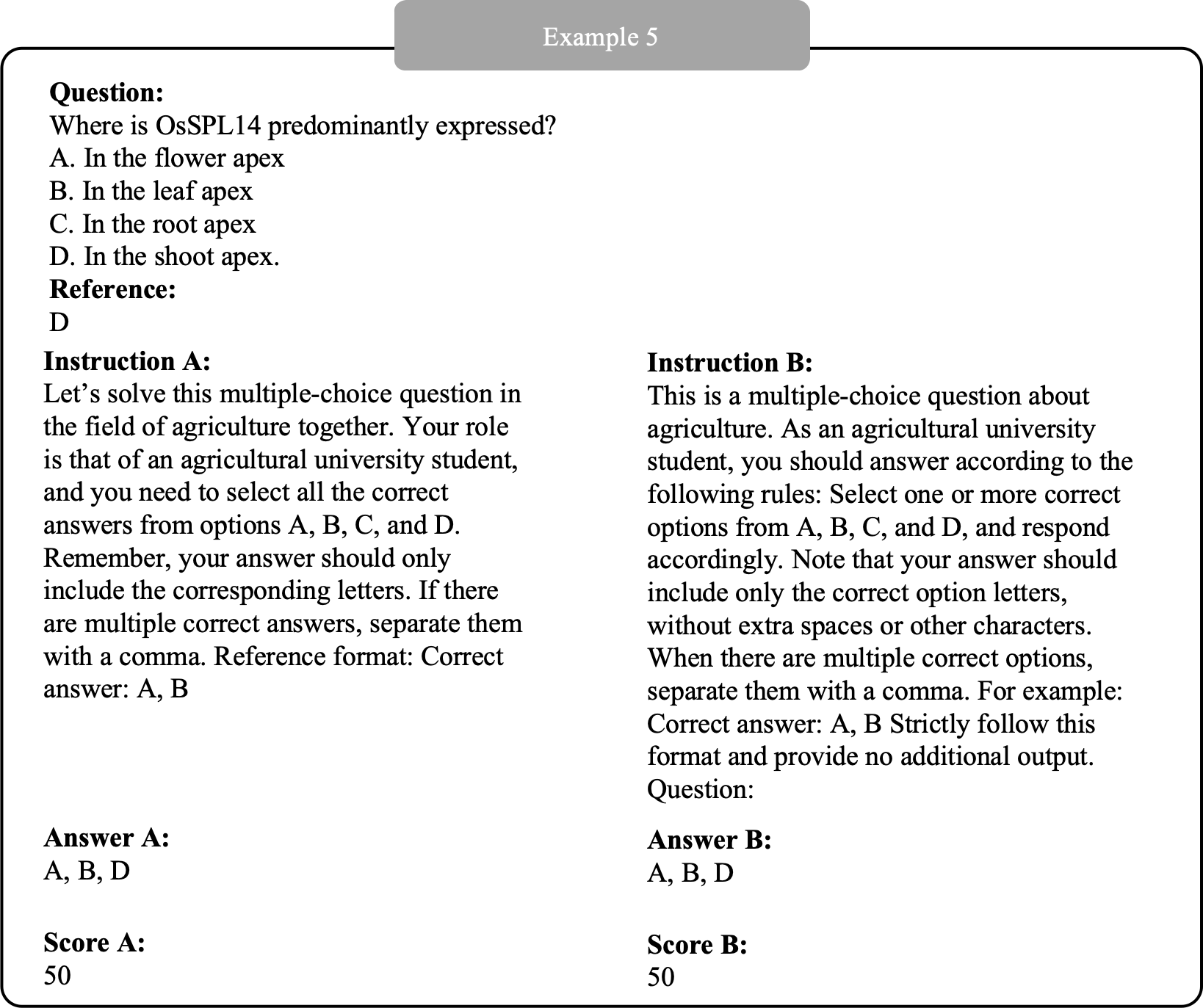}
\end{figure}

\newpage
\begin{figure}[htbp!]
\centering
\includegraphics[width=0.95\textwidth]{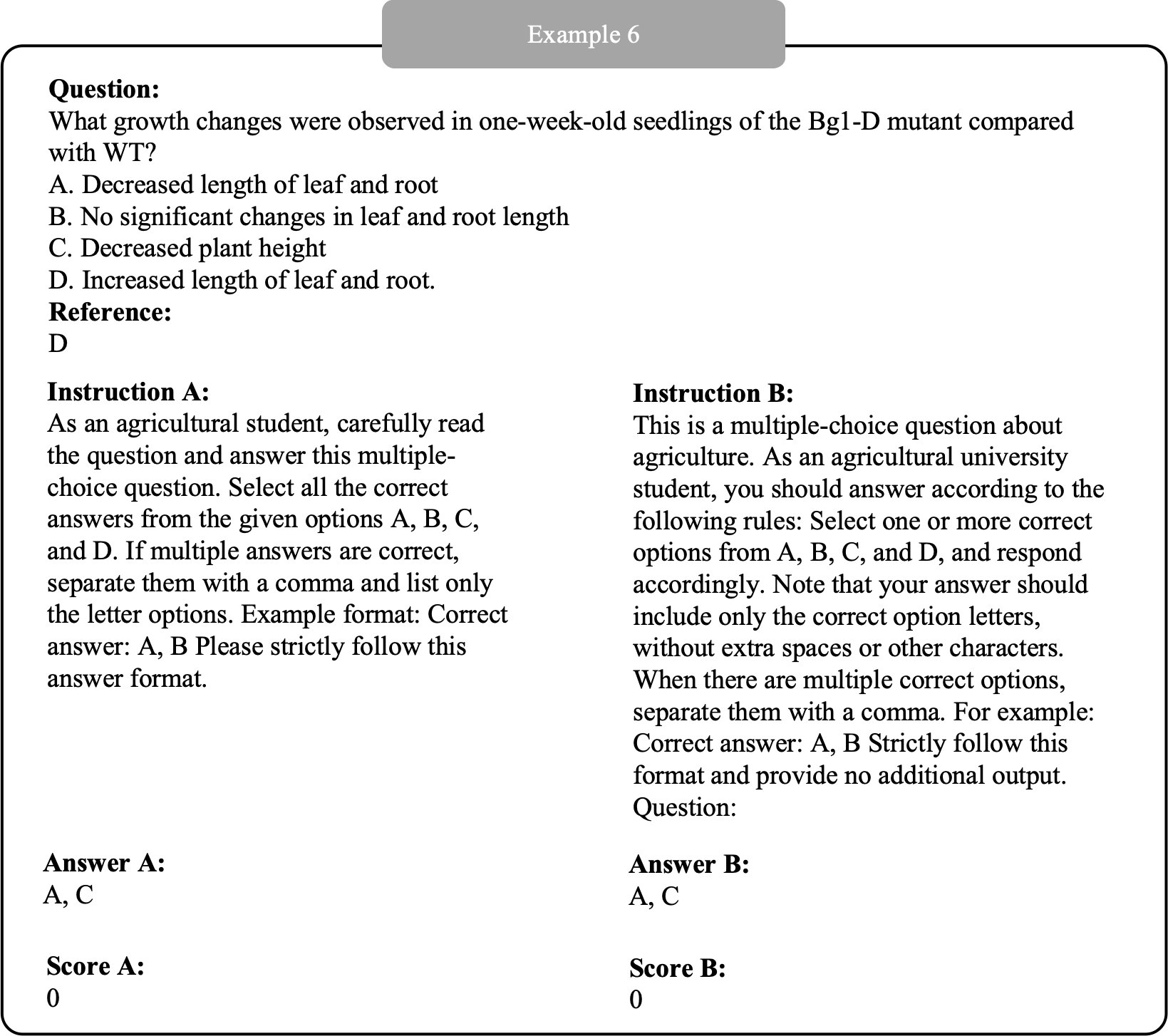}
\end{figure}

\newpage
\begin{figure}[htbp!]
\centering
\includegraphics[width=0.95\textwidth]{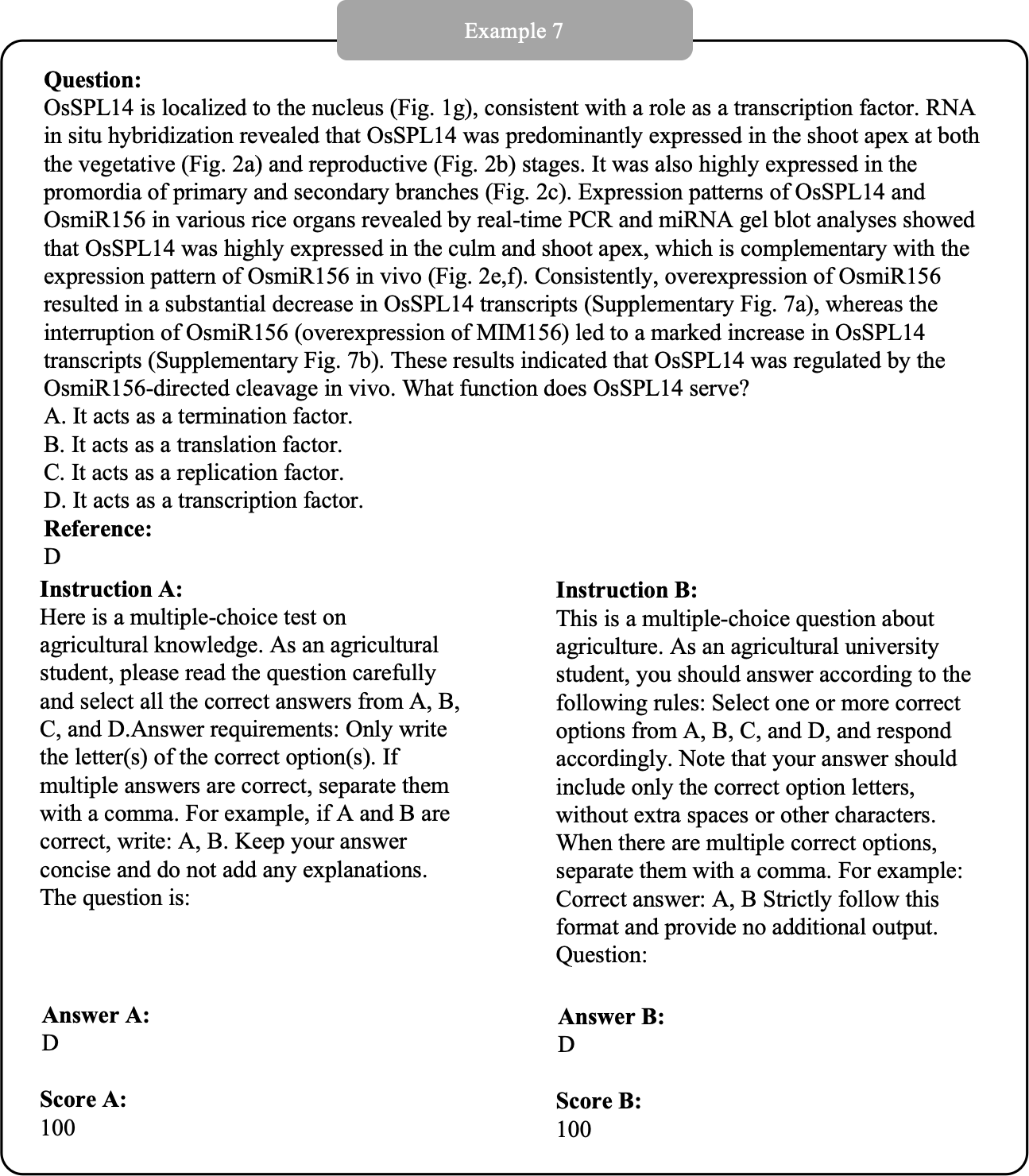}
\end{figure}

\newpage
\begin{figure}[htbp!]
\centering
\includegraphics[width=0.95\textwidth]{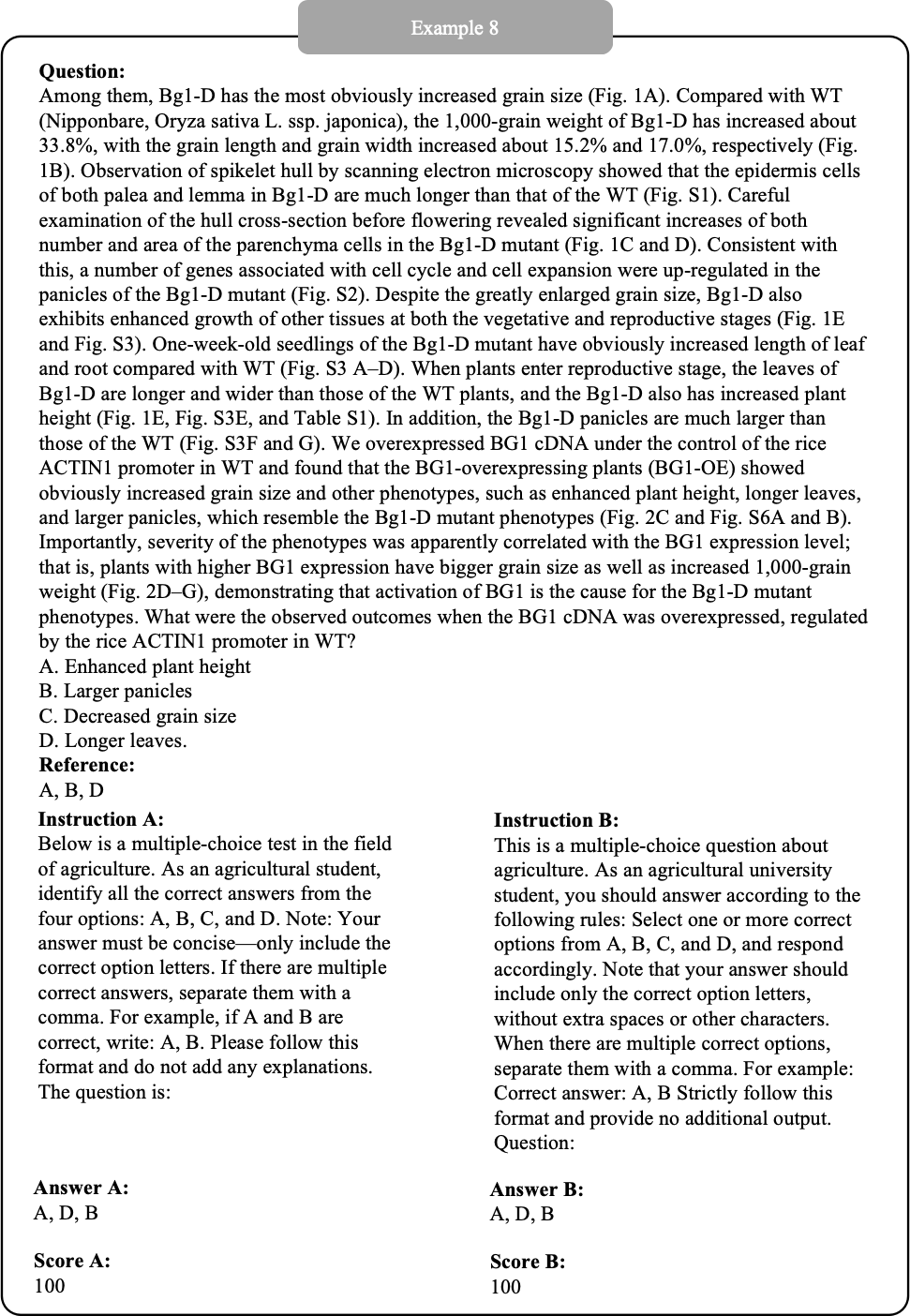}
\caption{Illustration of robustness evaluation for different prompt styles.}
\label{fig:prompt_robustness}
\end{figure}

\newpage
\section{Detailed Performance Comparison and Analysis}
\label{sec:H}
To analyze the experiment results from the provided table, we can examine several aspects, including the performance of different models, the comparison between different sizes of the same series, the variation in performance across tasks, and the relationship between model size and performance.

\subsection{Model Comparison on Different LLMs}
\label{sec:H.1}
\textbf{Proprietary Models:} Among the proprietary models, GPT-4 consistently performs well across most of the tasks, often achieving top or second place in many breeding subcategories (C1, C3, C5, C6). The model's average score is notably high at 62.06, which is the highest among proprietary models.

\noindent\textbf{Open-Source Models:} Models such as DeepSeek-V3-671B and Qwen2-57B exhibit strong performance in certain categories (e.g., Qwen2-57B performs exceptionally well in C3 and C7).

\subsection{Comparison Between Different Model Sizes of the Same Model Series}
\label{sec:H.2}
\textbf{Qwen Models:} The different model sizes of the Qwen models show varying performance trends. While Qwen2-7B, a mid-tier model, significantly outperforms Qwen2-0.5B, it falls short of Qwen2-72B, the largest model in the series. However, contrary to expectations, increasing model size does not guarantee progressive performance gains. For instance, Qwen2.5-72B fails to demonstrate clear improvements over Qwen2.5-14B in many subcategories, suggesting that parameter count alone may not dictate performance in this domain. Notably, Qwen2.5-14B excels in several subcategories (e.g., C6, C8), achieving top-tier rankings, whereas Qwen2.5-72B shows inconsistent superiority—exhibiting only marginal gains in C3 and C6. This pattern implies diminishing returns for larger models in specific subcategories, highlighting the importance of training data and strategies beyond mere scale.

\subsection{Top Performers in Each Subcategory}
\label{sec:H.4}

\hspace{1em}\textbf{C1}: (Gene Basic Information Query: GPT-4, 59.59)

\textbf{C2}: (Gene Expression Pattern Query: DeepSeek-V3, 62.42)

\textbf{C3}: (Gene Product Cellular Localization Query: GPT-4, 76.32)

\textbf{C4}: (Gene Function Experimental Observation: DeepSeek-V3, 63.17)

\textbf{C5}: (Gene Product Regulation of Downstream Genes Analysis: GPT-4, 56.34)

\textbf{C6}: (Gene Function Prediction: GPT-4, 59.35)

\textbf{C7}: (Variety Breeding Process Query: DeepSeek-V3, 68.23)

\textbf{C8}: (Variety Agronomic Trait Query: DeepSeek-V3, 69.04)

\textbf{C9}: (Variety Cultivation and Technical Key Points Query: DeepSeek-V3, 66.46)

\textbf{C10}: (Variety Suitable Planting Area Recommendation: DeepSeek-V3, 68.48)

\noindent\textbf{Consistent Top Performers:} DeepSeek-V3-671B is a standout performer, securing top-3 positions in all ten subcategories and ranking first in C2 (62.42), C4 (63.17), C7 (68.23), C8 (69.04), C9 (66.46), and C10 (68.48). Its dominance in moderately hard to medium tasks (e.g., C7, C8, C10) underscores its strength in breeding-related applications.

\noindent\textbf{Other Consistent Top Performers:} GPT-4 also demonstrates exceptional consistency, appearing in the top-3 for nine out of ten subcategories (C1, C2, C3, C4, C5, C6, C8, C9, C10) and ranking first in C1 (59.59), C3 (76.32), C5 (56.34), and C6 (59.35). Its strong performance across both simpler tasks (e.g., C3) and more complex ones (e.g., C5) highlights its versatility and robustness. GLM-4-Plus frequently appears in the top-3 for six subcategories (C2, C4, C7, C8, C9, C10), with notable scores such as 65.02 in C7 and 64.17 in C8, making it a reliable performer, particularly in medium-difficulty tasks. Additionally, Qwen2-57B and Qwen2-72B occasionally appear in the top-3 (e.g., C3, C7 for Qwen2-57B; C5 for Qwen2-72B), but their consistency is less pronounced compared to GPT-4 and DeepSeek-V3-671B.

\subsection{Comparison of Models with Same Parameter Scale}
\label{sec:H.6}
\begin{center}
\begin{tabular}{|l|l|l|l|}
\hline
\textbf{Parameter Level} & \textbf{Best Model (Avg.)} & \textbf{Second Best (Avg.)} & \textbf{Advantaged Domains} \\ 
\hline
7B & InternLM2.5-7B (53.51) & Qwen2.5-7B (48.45) & C3, C9 \\
70B & Qwen2-72B (57.62) & Llama3.1-70B (54.30) & C2, C5 \\
large-scale & DeepSeek-V3-671B (63.30) & GPT-4 (62.06) & C5, C10 \\
\hline
\end{tabular}
\end{center}

\subsection{Task-Specific Performance}
\label{sec:H.7}
\textbf{Task Difficulty Spectrum} (Based on average scores across all models):
\begin{center}
\begin{tabular}{|l|l|c|}
\hline
\textbf{Difficulty Level} & \textbf{Subcategory} & \textbf{Avg. Score}  \\
\hline
Most Difficult & C5 & 37  \\
Moderately Hard & C1, C2, C4, C9 & 42-45  \\
Medium & C6, C7, C8,C10 & 46-49  \\
Easiest & C3 & 58  \\
\hline
\end{tabular}
\end{center}

\noindent\textbf{Model Performance on Specific Subcategories (C1–C10)}

The performance of large language models (LLMs) across the ten breeding subcategories (C1–C10) reveals distinct patterns in task difficulty and model capability. By classifying tasks based on average model performance, we observe that larger models excel in complex tasks, while smaller models can achieve competitive results in simpler ones. The following analysis aligns with the provided difficulty classification, adjusted to reflect observed performance trends.

\noindent\textbf{Most Difficult Task (C5, Avg. Score ~37):} Gene Product Regulation of Downstream Genes Analysis (C5) is the most complex, with a wide performance gap. Larger models like GPT-4 (56.34) and DeepSeek-V3-671B (55.23) lead, but smaller models, such as Qwen2-0.5B (27.62), GLM-4-Chat-9B (16.20), and PLLaMa-7B (11.66), score significantly lower, aligning with the ~37 average for mid-to-low-tier models. This task demands advanced reasoning and knowledge integration, favoring larger models.

\noindent\textbf{Easiest Task (C3, Avg. Score ~58):} Gene Product Cellular Localization Query (C3) is the least challenging, with top models like GPT-4 (76.32) and DeepSeek-V3-671B (74.81) achieving high scores. Even smaller models, such as InternLM2.5-7B (67.88) and Qwen2.5-7B (66.01), perform well, indicating that C3 requires less computational capacity and is accessible to models with fewer parameters. The average score of ~58 reflects mid-tier model performance, though top performers significantly exceed this.

\noindent\textbf{Additional Insights:} Proprietary models (e.g., GPT-4) and large open-source models (e.g., DeepSeek-V3-671B) dominate across all categories, particularly in difficult tasks like C5. Domain-specific models, such as Aksara-v1-7B (24.26 in C5) and PLLaMa-13B (13.96 in C5), underperform, suggesting limited generalization. The consistent presence of GPT-4 and DeepSeek-V3-671B in top-3 rankings underscores the advantage of model scale in complex tasks, while smaller models remain viable for easier tasks.

In summary, the difficulty classification highlights that smaller models can perform competitively in easier tasks (e.g., C3), but larger models are essential for moderately hard to difficult tasks (e.g., C5, C1, C2). This emphasizes the importance of selecting appropriately scaled models to match task complexity in breeding-related applications.

\subsection{Conclusion}
\label{sec:H.8}

\textbf{Model Choice:} DeepSeek-V3-671B emerges as the top performer overall and should be considered the best model for most tasks. However, models like GPT-4 and  GLM-4-Plus demonstrate competitive performance and could be preferable for certain tasks.

\noindent\textbf{Task-Level Analysis:} For more difficult tasks such as C1 and C5, larger models like GPT-4 tend to excel, while Qwen2-57B performs strongly in easier categories such as C3. 



\noindent\textbf{Insights from Scaling Law:} 
Conventional scaling laws suggest that larger models, such as Qwen2.5-72B, should consistently outperform smaller models like Qwen2.5-7B and Qwen2.5-14B due to their greater model sizes and capacity. However, the experimental results challenge this expectation, as Qwen2.5-7B often performs comparably to or even surpasses its larger counterparts in several subcategories (e.g., 53.06 in C9 vs. 53.05 for Qwen2.5-72B). This discrepancy indicates that model size alone does not guarantee superior performance. Instead, factors such as training data distribution, task-specific optimization, and the nature of the tasks play critical roles in determining model effectiveness.

The findings suggest that larger models may not be fully optimized for the specific task set evaluated here, potentially due to mismatches between their training data and the demands of breeding-related subcategories. For certain tasks, smaller models like Qwen2.5-7B may be better suited, particularly when tasks align closely with their training or require less complex reasoning. These results challenge the universal applicability of scaling laws across all task types, highlighting the need for careful consideration of training data characteristics and task design when selecting model size for optimal performance.

\subsection{Assessment of the Generative Tasks using BERTScore}
\label{sec:H.9}

We have conducted an assessment of the generative tasks (QA-4, SUM-1, SUM-2, RC-3, RC-4) using BERTScore, and calculated the Pearson correlation coefficient between BERTScore and ROUGE-L. The results are as follows:

\begin{table}[h]
\centering
\caption{Pearson correlation between BERTScore and ROUGE-L on generative tasks.}
\begin{tabular}{lc}
\hline
\textbf{Task} & \textbf{Pearson Correlation (BERTScore vs. ROUGE-L)} \\
\hline
QA-4   & 0.7937 \\
SUM-1  & 0.6770 \\
SUM-2  & 0.5811 \\
RC-3   & 0.7515 \\
RC-4   & 0.7338 \\
\hline
\end{tabular}
\end{table}

The following table reports the BERTScore (\%) for five generative task subsets in SeedBench.

\begin{longtable}{lccccc}
\caption{BERTScore (\%) across models on five generative tasks.} \\
\hline
\textbf{Model} & \textbf{QA-4} & \textbf{SUM-1} & \textbf{SUM-2} & \textbf{RC-3} & \textbf{RC-4} \\
\hline
\endfirsthead
\hline
\textbf{Model} & \textbf{QA-4} & \textbf{SUM-1} & \textbf{SUM-2} & \textbf{RC-3} & \textbf{RC-4} \\
\hline
\endhead
Claude-3.5-Sonnet & 48.43 & 50.74 & 51.83 & 43.28 & 48.47 \\
Gemini-1.5-Pro & 72.39 & 74.89 & 73.89 & 95.59 & 79.48 \\
Gemini-2.0-Flash & 64.60 & 74.84 & 70.85 & 61.71 & 72.00 \\
GLM-4-Plus & 79.70 & 83.42 & 80.19 & 95.03 & 84.34 \\
GPT-4o mini & 81.10 & 83.50 & 82.79 & 93.98 & 86.01 \\
GPT-4 & 79.66 & 87.45 & 85.95 & 96.02 & 88.27 \\
OpenAI o1-mini & 77.36 & 75.18 & 69.91 & 94.61 & 81.97 \\
DeepSeek-V3 & 81.22 & 83.04 & 81.55 & 94.25 & 85.89 \\
GLM-4-Chat-9B & 56.10 & 69.45 & 74.72 & 51.17 & 60.93 \\
InternLM2-7B & 57.49 & 72.76 & 74.04 & 53.83 & 61.34 \\
InternLM2.5-7B & 76.97 & 84.07 & 81.63 & 91.01 & 85.72 \\
Llama3.1-8B & 74.91 & 82.11 & 77.87 & 83.28 & 81.18 \\
Llama3.1-70B & 77.97 & 83.84 & 82.62 & 91.12 & 82.98 \\
Llama3.3-70B & 76.35 & 81.89 & 79.08 & 87.91 & 82.52 \\
Mistral-v0.3-7B & 75.13 & 81.98 & 84.54 & 75.94 & 83.66 \\
Qwen2-0.5B & 75.27 & 78.37 & 76.46 & 64.88 & 79.64 \\
Qwen2-7B & 77.21 & 78.71 & 78.85 & 84.85 & 82.60 \\
Qwen2-57B & 80.31 & 82.82 & 83.93 & 92.00 & 85.83 \\
Qwen2-72B & 78.75 & 80.23 & 85.11 & 94.51 & 84.53 \\
Qwen2.5-7B & 77.66 & 81.19 & 82.19 & 82.82 & 83.35 \\
Qwen2.5-14B & 75.30 & 75.38 & 62.66 & 83.46 & 82.80 \\
Qwen2.5-72B & 79.50 & 84.38 & 83.50 & 93.16 & 83.14 \\
QwQ-32B & 70.16 & 70.43 & 69.57 & 65.62 & 73.06 \\
Aksara-v1-7B & 71.48 & 77.60 & 80.33 & 70.86 & 79.36 \\
PLLaMa-7B & 69.53 & 64.44 & 63.62 & 62.41 & 69.60 \\
PLLaMa-13B & 61.73 & 62.46 & 59.66 & 56.95 & 66.64 \\
\hline
\end{longtable}

\newpage
\subsection{Additional Quantitative Results}
\label{sec:H.10}
\begin{table*}[h!]
  \centering
  \resizebox{\textwidth}{!}{
    \small
    \setlength{\tabcolsep}{3.5pt}
    \begin{tabularx}{\textwidth}{l*{11}{r}}
      \hline
      \multirow{2}{*}{\textbf{Models}} & \multicolumn{10}{c}{\textbf{Breeding Subcategories}} & \multirow{2}{*}{\textbf{Average}}\\ 
      \cline{2-11}
       & \textbf{C1} & \textbf{C2} & \textbf{C3} & \textbf{C4} & \textbf{C5} & \textbf{C6} & \textbf{C7} & \textbf{C8} & \textbf{C9} & \textbf{C10}  &\\
      \hline
       \multicolumn{12}{l}{\textbf{Proprietary LLMs}} \\
      Claude-3.5-Sonnet & 48.77 & 57.72 & 66.02 & 57.54 & 47.82 & 49.36 & 57.47 & 60.11 & 58.06 & 58.89 & 55.45 \\
      Gemini-1.5-Pro & 47.00 & 59.55 & 62.42 & 59.56 & 43.11 & 49.55 & 53.41 & 56.18 & 52.51 & 53.71 & 53.58 \\
      Gemini-2.0-Flash & 33.67 & 27.37 & 53.04 & 32.07 & 25.87 & 44.41 & 33.57 & 36.77 & 31.78 & 31.70 & 34.24 \\
      GLM-4-Plus & 52.72 & \cellcolor{red!5}59.62 & 70.62 & \cellcolor{red!5}60.11 & 50.60 & 56.75 & \cellcolor{red!10}65.02 & \cellcolor{red!5}64.17 & \cellcolor{red!10}61.70 & \cellcolor{red!5}62.90 & \cellcolor{red!5}59.61 \\
      GPT-4 & \cellcolor{red!20}59.59 & \cellcolor{red!10}60.55 & \cellcolor{red!20}76.32 & \cellcolor{red!10}61.16 & \cellcolor{red!20}56.34 & \cellcolor{red!20}59.35 & 63.67 & \cellcolor{red!10}64.74 & \cellcolor{red!5}60.65 & \cellcolor{red!10}67.66 & \cellcolor{red!10}62.06 \\
      GPT-4o mini & \cellcolor{red!5}54.24 & 56.64 & 72.11 & 59.28 & 53.00 & \cellcolor{red!5}57.88 & 58.38 & 61.75 & 57.50 & 62.38 & 58.40 \\
      OpenAI o1-mini & 49.16 & 55.58 & 59.37 & 54.77 & 44.43 & 50.73 & 54.57 & 55.36 & 54.91 & 54.19 & 53.25 \\
      \hline
      \multicolumn{12}{l}{\textbf{Open-Source LLMs}} \\
      DeepSeek-V3-671B & \cellcolor{red!10}56.03 & \cellcolor{red!20}62.42 & \cellcolor{red!10}74.81 & \cellcolor{red!20}63.17 & \cellcolor{red!10}55.23 & \cellcolor{red!10}58.84 & \cellcolor{red!20}68.23 & \cellcolor{red!20}69.04 & \cellcolor{red!20}66.46 & \cellcolor{red!20}68.48 & \cellcolor{red!20}63.30 \\
      GLM-4-Chat-9B & 23.28 & 21.31 & 39.97 & 26.13 & 16.20 & 34.15 & 26.63 & 29.60 & 25.60 & 26.68 & 26.55 \\
        & (±0.11) & (±0.12) & (±0.43) & (±0.12) & (±0.01) & (±0.12) & (±0.08) & (±0.01) & (±0.02) & (±0.05) & (±0.03) \\
        InternLM2-7B & 27.55 & 21.14 & 39.64 & 28.57 & 15.16 & 36.12 & 28.74 & 30.80 & 27.32 & 29.22 & 28.71 \\
        & (±0.01) & (±0.01) & (±0.00) & (±0.00) & (±0.00) & (±0.03) & (±0.02) & (±0.04) & (±0.01) & (±0.02) & (±0.01) \\
        InternLM2.5-7B & 51.71 & 55.75 & 67.88 & 50.48 & 44.14 & 56.73 & 51.28 & 54.91 & 52.46 & 56.24 & 53.51 \\
        & (±0.01) & (±0.03) & (±0.02) & (±0.14) & (±0.01) & (±0.01) & (±0.02) & (±0.03) & (±0.01) & (±0.01) & (±0.03) \\
        Llama3.1-8B & 43.89 & 31.21 & 42.53 & 40.68 & 38.47 & 43.80 & 42.87 & 51.62 & 41.88 & 40.91 & 42.23 \\
        & (±0.35) & (±1.07) & (±0.30) & (±0.16) & (±0.38) & (±0.03) & (±0.05) & (±0.05) & (±0.51) & (±0.14) & (±0.23) \\
        Llama3.1-70B & 48.72 & 55.41 & 64.77 & 53.67 & 46.73 & 54.08 & 56.94 & 57.72 & 55.31 & 57.56 & 54.30 \\
        & (±0.07) & (±0.01) & (±0.00) & (±0.29) & (±0.00) & (±0.01) & (±0.12) & (±0.09) & (±0.11) & (±0.56) & (±0.13) \\
        Llama3.3-70B & 45.32 & 47.15 & 60.62 & 49.76 & 40.90 & 54.30 & 52.79 & 54.61 & 49.98 & 55.05 & 50.53 \\
        & (±0.00) & (±0.06) & (±0.00) & (±0.00) & (±0.00) & (±0.09) & (±0.06) & (±0.00) & (±0.02) & (±0.00) & (±0.01) \\
        Mistral-v0.3-7B & 42.61 & 38.28 & 57.02 & 40.41 & 29.97 & 44.22 & 36.31 & 43.98 & 39.92 & 43.51 & 41.59 \\
        & (±0.32) & (±0.19) & (±0.14) & (±0.15) & (±0.21) & (±0.06) & (±0.39) & (±0.23) & (±0.10) & (±0.05) & (±0.06) \\
        Qwen2-0.5B & 32.84 & 25.15 & 40.19 & 28.20 & 27.62 & 37.22 & 33.81 & 33.63 & 28.25 & 31.67 & 31.44 \\
        & (±0.26) & (±0.64) & (±0.64) & (±0.24) & (±0.49) & (±0.19) & (±0.58) & (±0.21) & (±0.33) & (±0.22) & (±0.13) \\
        Qwen2-7B & 44.21 & 40.41 & 63.00 & 47.36 & 35.37 & 52.30 & 45.61 & 48.73 & 44.88 & 46.89 & 46.51 \\
        & (±0.03) & (±0.10) & (±0.11) & (±0.10) & (±0.12) & (±0.27) & (±0.06) & (±0.14) & (±0.13) & (±0.17) & (±0.02) \\
        Qwen2-57B & 53.67 & 49.81 & \cellcolor{red!5}74.30 & 58.38 & 39.34 & 54.71 & \cellcolor{red!5}63.89 & 59.57 & 59.22 & 60.08 & 57.20 \\
        & (±0.21) & (±0.05) & \cellcolor{red!5}(±0.37) & (±0.04) & (±0.40) & (±0.16) & \cellcolor{red!5}(±0.37) & (±0.04) & (±0.01) & (±0.07) & (±0.01) \\
        Qwen2-72B & 51.16 & 58.10 & 74.07 & 59.72 & \cellcolor{red!5}51.58 & 57.76 & 58.85 & 61.63 & 56.69 & 59.11 & 57.62 \\
        & (±1.70) & (±4.07) & (±0.04) & (±3.44) & \cellcolor{red!5}(±0.42) & (±0.75) & (±2.35) & (±4.89) & (±2.16) & (±3.33) & (±2.60) \\
        Qwen2.5-7B & 45.16 & 39.50 & 66.01 & 44.61 & 35.72 & 50.00 & 53.60 & 53.31 & 53.06 & 51.05 & 48.45 \\
        & (±0.43) & (±0.19) & (±0.36) & (±0.19) & (±0.51) & (±0.24) & (±0.29) & (±0.21) & (±0.20) & (±0.31) & (±0.10) \\
        Qwen2.5-14B & 50.91 & 50.73 & 68.62 & 52.15 & 47.14 & 54.54 & 57.02 & 62.05 & 54.37 & 54.15 & 54.21 \\
        & (±0.00) & (±0.00) & (±0.00) & (±0.06) & (±0.00) & (±0.00) & (±0.03) & (±0.00) & (±0.12) & (±0.00) & (±0.02) \\
        Qwen2.5-72B & 46.86 & 47.41 & 70.99 & 51.89 & 46.17 & 57.60 & 55.35 & 56.31 & 53.05 & 54.75 & 52.63 \\
        & (±0.01) & (±0.01) & (±0.00) & (±0.00) & (±0.00) & (±0.13) & (±0.00) & (±0.01) & (±0.01) & (±0.01) & (±0.02) \\
        QwQ-32B & 32.24 & 21.06 & 47.11 & 29.14 & 28.56 & 39.68 & 38.17 & 39.56 & 34.70 & 34.52 & 33.55 \\
        & (±0.00) & (±0.00) & (±0.00) & (±0.00) & (±0.00) & (±0.00) & (±0.00) & (±0.00) & (±0.00) & (±0.00) & (±0.00) \\
        \hline
        \multicolumn{12}{l}{\textbf{Domain Specific LLMs}} \\
        Aksara-v1-7B & 36.72 & 36.69 & 48.32 & 35.41 & 24.26 & 36.83 & 31.17 & 34.64 & 31.15 & 34.14 & 35.04 \\
        & (±0.18) & (±0.10) & (±0.27) & (±0.06) & (±0.02) & (±0.22) & (±0.03) & (±0.02) & (±0.01) & (±0.05) & (±0.05) \\
        PLLaMa-7B & 17.85 & 13.69 & 17.99 & 16.81 & 11.66 & 21.67 & 14.34 & 17.36 & 12.39 & 16.11 & 16.46 \\
        & (±0.03) & (±0.07) & (±0.44) & (±0.14) & (±0.20) & (±0.34) & (±0.21) & (±0.03) & (±0.17) & (±0.22) & (±0.01) \\
        PLLaMa-13B & 15.10 & 14.18 & 28.41 & 18.83 & 13.96 & 23.28 & 18.53 & 17.37 & 14.15 & 18.51 & 17.57 \\
        & (±0.04) & (±0.03) & (±0.36) & (±0.03) & (±0.01) & (±0.07) & (±0.12) & (±0.21) & (±0.08) & (±0.03) & (±0.03) \\
        \hline
    \end{tabularx}}
\caption{Evaluation of 26 LLMs on SeedBench. Performance is stratified by breeding subcategories, with open-source/domain-specific models evaluated through 3 repeated trials (mean scores reported). The scores represent averages across three different metrics for 11 task types. The columns delineate ten subcategories in breeding: (C1) Gene Basic Information Query, (C2) Gene Expression Pattern Query, (C3) Gene Product Cellular Localization Query, (C4) Gene Function Experimental Observation, (C5) Gene Product Regulation of Downstream Genes Analysis, (C6) Gene Function Prediction, (C7) Variety Breeding Process Query, (C8) Variety Agronomic Trait Query, (C9) Variety Cultivation and Technical Key Points Query, (C10) Variety Suitable Planting Area Recommendation. Top-$3$ performers per column are highlighted in red.}
  \label{tab:performance_std}
\end{table*}

\newpage
\begin{table*}[h!]
\centering
\resizebox{\textwidth}{!}{
\small
\setlength{\tabcolsep}{0.7pt}
\begin{tabularx}{\textwidth}{l*{12}{r}}
\cline{1-13}
\multirow{2}{*}{\textbf{Models}} & \multicolumn{11}{c}{\textbf{Question Type}} & \multirow{2}{*}{\textbf{Average}} \\
\cline{2-12}
& \textbf{QA-1} & \textbf{QA-2} & \textbf{QA-3} & \textbf{QA-4} & \textbf{SUM-1} & \textbf{SUM-2} & \textbf{RC-1} & \textbf{RC-2} & \textbf{RC-3} & \textbf{RC-4} & \textbf{RC-5} & \\
\cline{1-13}
\multicolumn{13}{l}{\textbf{Proprietary LLMs}} \\
\textbf{Claude-3.5-Sonnet} &
  57.50 & 74.68 & \cellcolor{red!5}25.84 & 21.82 & 33.97 & 44.94 & 98.23 & \cellcolor{red!20}97.67 & 85.97 & 44.58 & 88.17 & 61.22\\
\textbf{Gemini-1.5-Pro} &
  56.50 & 73.26 & \cellcolor{red!10}27.37 & 16.96 & 22.99 & 28.88 & 99.12 & \cellcolor{red!5}97.01 & \cellcolor{red!5}86.75 & 43.18 & 84.95 & 57.91\\
\textbf{Gemini-2.0-Flash} &
  62.00 & 58.59 & 1.83 & 11.04 & 31.01 & 27.76 & 97.35 & 65.25 & 7.37 & 25.74 & \cellcolor{red!10}95.34 & 43.93\\
\textbf{GLM-4-Plus} &
  \cellcolor{red!5}64.50 & \cellcolor{red!5}75.40 & 25.82 & 32.51 & 48.53 & 51.53 & 99.12 & 97.02 & 84.39 & 49.31 & 85.30 & 64.85\\
\textbf{GPT-4o mini} &
  57.50 & 72.33 & 17.88 & \cellcolor{red!20}44.47 & 49.51 & 60.66 & 97.35 & 95.38 & 85.30 & \cellcolor{red!5}57.73 & 84.23 & \cellcolor{red!5}65.67\\
\textbf{GPT-4} &
  60.50 & 73.87 & 21.35 & 36.07 & \cellcolor{red!20}58.73 & \cellcolor{red!10}62.89 & \cellcolor{red!20}100.00 & 96.44 & \cellcolor{red!10}87.86 & \cellcolor{red!20}62.29 & 86.74 & \cellcolor{red!10}67.88\\
\textbf{OpenAI o1-mini} &
  57.50 & 73.81 & 22.25 & 27.93 & 38.40 & 37.80 & \cellcolor{red!20}100.00 & 96.17 & 83.46 & 36.40 & 82.80 & 59.68\\
\textbf{DeepSeek-V3} &
  \cellcolor{red!20}72.50 & \cellcolor{red!20}79.84 & \cellcolor{red!20}29.29 & \cellcolor{red!10}40.63 & 48.06 & 54.67 & \cellcolor{red!20}100.00 & \cellcolor{red!10}97.22 & \cellcolor{red!20}87.89 & 55.19 & 86.74 & \cellcolor{red!20}68.37\\
\cline{1-13}
\multicolumn{13}{l}{\textbf{Open-Source LLMs}} \\
\textbf{GLM-4-Chat-9B} &
  53.00 & 62.41 & 0.95 & 1.57 & 5.69 & 13.38 & 98.53 & 68.51 & 6.26 & 2.29 & \cellcolor{red!5}93.55 & 36.92\\
& (±0.00) & (±0.10) & (±0.00) & (±0.01) & (±0.03) & (±0.02) & (±0.51) & (±0.29) & (±0.01) & (±0.01) & (±0.36) \\
\textbf{InternLM2-7B} &
  56.50 & 66.93 & 1.85 & 2.14 & 8.66 & 13.88 & 98.23 & 85.88 & 14.31 & 6.51 & \cellcolor{red!20}96.77 & 41.06\\
& (±0.00) & (±0.17) & (±0.00) & (±0.00) & (±0.07) & (±0.03) & (±0.00) & (±0.00) & (±0.02) & (±0.02) & (±0.00) \\
\textbf{InternLM2.5-7B} &
  58.50 & 65.86 & 12.99 & 32.69 & 50.98 & 53.64 & 99.12 & 94.48 & 73.39 & 55.41 & 70.37 & 60.67\\
& (±0.00) & (±0.00) & (±0.05) & (±0.07) & (±0.10) & (±0.02) & (±0.00) & (±0.00) & (±0.00) & (±0.01) & (±0.21) \\
\textbf{Llama3.1-8B} &
  48.00 & 64.33 & 10.60 & 24.57 & 48.11 & 42.62 & 94.69 & 81.05 & 54.31 & 45.63 & 74.67 & 53.51\\
& (±0.00) & (±0.30) & (±0.24) & (±0.12) & (±0.14) & (±0.09) & (±0.00) & (±0.09) & (±0.76) & (±0.47) & (±0.21) \\
\textbf{Llama3.1-70B} &
  56.00 & 73.73 & 19.53 & 34.65 & 52.09 & 54.83 & 99.12 & 96.62 & 74.76 & 51.28 & 83.15 & 63.25\\
& (±0.71) & (±0.07) & (±0.09) & (±0.00) & (±0.00) & (±0.02) & (±0.00) & (±0.00) & (±0.98) & (±0.01) & (±0.01) \\
\textbf{Llama3.3-70B} &
  58.50 & 71.50 & 18.65 & 26.55 & 47.79 & 53.03 & 99.12 & 93.88 & 64.92 & 49.96 & 80.29 & 60.38\\
& (±0.00) & (±0.00) & (±0.00) & (±0.01) & (±0.13) & (±0.01) & (±0.00) & (±0.00) & (±0.09) & (±0.02) & (±0.00) \\
\textbf{Mistral-v0.3-7B} &
  39.50 & 58.14 & 5.91 & 30.01 & 45.02 & \cellcolor{red!5}62.03 & 83.19 & 76.43 & 38.51 & 50.31 & 73.12 & 51.11\\
& (±0.00) & (±0.19) & (±0.18) & (±0.42) & (±0.05) & (±0.03) & (±0.00) & (±0.00) & (±0.22) & (±0.53) & (±0.00) \\
\textbf{Qwen2-0.5B} &
  40.50 & 62.20 & 2.82 & 31.21 & 44.72 & 53.73 & 55.46 & 65.89 & 15.98 & 42.04 & 50.18 & 42.25\\
& (±0.00) & (±0.29) & (±0.07) & (±0.47) & (±0.20) & (±0.34) & (±1.35) & (±0.61) & (±0.08) & (±0.76) & (±0.72) \\
\textbf{Qwen2-7B} &
  57.00 & 69.44 & 12.72 & 32.42 & 31.18 & 40.30 & 97.35 & 82.36 & 57.83 & 46.96 & 84.59 & 55.65\\
& (±0.00) & (±0.09) & (±0.17) & (±0.08) & (±0.00) & (±0.08) & (±0.00) & (±0.17) & (±0.20) & (±0.28) & (±0.00) \\
\textbf{Qwen2-57B} &
  56.00 & 74.79 & 21.24 & \cellcolor{red!5}39.56 & 46.04 & 60.27 & 99.12 & 95.95 & 76.55 & \cellcolor{red!10}56.66 & 80.83 & 64.27\\
& (±0.00) & (±0.03) & (±0.11) & (±0.23) & (±0.03) & (±0.64) & (±0.00) & (±0.00) & (±0.09) & (±0.25) & (±0.25) \\
\textbf{Qwen2-72B} &
  59.50 & \cellcolor{red!10}75.98 & 19.55 & 31.62 & 31.08 & \cellcolor{red!20}63.09 & 99.12 & 94.24 & 72.20 & 51.58 & 89.96 & 62.54\\
& (±0.00) & (±0.00) & (±0.00) & (±0.00) & (±0.00) & (±0.03) & (±0.00) & (±0.00) & (±0.00) & (±0.00) & (±0.00) \\
\textbf{Qwen2.5-7B} &
  57.00 & 71.19 & 17.46 & 32.47 & 43.96 & 54.70 & 96.46 & 88.33 & 54.47 & 47.97 & 79.21 & 58.47\\
& (±1.00) & (±0.24) & (±0.10) & (±0.09) & (±0.01) & (±0.07) & (±0.00) & (±0.07) & (±0.00) & (±0.02) & (±0.00) \\
\textbf{Qwen2.5-14B} &
  57.00 & 72.49 & 17.43 & 22.66 & \cellcolor{red!10}54.88 & 56.20 & 99.12 & 93.36 & 67.72 & 51.57 & 83.51 & 61.45\\
& (±0.00) & (±0.00) & (±0.00) & (±0.00) & (±0.08) & (±0.00) & (±0.00) & (±0.00) & (±0.25) & (±0.00) & (±0.00) \\
\textbf{Qwen2.5-72B} &
  \cellcolor{red!10}70.50 & 73.71 & 17.86 & 29.84 & \cellcolor{red!5}51.33 & 59.52 & \cellcolor{red!20}100.00 & 87.79 & 57.07 & 48.41 & 83.51 & 61.77\\
& (±0.00) & (±0.00) & (±0.00) & (±0.01) & (±0.00) & (±0.03) & (±0.00) & (±0.00) & (±0.00) & (±0.00) & (±0.00) \\
\textbf{QwQ-32B} &
  61.50 & 58.81 & 3.00 & 14.54 & 15.60 & 26.93 & 91.15 & 64.17 & 19.23 & 19.09 & 91.04 & 42.28\\
& (±0.00) & (±0.00) & (±0.00) & (±0.00) & (±0.00) & (±0.00) & (±0.00) & (±0.00) & (±0.00) & (±0.00) & (±0.00) \\
\cline{1-13}
\multicolumn{13}{l}{\textbf{Domain Specific LLMs}} \\
\textbf{Aksara-v1-7B} &
  34.167 & 53.4 & 4.28 & 17.263 & 26.503 & 38.32 & 87.61 & 72.89 & 30.037 & 35.397 & 66.67 & 42.41 \\
 & (±0.289) & (±0.139) & (±0.061) & (±0.025) & (±0.006) & (±0.00) & (±0.00) & (±0.00) & (±0.075) & (±0.396) & (±0.00) & \\
\textbf{PLLaMa-7B} &
  6.5 & 35.47 & 2.613 & 20.223 & 13.753 & 20.577 & 6.19 & 35.03 & 12.76 & 24.583 & 4.78 & 16.59 \\
 & (±0.5) & (±0.485) & (±0.032) & (±0.326) & (±0.126) & (±0.046) & (±0.00) & (±0.00) & (±0.00) & (±0.167) & (±0.208) & \\
 \textbf{PLLaMa-13B} &
  24.167 & 50.433 & 1.013 & 18.17 & 13.37 & 14.84 & 38.94 & 57.37 & 6.95 & 18.007 & 40.98 & 25.84 \\
 & (±0.289) & (±0.133) & (±0.04) & (±0.171) & (±0.00) & (±0.00) & (±0.00) & (±0.00) & (±0.00) & (±0.243) & (±0.416) & \\
 
\cline{1-13}
\end{tabularx}
}
\caption{Evaluation results (zero-shot) on different question type. Top-3 per column are highlighted.}
\label{tab:res_task_type_zeroshot}
\end{table*}

\newpage
\begin{table}[h!]
\centering
\resizebox{\textwidth}{!}{
\small
\setlength{\tabcolsep}{1.1pt}
\begin{tabularx}{\textwidth}{l*{12}{r}}
\cline{1-13}
\multirow{2}{*}{\textbf{Models}} & \multicolumn{11}{c}{\textbf{Question Type}} & \multirow{2}{*}{\textbf{Average}} \\
\cline{2-12}
& \textbf{QA-1} & \textbf{QA-2} & \textbf{QA-3} & \textbf{QA-4} & \textbf{SUM-1} & \textbf{SUM-2} & \textbf{RC-1} & \textbf{RC-2} & \textbf{RC-3} & \textbf{RC-4} & \textbf{RC-5} & \\
\cline{1-13}
\multicolumn{13}{l}{\textbf{Proprietary LLMs}} \\
\textbf{Claude-3.5-Sonnet} &
  53.50 & 75.54 & \cellcolor{red!10}27.92 & 23.99 & 32.36 & 42.88 & 98.23 & \cellcolor{red!20}97.67 & 85.68 & 45.40 & 86.02 & 60.84\\
\textbf{Gemini-1.5-Pro} &
  55.50 & 73.49 & \cellcolor{red!5}27.13 & 27.03 & 28.98 & 35.65 & 99.12 & \cellcolor{red!10}97.45 & \cellcolor{red!10}86.98 & 46.08 & 86.74 & 60.38\\
\textbf{Gemini-2.0-Flash} &
  64.50 & 59.30 & 1.73 & 12.99 & 31.99 & 36.43 & 97.35 & 65.91 & 6.25 & 26.63 & \cellcolor{red!20}93.19 & 45.12\\
\textbf{GLM-4-Plus} &
  \cellcolor{red!5}67.00 & \cellcolor{red!5}76.46 & 26.47 & \cellcolor{red!5}46.13 & 51.35 & 52.73 & 99.12 & 96.62 & 85.99 & \cellcolor{red!5}59.14 & 80.29 & 67.39\\
\textbf{GPT-4o mini} &
  55.00 & 71.89 & 19.10 & 41.59 & 50.21 & 57.59 & 97.35 & 95.00 & 86.00 & 53.76 & 82.08 & 64.51\\
\textbf{GPT-4} &
  57.50 & 73.24 & 23.43 & \cellcolor{red!10}46.19 & \cellcolor{red!20}63.93 & \cellcolor{red!20}68.07 & \cellcolor{red!20}100.00 & 96.86 & \cellcolor{red!20}88.97 & \cellcolor{red!20}64.53 & 83.15 & 69.63\\
\textbf{OpenAI o1-mini} &
  61.50 & 74.78 & 19.82 & 35.75 & 32.89 & 25.91 & 99.12 & 96.84 & 83.31 & 47.12 & 80.65 & 59.79\\
\textbf{DeepSeek-V3} &
  \cellcolor{red!20}71.50 & \cellcolor{red!20}79.16 & \cellcolor{red!20}32.28 & \cellcolor{red!20}46.59 & 50.16 & 58.52 & \cellcolor{red!20}100.00 & \cellcolor{red!5}97.00 & \cellcolor{red!5}86.85 & \cellcolor{red!10}60.59 & 85.30 & 69.81\\
\cline{1-13}
\multicolumn{13}{l}{\textbf{Open-Source LLMs}} \\
\textbf{GLM-4-Chat-9B} &
  55.50 & 59.31 & 0.51 & 1.24 & 4.46 & 11.12 & 97.35 & 64.35 & 7.67 & 1.92 & \cellcolor{red!10}91.16 & 35.87 \\
 & (±0.00) & (±0.00) & (±0.00) & (±0.01) & (±0.01) & (±0.00) & (±0.00) & (±0.00) & (±0.00) & (±0.00) & (±0.41) & \\
\textbf{InternLM2-7B} &
  50.50 & 59.30 & 0.16 & 1.55 & 7.82 & 12.76 & 94.69 & 75.12 & 6.35 & 2.98 & \cellcolor{red!5}88.17 & 36.31 \\
 & (±0.00) & (±0.03) & (±0.00) & (±0.01) & (±0.00) & (±0.00) & (±0.00) & (±0.00) & (±0.00) & (±0.01) & (±0.97) & \\
\textbf{InternLM2.5-7B} &
  58.50 & 69.42 & 14.74 & 35.05 & 53.63 & 56.07 & 99.12 & 95.18 & 75.04 & 57.03 & 68.22 & 62.00 \\
 & (±0.00) & (±0.00) & (±0.00) & (±0.12) & (±0.00) & (±0.00) & (±0.00) & (±0.00) & (±0.00) & (±0.11) & (±0.06) & \\
\textbf{Llama3.1-8B} &
  35.50 & 49.10 & 7.97 & 32.90 & 49.22 & 46.62 & 86.43 & 75.39 & 54.04 & 41.27 & 68.10 & 49.68 \\
 & (±0.00) & (±0.00) & (±0.01) & (±0.63) & (±0.07) & (±0.01) & (±0.51) & (±0.38) & (±0.00) & (±0.01) & (±0.62) & \\
\textbf{Llama3.1-70B} &
  55.00 & 70.01 & 19.60 & 33.17 & 52.24 & 59.04 & 99.12 & 96.02 & 74.49 & 51.42 & 80.65 & 62.80 \\
 & (±0.00) & (±0.06) & (±0.00) & (±0.07) & (±0.00) & (±0.06) & (±0.00) & (±0.00) & (±0.33) & (±0.00) & (±0.33) & \\
\textbf{Llama3.3-70B} &
  54.75 & 73.55 & 17.35 & 30.61 & 50.09 & 45.12 & 99.12 & 93.52 & 63.22 & 45.62 & 79.21 & 59.29 \\
 & (±0.35) & (±0.13) & (±0.00) & (±0.06) & (±0.00) & (±0.01) & (±0.00) & (±0.00) & (±0.00) & (±0.06) & (±0.22) & \\
\textbf{Mistral-v0.3-7B} &
  39.50 & 60.85 & 3.82 & 31.09 & 45.74 & 62.27 & 79.06 & 75.79 & 35.65 & 48.81 & 74.55 & 50.65 \\
 & (±0.50) & (±0.13) & (±0.01) & (±0.12) & (±0.00) & (±0.00) & (±0.51) & (±0.00) & (±0.30) & (±0.10) & (±0.75) & \\
\textbf{Qwen2-0.5B} &
  44.50 & 62.94 & 2.28 & 33.36 & 36.05 & 38.12 & 67.26 & 68.93 & 12.10 & 40.83 & 25.09 & 39.22 \\
 & (±0.00) & (±0.31) & (±0.25) & (±0.65) & (±0.38) & (±0.44) & (±0.00) & (±0.42) & (±0.98) & (±0.83) & (±0.46) & \\
\textbf{Qwen2-7B} &
  52.17 & 65.62 & 14.13 & 34.50 & 43.35 & 52.33 & 88.50 & 78.53 & 52.03 & 48.77 & 85.42 & 55.94 \\
 & (±0.58) & (±0.79) & (±0.02) & (±0.18) & (±0.00) & (±0.00) & (±0.00) & (±0.08) & (±0.24) & (±0.01) & (±0.31) & \\
\textbf{Qwen2-57B} &
  58.00 & 76.42 & 21.17 & 43.49 & 49.60 & 60.53 & 97.35 & 95.28 & 76.86 & 56.87 & 77.42 & 64.82 \\
 & (±0.00) & (±0.00) & (±0.09) & (±0.36) & (±0.01) & (±0.12) & (±0.00) & (±0.00) & (±0.16) & (±0.16) & (±0.63) & \\
\textbf{Qwen2-72B} &
  65.00 & \cellcolor{red!10}77.73 & 24.51 & 41.70 & 54.36 & \cellcolor{red!10}66.48 & 99.12 & 96.13 & 78.54 & 57.85 & 84.41 & 67.80 \\
 & (±0.00) & (±0.13) & (±2.57) & (±0.96) & (±0.07) & (±0.01) & (±0.00) & (±1.57) & (±7.40) & (±0.69) & (±0.57) & \\
\textbf{Qwen2.5-7B} &
  57.667 & 70.32 & 16.867 & 38.27 & 43.897 & 59.597 & 92.92 & 80.093 & 52.003 & 52.46 & 75.987 & 58.19 \\
 & (±0.00) & (±1.11) & (±0.62) & (±0.29) & (±0.18) & (±0.40) & (±0.00) & (±0.00) & (±0.30) & (±0.33) & (±0.39) & \\
\textbf{Qwen2.5-14B} &
  63.5 & 72.57 & 18.79 & 38.4 & \cellcolor{red!20}56.24 & \cellcolor{red!5}62.14 & 99.12 & 94.39 & 69.877 & 56.83 & 82.08 & 64.90 \\
 & (±0.00) & (±0.00) & (±0.00) & (±0.00) & (±0.00) & (±0.00) & (±0.00) & (±0.00) & (±0.335) & (±0.00) & (±0.00) & \\
 \textbf{Qwen2.5-72B} &
  \cellcolor{red!10}67.50 & 74.79 & 21.39 & 40.425 & \cellcolor{red!5}55.86 & 62.01 & 99.12 & 89.11 & 55.995 & 51.38 & 82.8 & 63.67 \\
 & (±0.00) & (±0.00) & (±0.00) & (±0.078) & (±0.00) & (±0.00) & (±0.00) & (±0.00) & (±0.007) & (±0.028) & (±0.00) & \\
\textbf{QwQ-32B} &
  63.50 & 60.26 & 6.81 & 25.35 & 13.12 & 20.35 & 93.81 & 65.07 & 22.72 & 27.84 & 75.99 & 43.17 \\
 & (±0.00) & (±0.00) & (±0.00) & (±0.00) & (±0.00) & (±0.00) & (±0.00) & (±0.00) & (±0.00) & (±0.00) & (±0.00) & \\
\cline{1-13}
\multicolumn{13}{l}{\textbf{Domain Specific LLMs}} \\
\textbf{Aksara-v1-7B} &
  41.667 & 55.627 & 2.653 & 23.35 & 38.7 & 57.14 & 84.07 & 70.88 & 27.037 & 40.087 & 71.33 & 50.65 \\
 & (±0.289) & (±0.182) & (±0.072) & (±0.157) & (±0.00) & (±0.00) & (±0.00) & (±0.00) & (±0.167) & (±0.389) & (±0.00) & \\
 
\textbf{PLLaMa-7B} &
  11.167 & 37.22 & 6.437 & 25.203 & 30.42 & 27.71 & 5.31 & 28.38 & 20.27 & 24.723 & 29.75 & 22.42 \\
 & (±0.577) & (±0.859) & (±0.202) & (±0.117) & (±0.00) & (±0.00) & (±0.00) & (±0.00) & (±0.208) & (±0.067) & (±0.00) & \\

 \textbf{PLLaMa-13B} &
  24.0 & 52.697 & 0.507 & 7.167 & 12.34 & 13.73 & 30.09 & 50.51 & 3.81 & 15.55 & 32.26 & 22.06 \\
 & (±0.00) & (±0.029) & (±0.006) & (±0.086) & (±0.00) & (±0.00) & (±0.00) & (±0.00) & (±0.035) & (±0.00) & (±0.00) & \\
\cline{1-13}
\end{tabularx}
}
\caption{Evaluation results (one-shot) on different question type. Top-3 per column are highlighted.}
\label{tab:res_task_type_oneshot}

\end{table}

\begin{figure*}[h!]
  \centering
    \includegraphics[width=\linewidth]{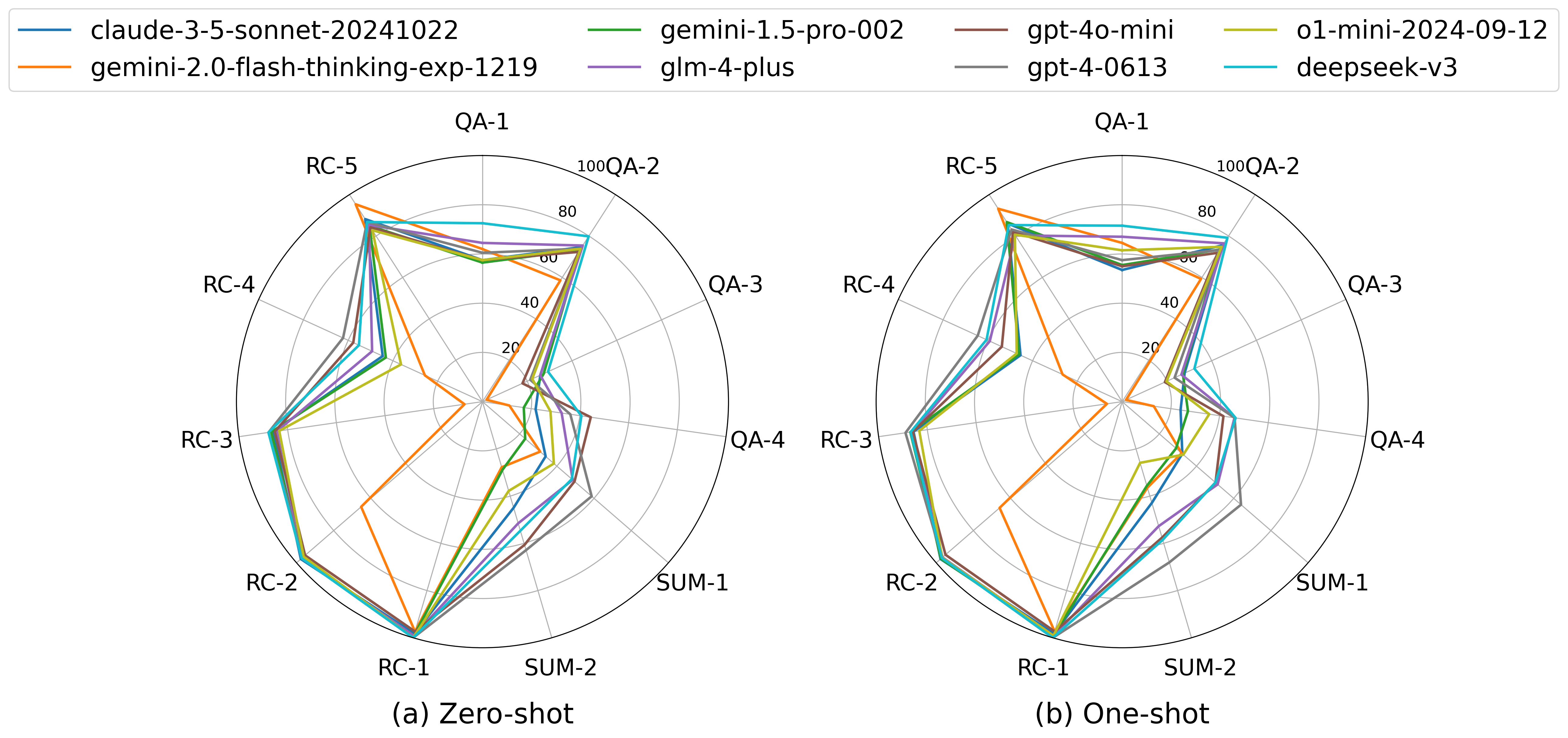}
    \caption{Evaluation of Proprietary LLMs on SeedBench. Performance is stratified by task-type.}
    \label{ffiig:redar1}
\end{figure*}

\begin{figure*}[h!]
  \centering
    \includegraphics[width=\linewidth]{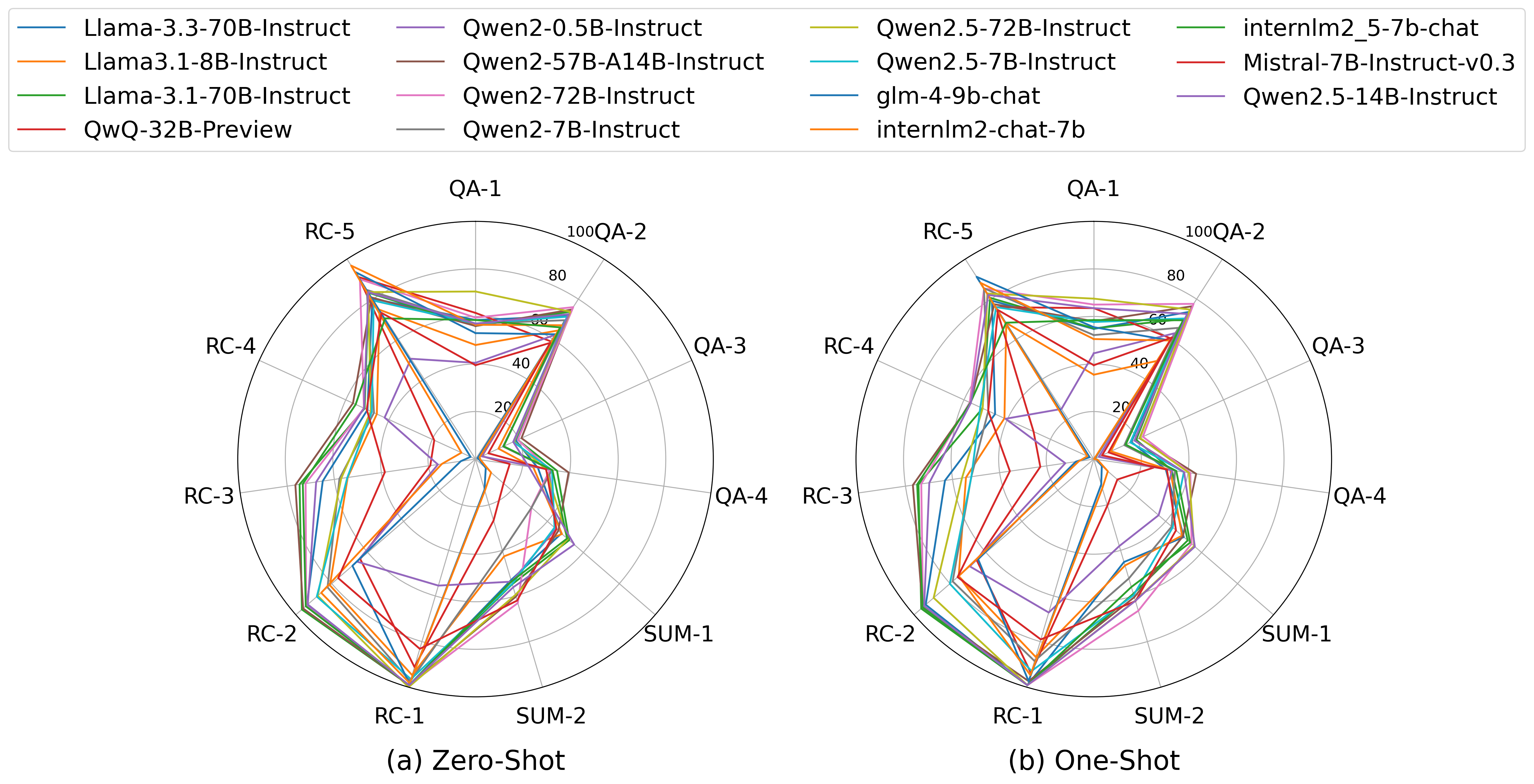}
    \caption{Evaluation of Open-Source LLMs on SeedBench. Performance is stratified by task-type.}
    \label{ffiig:redar2}
\end{figure*}

\end{document}